\documentclass[11pt]{article}
\usepackage[english]{babel}
\usepackage[utf8]{inputenc}
\usepackage[round]{natbib}
\usepackage[margin=0.94in]{geometry}
\usepackage{booktabs}
\usepackage{amsmath}
\usepackage{amssymb}
\usepackage{mathtools}
\usepackage{mathrsfs}
\usepackage{bm}
\usepackage{multirow}
\usepackage{color}
\usepackage{xcolor}
\PassOptionsToPackage{dvipsnames,svgnames,table}{xcolor}
\definecolor{darkblue}{rgb}{0,0,1}
\usepackage[utf8]{inputenc} 
\usepackage[T1]{fontenc}    
\usepackage[colorlinks=true,
            urlcolor=purple,
            linkcolor=red,
            hyperfootnotes=false,
            citecolor=blue]{hyperref}       
\usepackage{url}            
\usepackage{booktabs}       
\usepackage{amsfonts}       
\usepackage{nicefrac}       
\usepackage{microtype}      
\usepackage{graphicx}
\usepackage{subcaption}
\usepackage{amsmath}
\usepackage{amssymb}
\usepackage{amsthm}
\usepackage{wrapfig}
\usepackage{algorithm}
\usepackage{algpseudocode}
\usepackage{makecell}
\usepackage{float}
\usepackage{array}
\usepackage{colortbl}
\usepackage{comment}
\usepackage{longtable}
\usepackage[misc]{ifsym}

\newcommand{\m}{\mathbf}

\usepackage{bm}
\usepackage{bbm}

\usepackage{booktabs}
\usepackage{makecell}
\usepackage{multirow}

\usepackage{comment}

\makeatletter
\renewcommand{\fnum@algorithm}{\fname@algorithm}
\makeatother

\newtheorem{theorem}{Theorem}[section]

\newtheorem{assumption}{A\hspace{-4pt}}

\usepackage{mathrsfs}

\newtheorem{lemmaA}{Lemma}

\usepackage{cleveref}
\crefname{assumption}{assumption}{assumptions}
\Crefname{assumption}{Assumption}{Assumptions}

\crefname{theorem}{theorem}{theorems}
\Crefname{theorem}{Theorem}{Theorems}

\crefname{lemmaA}{lemma}{lemmas}
\Crefname{lemmaA}{Lemma}{Lemmas}

\crefname{figure}{fig.}{}
\Crefname{figure}{Fig.}{}

\newcommand{\MYhref}[3][blue]{\href{#2}{\color{#1}{#3}}}%

\allowdisplaybreaks

\title{\textbf{Robust Classification of High-Dimensional Data\\ using Data-Adaptive Energy Distance}}

\usepackage[max2,noblocks]{authblk}
\usepackage[symbol]{footmisc}

\newcommand\blfootnote[1]{%
  \begingroup
  \renewcommand\thefootnote{}\footnote{#1}%
  \addtocounter{footnote}{-1}%
  \endgroup
}

\newcommand{\myFootnote}[1]{%
  \renewcommand{\thefootnote}{$*$}%
  \footnote{#1}%
  \setcounter{footnote}{1}
  \renewcommand{\thefootnote}{\fnsymbol{footnote}}
}

\author[1]{Jyotishka Ray Choudhury{\protect\myFootnote{This work was completed while the author was affiliated with the \textsc{Indian Statistical Institute}, Kolkata, India.}}\thanks{\textbf{Joint first authors contributed equally to this research.}\vspace{1mm}}}
\renewcommand{\thefootnote}{\fnsymbol{footnote}}
\author[2]{Aytijhya Saha\protect\footnotemark[2]}
\author[3]{Sarbojit Roy}
\author[4,5]{Subhajit Dutta}

\affil[1]{School of Industrial and Systems Engineering, \textsc{Georgia Institute of Technology}, Atlanta, USA}
\affil[2]{\textsc{Indian Statistical Institute}, Kolkata, India}
\affil[3]{Computer, Electrical and Mathematical Sciences and Engineering Division, \textsc{King Abdullah University of Science and Technology}, Saudi Arabia} 
\affil[4]{Applied Statistics Unit, \textsc{Indian Statistical Institute}, Kolkata, India}
\affil[5]{Department of Mathematics and Statistics, \textsc{Indian Institute of Technology Kanpur}, India}

\date{\vspace{-5ex}}

\begin{document}
\maketitle

\makeatletter\def\Hy@Warning#1{}\makeatother 


\begin{abstract}
Classification of high-dimensional low sample size (HDLSS) data poses a challenge in a variety of real-world situations, such as gene expression studies, cancer research, and medical imaging. This article presents the development and analysis of some classifiers that are specifically designed for HDLSS data. These classifiers are free of tuning parameters and are robust, in the sense that they are devoid of any moment conditions of the underlying data distributions. It is shown that they yield perfect classification in the HDLSS asymptotic regime, under some fairly general conditions. The comparative performance of the proposed classifiers is also investigated. Our theoretical results are supported by extensive simulation studies and real data analysis, which demonstrate promising advantages of the proposed classification techniques over several widely recognized methods.
\end{abstract}

\vspace{-0.4cm}
\begin{center}
\end{center}


\section{Introduction}
\label{introduction}

High-dimensional low sample size data is characterized by having a large number of features or variables, but only a few samples or observations. The problem of HDLSS classification has been an important problem in the statistics and machine learning communities. In today’s world, high-dimensional low sample size problems are frequently encountered in scientific areas including  microarray gene expression studies, medical image analysis, and spectral measurements in chemometrics to name a few.
\blfootnote{A version of this work was published at the \textsl{European Conference on Machine Learning and Principles and Practice of Knowledge Discovery in Databases} (ECML PKDD), 2023. \textsc{DOI:} \textsf{\MYhref[purple]{https://doi.org/10.1007/978-3-031-43424-2_6}{https://doi.org/10.1007/978-3-031-43424-2\_6}}. All relevant codes and implementation details are available at \textsf{\MYhref[blue]{https://github.com/jyotishkarc/sub-834-ecml-2023}{https://github.com/jyotishkarc/sub-834-ecml-2023}}.}

Traditional classification techniques such as logistic regression, support vector machines (SVM) \citep{cortes1995support}, and $k$-nearest neighbors (k-NN) \citep{knn-paper-2,knn-paper-1} often fail on this type of data \citep{PESTOV20131427} when certain regularity conditions on the underlying distributions are not met. In case of $k$-nearest neighbors, for example, when the dimension of the data is far greater than the number of observations, the concept of neighbors becomes loose and ill-defined. Consequently, the $k$-nearest neighbor classifier exhibits erratic behavior \citep{beyer-et-al}. Due to distance concentration, Euclidean distance (ED)-based classifiers suffer certain limitations in HDLSS situations \citep{aggarwal2001surprising,franccois2007concentration}. Some recent work has studied the effect of distance concentration on some widely used classifiers based on Euclidean distances, such as 1-nearest neighbor (1-NN) classifier \citep{hastie2009elements}, SVM, etc. They derived conditions under which these classifiers yield perfect classification in the HDLSS setup \citep{hall2005geometric}. Moreover, ED-based classifiers lack robustness to outliers, since ED is sensitive to outliers.

For the HDLSS setup, numerous studies adopt dimension reduction approach as a pre-processing step before performing the classification. These work include modern classifiers and learning techniques centered mainly on feature selection (e.g., correlation-based, information theory-based, feature clustering \citep{song2011fast}, etc.), projection based on transformation \citep{abdi2010principal,hyvarinen2000independent}, regularization (ridge, LASSO, SCAD, and Elastic-net \citep{zou2005regularization}), deep learning (autoencoders \citep{hinton2006reducing,wang2014generalized}), etc. However, this is not optimal when the dimension reduction step is conducted independently of the goals of finding reduced features that maximize the separation between classes of signals. In fact, it is inevitable that some information is lost via dimension reduction if a large number of features turn out to be relevant and weakly dependent upon each other. A few studies have conducted classification of HDLSS data without employing dimension reduction (see, e.g., \cite{roy2022generalizations,shen2022classification,yin2020population}). 

Energy distance was introduced in \cite{baringhaus2004new,szekely-2004} as a statistical measure of distance between two probability distributions on $\mathbb{R}^d$. It was primarily designed with a goal of testing for equality of two or more multivariate distributions, and worked particularly well with high-dimensional data. Recently, energy distances have been utilized in the context of classification (see, e.g., \cite{pmlr-v151-roy22a,roy2022generalizations,roy2022exact}) as well. In this article, we develop classifiers based on a more general version of energy distance that yield asymptotically perfect classification (i.e., zero misclassification rate) under fairly general assumptions in an HDLSS setting, without maneuvering dimension reduction.

Suppose $\mathbf{X}$ and $\mathbf{Y}$ are two $d$-variate random vectors following the distribution functions $\mathbf{F}$ and $\mathbf{G}$, respectively. In the context of testing for equality of $\mathbf{F}$ and $\mathbf{G}$, a constant multiple of the following was introduced in \cite{10.1214/19-AOS1936} as squared multivariate Cram\'{e}r-von Mises (CvM) distance between $\mathbf{F}$ and $\mathbf{G}$. It is a special case of the generalized energy distance \citep{generalized-energy-dist}:
\begin{equation}
\label{eq:energy}
\mathcal{W}^{*}_{\mathbf{FG}}=2\int\int\left(\mathbf{F}_\beta(t) - \mathbf{G}_\beta(t)\right)^{2} \,d\mathbf{H}(\boldsymbol{\beta}, t) ~,
\end{equation}
where $\boldsymbol{\beta} \in \mathbb{R}^d$, and $\mathbf{F}_\beta(t) = \mathbb{P}\left[\boldsymbol{\beta}^{\top} \mathbf{X} \leq t\right]$ and $\mathbf{G}_\beta(t) = \mathbb{P}\left[\boldsymbol{\beta}^{\top} \mathbf{Y} \leq t\right]$, for $t \in \mathbb{R}$, are the cumulative distribution functions of $\beta^{\top}\mathbf{X}$ and $\beta^{\top}\mathbf{Y}$ respectively, evaluated at $t$, $\,d\mathbf{H}(\boldsymbol{\beta}, t) = \,d \mathbf{H}_{\boldsymbol{\beta}}(t) \,d \lambda(\boldsymbol{\beta})$ with $\lambda(\boldsymbol{\beta})$ being the uniform probability measure on $d$-dimensional unit sphere $\mathbb{S}^{d-1} = \{\mathbf{x} \in \mathbb{R}^d : \mathbf{x}^{\top}\mathbf{x}=1\}$, and $\mathbf{H}_{\boldsymbol{\beta}}(t)=\alpha\mathbf{F}_\beta(t) + (1-\alpha)\mathbf{G}_\beta(t)$. Here, $\alpha$ is a fixed value in $(0,1)$. In the same context of hypothesis testing, \cite{pmlr-v119-li20s} considered a constant multiple of \eqref{eq:energy}, with $\mathbf{H}(\boldsymbol{\beta}, t)$ as the distribution function of a $(d+1)$-dimensional normal random vector with mean $\mathbf{0}_{d+1}$, the $(d+1)$-dimensional zero vector, and covariance matrix $I_{d+1}$, the identity matrix of order $(d+1)$. However, considering such a fixed distribution which is not data-dependent may not be useful in general. On the other hand, the weight function $\mathbf{H}_\beta$ considered in \cite{10.1214/19-AOS1936} is more flexible, since it adapts according to the underlying class distributions. In that sense, $\mathcal{ W}^{*}_{\mathbf{FG}}$ is referred to as a data-adaptive energy distance between $\mathbf{F}$ and $\mathbf{G}$. It was shown in \cite{10.1214/19-AOS1936} that $\mathcal{W}^{*}_{\mathbf{FG}}=0 \text{ if and only if } \mathbf{F}=\mathbf{G}$. This property of $\mathcal{W}^{*}_{\mathbf{FG}}$ says that it has the capability of discriminating between two different distributions. This motivates us to utilize $\mathcal{W}^{*}_{\mathbf{FG}}$ in the context of binary classification problems.

\subsection{Our contributions}
\label{contri}
In this article, we start off by developing a classifier based on $\mathcal{W}^{*}_{\mathbf{FG}}$. However, it suffers certain limitations in the HDLSS setting. We investigate and address those issues by modifying $\mathcal{W}^{*}_{\mathbf{FG}}$ in different ways, and based on the new measures of distance, we develop classifiers that are robust in the sense that their performance does not depend on the existence of the moments of the underlying class distributions. Moreover, the proposed classifiers are free from tuning parameters and admit strong theoretical guarantees under fairly general assumptions, in an HDLSS setup.
 
The rest of the paper is organized as follows. In \Cref{Methodolgy}, we develop a classifier based on $\mathcal{W}^{*}_{\mathbf{FG}}$, discuss its limitations and modify it to obtain three robust classifiers to achieve asymptotically perfect classification under milder conditions. \Cref{Asymptotics} provides an analysis of the asymptotic behaviors and a relative comparison of the proposed classifiers. \Cref{Empirical} demonstrates convincing advantages of the proposed classifiers using numerical simulations and real data analysis. Proofs of all the theoretical results are included in Section \ref{AppendixA} of the Supplementary Material. Lastly, Section \ref{AppendixB} of the Supplementary Material contains some additional details on the simulation studies and real data analysis. The supplementary material and the relevant R codes for simulation studies and real data analysis are available at \textsf{\MYhref[blue]{https://github.com/jyotishkarc/sub-834-ecml-2023}{https://github.com/jyotishkarc/sub-834-ecml-2023}}.


\section{Methodology}

\label{Methodolgy}
Consider two mutually independent samples
\begin{eqnarray*}
    \m{X}_{1}^{(d)}, \m{X}_{2}^{(d)}, \ldots, \m{X}_{m}^{(d)} &\stackrel{\text{i.i.d.}}{\sim} \mathbf{F}_d~~\text{ and }~~
    \m{Y}_{1}^{(d)}, \m{Y}_{2}^{(d)}, \ldots, \m{Y}_{n}^{(d)} &\stackrel{\text{i.i.d} .}{\sim} \mathbf{G}_d
\end{eqnarray*}
where $\m{X}_i^{(d)} = (X_{i1}, X_{i2}, \ldots, X_{id})^{\top}$ for $i=1,2,\ldots,m,$ and $\m{Y}_j^{(d)}= (Y_{j1}, Y_{j2}, \ldots,$ $Y_{jd})^{\top}$ for $j=1,\ldots, n$ are $d$-dimensional random vectors arising from two different population distributions $\m{F}_d$ and $\m{G}_d$. For the sake of convenience, we shall drop $d$ from notations where dependence on $d$ is obvious. As mentioned in \Cref{contri}, we shall keep the sample sizes $m$ and $n$ fixed throughout our analysis.

The angular distance between any $\mathbf{u},\mathbf{v} \in \mathbb{R}^d$ was defined in \cite{10.1214/19-AOS1936} as follows:
\begin{eqnarray}
\label{eq:rho}
\rho\left(\mathbf{u},\mathbf{v}\right)=\mathbb{E}\left[\rho_0\left(\mathbf{u},\mathbf{v} ; \mathbf{Q}\right)\right]~ \text{with}~\mathbf{Q} \sim \alpha \mathbf{F}+(1-\alpha) \mathbf{G}.\\
\label{eq:rho_0}
 \rho_0\left(\mathbf{u},\mathbf{v} ; \mathbf{w}\right) =
  \begin{cases}
  \dfrac{1}{\pi} \angle(\mathbf{u}-\mathbf{w}, \mathbf{v}-\mathbf{w}) &\text{ if } \mathbf{u}\neq \mathbf{w} \text{ and } \mathbf{v}\neq \mathbf{w},\\
  0 &\text{ otherwise,}
  \end{cases}
\end{eqnarray}
  with $\alpha = \frac{m}{m+n}~,$ and $\angle(\m a, \m b)=\text{cos}^{-1}\left(\frac{ \m a^{\top} \m b}{\|\m a\|_2\|\m b\|_2}\right)$ with $\|\m v\|_2$ as the $l_2$ norm of $\m v$. Note that $\rho \in [0,1]$ since $\rho_0$ takes values in $[0,1]$. It was shown in \cite{10.1214/19-AOS1936} that $\mathcal{W}^{*}_{\mathbf{FG}}$, defined in \eqref{eq:energy}, has the following closed-form expression:
 \begin{equation}
    \mathcal{W}^{*}_{\mathbf{FG}}=\mathbb{E}\left[2 \rho\left(\mathbf{X}_1, \mathbf{Y}_1\right)-\rho\left(\mathbf{X}_1, \mathbf{X}_2\right)-\rho\left(\mathbf{Y}_1, \mathbf{Y}_2\right)\right].
    \label{eq:w*}
\end{equation}
 
\subsection{A classifier based on $\mathcal{W}^{*}_{\mathbf{FG}}$}
\label{prelims}
Let us consider the unknown expectations $t_{\mathbf{F F}}=\mathbb{E}\left[{{\rho}}\left(\mathbf{X}_{1}, \mathbf{X}_{2}\right)\right]$ and $t_{\mathbf{G G}}=\mathbb{E}\left[{{\rho}}\left(\mathbf{Y}_{1}, \mathbf{Y}_{2}\right)\right]$. We start off by defining estimators of $t_{\mathbf{FF}}$ and $t_\mathbf{GG}$ as follows.
$$\hat{t}_{\mathbf{F F}}=\frac{1}{m(m-1)} \sum_{i \neq j} \hat{{\rho}}\left(\mathbf{X}_{i},\mathbf{X}_{j}\right) ~\text{ and }~ \hat{t}_{\mathbf{G G}}=\frac{1}{n(n-1)} \sum_{i \neq j} {\hat{\rho}}\left(\mathbf{Y}_{i},\mathbf{Y}_{j}\right),$$
where $\hat{\rho}$ is defined as a sample version of $\rho$ in the following manner:
\begin{equation}
        \label{rho-hat}
       \hat{\rho}(\mathbf{u},\mathbf{v})=\frac{1}{m+n}\Big( \sum_{i=1}^m \rho_0(\mathbf{u},\mathbf{v},\mathbf{X}_{i})+\sum_{j=1}^n \rho_0(\mathbf{u},\mathbf{v},\mathbf{Y}_{j})\Big).
\end{equation}
Similarly, for $\mathbf{z} \in \mathbb{R}^d$, we define
\begin{equation}
\begin{array}{r@{}l}
\hat{t}_{\mathbf{F}}(\mathbf{z})=\frac{1}{m} \sum_{i} \hat{{\rho}}\left(\mathbf{X}_{i},\mathbf{z}\right)&,~\hat{t}_{\mathbf{G}}(\mathbf{z})=\frac{1}{n} \sum_{j} \hat{{\rho}}\left(\mathbf{Y}_{j}, \mathbf{z}\right) , \\
l_{\mathbf{F}}(\mathbf{z})=\hat{t}_{\mathbf{F}}(\mathbf{z})-\frac{1}{2}\hat{t}_{\mathbf{F F}} &,~l_{\mathbf{G}}(\mathbf{z})=\hat{t}_{\mathbf{G}}(\mathbf{z})-\frac{1}{2}\hat{t}_{\mathbf{G G}} .
\end{array}
\end{equation}
Finally, the classifier $\delta_0$ is defined as follows:
\begin{equation}
\label{eq:0}
\delta_{0}(\mathbf{z})=\left\{\begin{array}{l}
1~ \text { if } l_{\mathbf{G}}(\mathbf{z})-l_{\mathbf{F}}(\mathbf{z})>0, \\
2~ \text { otherwise,}
\end{array}\right.
\end{equation}
where $\delta_{0}(\mathbf{z}) = 1$ or $\delta_{0}(\mathbf{z}) = 2$ correspond to assigning $\mathbf{z}$ to class $1$ or class $2$ having data distribution $\mathbf{F}$ or $\mathbf{G}$, respectively. Let $\bm{\mu}_\mathbf{F}$ and $\bm{\mu}_\mathbf{G}$ denote the mean vectors for $\mathbf{F}$ and $\mathbf{G}$, respectively, and $\bm{\Sigma}_\mathbf{F}$ and $\bm{\Sigma}_\mathbf{G}$ denote the covariance matrices for $\mathbf{F}$ and $\mathbf{G}$, respectively.
In order to analyze the behavior of the $\delta_0$ in HDLSS setup, consider the following assumptions:
\begin{assumption}
\label{ass:0.1}
     There exists a constant $c < \infty$ such that $E[|U_k|^4] < c$ for all $1 \leq k \leq d$, where $\mathbf{U} =(U_1, \cdots , U_d)^{\top}$ follows either $\mathbf{F}$ or $\mathbf{G}$.
\end{assumption}

\begin{assumption}
\label{ass:0.2}
 $\displaystyle\lambda_{\mathbf{FG}}=\lim_{d \to \infty} \Big\{\frac{1}{d}\|\bm{\mu}_\mathbf{F}-\bm{\mu}_\mathbf{G}\|^2\Big\}$ and $\sigma^2_I=\displaystyle\lim_{d \to \infty} \Big\{\frac{1}{d}\operatorname{trace}(\bm{\Sigma}_I)\Big\}$ exist for $I\in\{\mathbf{F},\mathbf{G}\}$. 
\end{assumption}
\begin{assumption}
\label{ass:0.3}
    Let $\mathbf{U}$, $\mathbf{V}$ and $\mathbf{Z}$ be independent random vectors such that each of them follows either $\mathbf{F}$ or $\mathbf{G}$. Then, 
    $$\sum_{i<j}\operatorname{cov}(({U}_i-{Z}_i)({V}_i-{Z}_i),({U}_j-{Z}_j)({V}_j-{Z}_j))=o(d^2) .$$
\end{assumption}
\Cref{ass:0.1} requires finiteness of the fourth moments of all marginals of $\mathbf{F}$ and $\m G$. \Cref{ass:0.2} demands the existence of the limiting values of the average of the squared mean difference between the marginals of two distributions and the variances of the marginals to exist.
\Cref{ass:0.3} is trivially satisfied when the component variables of the underlying populations are independent. It also holds with certain additional constraints on their dependence structure, e.g., when the sequence
$\{({U}_k-{Z}_k)({V}_k-{Z}_k)\}_{k\geq 1}$ has $\rho$-mixing property. In fact, if the sequences $\{U_k\}_{k\geq 1}$, $\{V_k\}_{k\geq 1}$ and $\{Z_k\}_{k\geq 1}$ all have $\rho$-mixing property, then the sequence
$\{h(U_k,V_k,Z_{k})\}_{k\geq 1}$ also has $\rho$-mixing property, for any Borel measurable function $h$ (see \cite{bradley2007} for more details).
\begin{theorem}
    \label{thm:d0}
    Suppose \cref{ass:0.1,ass:0.2,ass:0.3} are satisfied. Then, $\theta^{*}_{\mathbf{FG}}=\lim_{d \to \infty} \mathcal{W}^{*}_{\mathbf{FG}}$ is finite, and for a test observation $\mathbf{Z} ,$
\begin{itemize}
    \item[(a)] if $\mathbf{Z} \sim \mathbf{F}$, then 
    $l_{\mathbf{G}}(\mathbf{Z})-l_{\mathbf{F}}(\mathbf{Z})\stackrel{\mathbb{P}}{\rightarrow}\frac{1}{2}\theta^{*}_{\mathbf{FG}}  ~ \text{ as } d \to\infty;$

    \item[(b)] if $\mathbf{Z} \sim \mathbf{G}$, then 
    $l_{\mathbf{G}}(\mathbf{Z})-l_{\mathbf{F}}(\mathbf{Z})\stackrel{\mathbb{P}}{\rightarrow}-\frac{1}{2}\theta^{*}_{\mathbf{FG}} ~ \text{ as } d \to\infty.$
\end{itemize}
\end{theorem}
As $d \to \infty$, $l_{\mathbf{G}}(\mathbf{Z})-l_{\mathbf{F}}(\mathbf{Z})$ converges in probability to the limit of $\frac{1}{2} \mathcal{W}^{*}_{\mathbf{FG}}$ if $\mathbf{Z}\sim \mathbf{F}$, and to the negative of it, if $\mathbf{Z}\sim \mathbf{G}$. This justifies the construction of the classifier $\delta_0$ in \eqref{eq:0}. The probability of misclassification of a classifier $\delta$ is defined as $\Delta = \alpha P[\delta(\mathbf{Z}) = 2|\mathbf{Z} \sim \mathbf{F}] +(1-\alpha)P[\delta(\mathbf{Z}) = 1|\mathbf{Z} \sim \mathbf{G}].$ Now, we state a result on the convergence of the misclassification probabilities of the classifier $\delta_0$ (denoted by $\Delta_0$), under the set of assumptions stated above.
\begin{theorem}
\label{thm:0}
Suppose that \cref{ass:0.1,ass:0.2,ass:0.3} are satisfied, and either
   $\displaystyle\lambda_\mathbf{FG}\neq0~$ or $\sigma^2_\mathbf{F}\neq\sigma^2_\mathbf{G}~$ holds. Then, $\Delta_0 \xrightarrow{} 0$ as $d \xrightarrow{} \infty.$
\end{theorem} 
It follows from \Cref{thm:0} that if $\mathbf{F}$ and $\mathbf{G}$ differ in their locations and/or scales, then $\Delta_0$ converges to $0$ as the dimension grows. Clearly, the asymptotic properties of the classifier $\delta_0$ are governed by the limiting constants, $\lambda_\mathbf{FG}, \sigma_\mathbf{F}$, and $\sigma_\mathbf{G}$. Similar issues regarding assumptions on the existence of moments of class distributions were also present in the two-sample test based on $\mathcal{W}^{*}_{\mathbf{FG}}$ in \cite{10.1214/19-AOS1936}. Let us now consider the following two examples:
\begin{itemize}
    \item[] \textbf{Example 1:} $X_{1k} \stackrel{\text{i.i.d.}}{\sim} N(1,1)$ and $Y_{1k} \stackrel{\text{i.i.d.}}{\sim} N(1,2)$
    \item[] \textbf{Example 2:} $X_{1k} \stackrel{\text{i.i.d.}}{\sim} N(0,3)$ and $Y_{1k} \stackrel{\text{i.i.d.}}{\sim} t_3$
\end{itemize}
for $1\leq k \leq d$. Here, $N(\mu,\sigma^2)$ refers to a Gaussian distribution with mean $\mu$ and variance $\sigma^2$, and $t_s$ denotes the Student's t-distribution with $s$ degrees of freedom.

\begin{figure}[H]
    \centering
    \includegraphics[width=0.70\textwidth]{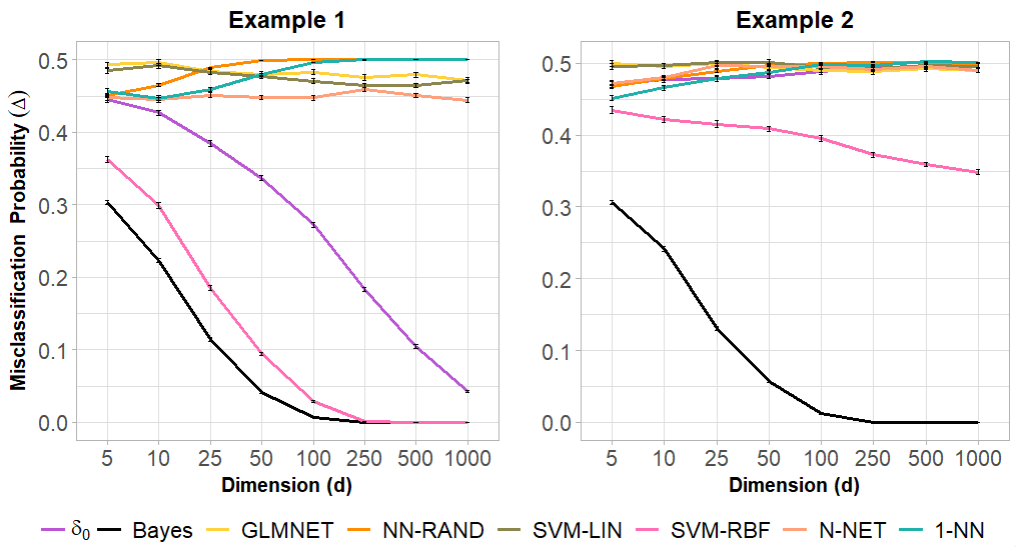}
    \caption{Average misclassification rates with errorbars for $\delta_0$, along with some popular classifiers for increasing dimensions. Bayes classifier is treated as a benchmark.}
    \label{fig:delta-0-example}
\end{figure}

In \textbf{Example 1}, $\|\bm{\mu}_\mathbf{F}-\bm{\mu}_\mathbf{G}\|^2=0$ but $\sigma^2_\mathbf{F}=1, \sigma^2_\mathbf{G}=2$. It can be observed from \Cref{fig:delta-0-example} that $\delta_0$ identifies this difference in scale and its performance improves as $d$ increases, whereas most of the popular classifiers misclassify nearly $45\%$ of the observations.

In \textbf{Example 2}, $\|\bm{\mu}_\mathbf{F}-\bm{\mu}_\mathbf{G}\|^2=0$ and $\sigma^2_\mathbf{F}=\sigma^2_\mathbf{G}=3$, i.e.,
there is no difference between either location parameters or scale parameters. Consequently, $\delta_0$ (as well as the popular classifiers) fails to classify the test observations correctly since the assumptions in \Cref{thm:0} are not met.

These simulations support  \Cref{thm:0} (see \Cref{fig:delta-0-example}) and illustrate the limitations of the classifier $\delta_0$ when there is no difference in either location parameters or scale parameters.
In the next subsection, we refine $\delta_0$ to develop some classifiers whose asymptotic properties are free of any moment conditions, and the limiting constants $\lambda_{\mathbf{FG}}, \sigma_{\mathbf{F}}^2$, and $\sigma_{\mathbf{G}}^2$, mentioned in \cref{ass:0.1,ass:0.2}.

\subsection{Refinements of $\delta_0$}
 \label{refinements-of-delta0}
\subsubsection{A New Measure of Distance}

We modify $\mathcal{W}^{*}_{\mathbf{FG}}$ by taking the average of the distances between each $F_{k}$ and $G_{k}$, the $k$-th marginals of $\mathbf{F}$ and $\mathbf{G}$, respectively. Define
${\overline{\mathcal{W}}}^{*}_\mathbf{FG}=\frac{1}{d} \sum_{k=1}^{d} \mathcal{W}^{*}_{F_{k}G_{k}}$. For each $1\leq k \leq d$, it follows from \eqref{eq:w*} the quantity $\mathcal{W}^{*}_{F_{k} G_{k}}$ has the following closed-form expression:
\begin{equation}
    \mathcal{W}^{*}_{F_{k} G_{k}}=\mathbb{E}\left[2 \rho\left(X_{1 k}, Y_{1 k}\right)-\rho\left(X_{1 k}, X_{2 k}\right)-\rho\left(Y_{1 k}, Y_{2 k}\right)\right],
\end{equation}
where $\mathbf{X}_{1}, \mathbf{X}_{2} \stackrel{i.i.d.}{\sim} \mathbf{F}$ and $\mathbf{Y}_{1}, \mathbf{Y}_{2} \stackrel{i.i.d.}{\sim} \mathbf{G}.$

Recall the definition of $\rho$ in \eqref{eq:rho}. For any two $d$-dimensional random variables $\mathbf{u} = (u_1,u_2,\ldots,u_d)^{\top}$ and $\mathbf{v}=(v_1,v_2,\ldots,v_d)^{\top}$, we define
\begin{equation}
    \bar{\rho}(\mathbf{u}, \mathbf{v})=\frac{1}{d} \sum_{k=1}^{d} \rho\left(u_{k}, v_{k}\right) .
\end{equation}
We now introduce some notations:
 $$T_{\mathbf{F G}}=\mathbb{E}\left[\bar{\rho}\left(\mathbf{X}_{1}, \mathbf{Y}_{1}\right)\right],~ T_{\mathbf{F F}}=\mathbb{E}\left[\bar{\rho}\left(\mathbf{X}_{1}, \mathbf{X}_{2}\right)\right], \text{~and~}  T_{\mathbf{G G}}=\mathbb{E}\left[\bar{\rho}\left(\mathbf{Y}_{1}, \mathbf{Y}_{2}\right)\right].$$ This implies that ${\overline{\mathcal{W}}}^{*}_\mathbf{FG}=2 T_{\mathbf{F G}}-T_{\mathbf{F F}}-T_{\mathbf{G G}} .$ Note that $\mathcal{W}^{*}_{F_{k} G_{k}}\geq 0$, and equality holds iff $F_k = G_k$. Thus, ${\overline{\mathcal{W}}}^{*}_\mathbf{FG}=0$ iff $F_k = G_k$ for all $1 \leq k \leq d$. This property of ${\overline{\mathcal{W}}}^{*}_\mathbf{FG}$ suggests that it can be utilized as a measure of separation between $\mathbf{F}$ and $\mathbf{G}$. Since $T_{\mathbf{F F}}, T_{\mathbf{G G}}$, and $T_{\mathbf{F G}}$ are all unknown quantities, we consider the following estimators based on the sample observations:
\begin{align*}
 \hat{T}_{\mathbf{F F}} = \frac{1}{m(m-1)} \sum_{i \neq j} \hat{\bar{\rho}}\left(\mathbf{X}_{i},\mathbf{X}_{j}\right),~~ \hat{T}_{\mathbf{G G}} = \frac{1}{n(n-1)} \sum_{i \neq j} \hat{\bar{\rho}}\left(\mathbf{Y}_{i}, \mathbf{Y}_{j}\right),~~ \hat{T}_{\mathbf{F G}} = \frac{1}{mn}\displaystyle \sum_{i , j} \hat{\bar{\rho}}\left(\mathbf{X}_{i}, \mathbf{Y}_{j}\right),
\end{align*}
where $\hat{\bar{\rho}}(\mathbf{u}, \mathbf{v})$ is a natural estimator of $\bar{\rho}(\mathbf{u}, \mathbf{v})$, defined as follows:
\begin{equation}
    \hat{\bar{\rho}}(\mathbf{u}, \mathbf{v})=\frac{1}{d} \sum_{k=1}^{d} \hat\rho\left(u_{k}, v_{k}\right).
\end{equation}
This leads to an empirical version of ${\overline{\mathcal{W}}}^{*}_\mathbf{FG}$ defined as
\begin{equation}
    \hat{\overline{\mathcal{W}}}^{*}_\mathbf{FG} = 2\hat{T}_{\mathbf{F G}} - \hat{T}_{\mathbf{F F}} - \hat{T}_{\mathbf{G G}}.
\end{equation}
For $\mathbf{z}=(z_1,z_2,\ldots,z_d)^{\top} \in \mathbb{R}^d$, we define:
  \begin{eqnarray*}
&\hat{T}_{\mathbf{F}}(\mathbf{z})=\frac{1}{m} \sum_{i} \hat{\bar{\rho}}\left(\mathbf{X}_{i}, z_i\right)~,~\hat{T}_{\mathbf{G}}(\mathbf{z})=\frac{1}{n} \sum_{j} \hat{\bar{\rho}}\left(\mathbf{Y}_{j}, z_j\right),\\
  &L_{\mathbf{F}}(\mathbf{z})=\hat{T}_{\mathbf{F}}(\mathbf{z})-\frac{1}{2}\hat{T}_{\mathbf{F F}}~,~L_{\mathbf{G}}(\mathbf{z})=\hat{T}_{\mathbf{G}}(\mathbf{z})-\frac{1}{2}\hat{T}_{\mathbf{G G}},\\
  &S(\mathbf{z})=\hat{T}_{\mathbf{F}}(\mathbf{z})+\hat{T}_{\mathbf{G}}(\mathbf{z})-\frac{1}{2}\left(\hat{T}_{\mathbf{F F}}+\hat{T}_{\mathbf{G G}}\right)-\hat{T}_{\mathbf{F G}}.
  \end{eqnarray*}

\subsubsection{Classifier Based on $\overline{\mathcal{W}}^{*}_{\mathbf{F G}}$}

 Define $\mathscr{D}_1(\mathbf{z}) =L_{\mathbf{G}}(\mathbf{z})-L_{\mathbf{F}}(\mathbf{z})$. We prove that $\mathscr{D}_1(\mathbf{Z})$ converges in probability to $\frac{1}{2}{\overline{\mathcal{W}}}^{*}_\mathbf{FG}$, if $\mathbf{Z}\sim \mathbf{F}$ and to $-\frac{1}{2}{\overline{\mathcal{W}}}^{*}_\mathbf{FG}$,
 if $\mathbf{Z}\sim \mathbf{G}$, as $d \to \infty$ (see \Cref{thm:1}). This, along with the fact that the average energy distance, ${\overline{\mathcal{W}}}^{*}_\mathbf{FG}$ is non-negative, motivates us to consider the following~classifier:
\begin{eqnarray}
    \delta_{1}(\mathbf{z})=\left\{\begin{array}{l}
1~ \text { if } \mathscr{D}_1(\mathbf{z}) > 0,\\
2~ \text { otherwise. }
\end{array}\right. \label{eq:1}
\end{eqnarray}
Recall that $(T_{\mathbf{F G}}-T_{\mathbf{F F}})$ and $(T_{\mathbf{F G}}-T_{\mathbf{G G}})$ sum up to $\overline{\mathcal{W}}_{\mathbf{FG}}^{*}$. So, in case $T_{\mathbf{F G}}$ lies between $T_{\mathbf{F F}}$ and $T_{\mathbf{G G}}$, adding them up might nearly cancel each other out, resulting in a very small value of $\overline{\mathcal{W}}_{\mathbf{FG}}^{*}$. Consequently, it may not fully capture the actual dissimilarity between $\mathbf{F}$ and $\mathbf{G}$. A natural way to address this problem is to square the two quantities before adding them. We define
\begin{equation*}
    \bar{\tau}_{\mathbf{FG}}=\left(T_{\mathbf{F G}}-T_{\mathbf{F F}}\right)^{2}+\left(T_{\mathbf{F G}}-T_{\mathbf{G G}}\right)^{2} .
\end{equation*}
It follows from simple calculations that one may write $\bar{\tau}_{\mathbf{FG}}$ in the following form:
\begin{equation}
\label{eq:tau}
    \bar{\tau}_{\mathbf{FG}}=\frac{1}{2}\overline{\mathcal{W}}^{*2}_\mathbf{FG}+\frac{1}{2}\left(T_{\mathbf{F F}}-T_{\mathbf{G G}}\right)^{2}.
\end{equation} 
Note that $\bar{\tau}_{\mathbf{FG}}$ being a convex combination of squares of $\overline{\mathcal{W}}^{*}_{\mathbf{FG}} = 2T_{\mathbf{F G}}-T_{\mathbf{F F}}-T_{\mathbf{G G}}$ and $T_{\mathbf{F F}}-T_{\mathbf{G G}}$, both of which are measures of disparity between $\mathbf{F}$ and $\mathbf{G}$, can be considered as a new measure of disparity between the two distributions. The modification approach proposed in \eqref{eq:tau} is similar to what had been suggested in the literature of two-sample hypothesis tests to improve the power of some two-sample tests for HDLSS data in \cite{biswas2014nonparametric}.

\subsubsection{Classifier Based on $\bar{\tau}_{\mathbf{F G}}$}

We now develop a classifier that utilizes $\bar{\tau}_{\mathbf{F G}}$. Recall the definitions of $\mathscr{D}_1(\mathbf{z})$ and $S(\mathbf{z})$.
For $\mathbf{z} \in \mathbb{R}^d$, define 
\begin{equation*}
    \mathscr{D}_2(\mathbf{z}) = \frac{1}{2} \hat{\overline{\mathcal{W}}}^{*}_\mathbf{FG}\cdot \mathscr{D}_1(\mathbf{z}) + \frac{1}{2} \left(\hat T_{\mathbf{F F}}-\hat T_{\mathbf{G G}}\right) \cdot S(\mathbf{z}).
\end{equation*}
We show that as $d \to \infty$, $\mathscr{D}_2(\mathbf{Z})$ converges in probability to $\bar{\tau}_{\mathbf{FG}}~(>0)$ if $\mathbf{Z}\sim \mathbf{F}$, and to $-\bar{\tau}_{\mathbf{FG}}~(<0)$ if $\mathbf{Z}\sim \mathbf{G}$ (see \Cref{thm:1} in \Cref{Asymptotics} below). This motivates us to consider the following classifier:
\begin{eqnarray}
\label{eq:2}
    \delta_{2}(\mathbf{z})=\left\{\begin{array}{l}
1~ \text { if } \mathscr{D}_2(\mathbf{z}) > 0,\\
2~ \text { otherwise.}
\end{array}\right.
\end{eqnarray}
We consider another measure of disparity between $\mathbf{F}$ and $\mathbf{G}$ (say, $\bar{\psi}_{\mathbf{FG}}$), by simply replacing the squares of $\overline{\mathcal{W}}^{*}_\mathbf{FG}$ and $S_{\mathbf{F G}}$  by their absolute values in the expression for $\bar{\tau}_{\mathbf{FG}}$. A similar modification has already been considered, in the context of two-sample testing (see, e.g., \cite{Tsukada2019}). Based on this, we define yet another measure of separation:
$$\bar{\psi}_{\mathbf{FG}}=\frac{1}{2}\overline{\mathcal{W}}^{*}_\mathbf{FG}+\frac{1}{2}\left| T_{\mathbf{F F}}-T_{\mathbf{G G}}\right| .$$

\subsubsection{Classifier Based on $\bar{\psi}_{\mathbf{FG}}$}

For $\mathbf{z} \in \mathbb{R}^d$, we define 
\begin{equation*}
    \begin{split}
        \mathscr{D}_3(\mathbf{z}) = \frac{1}{2} \hat{\overline{\mathcal{W}}}^{*}_{\mathbf{FG}} \operatorname{sign}\left(\mathscr{D}_1(\mathbf{z})\right)
        +\frac{1}{2} \left(\hat T_{\mathbf{F F}}-\hat T_{\mathbf{G G}}\right)\cdot \operatorname{sign}\left(S(\mathbf{z})\right)
    \end{split}
  \end{equation*}
where $\operatorname{sign}(\cdot)$ is defined as $\operatorname{sign}(x) = \frac{x}{|x|}$ for $x \neq 0$, and $0$ for $x = 0$.

We prove that as $d \to \infty$, $\mathscr{D}_3(\mathbf{Z})$ converges in probability to $\bar{\psi}_{\mathbf{FG}}$, a positive quantity if $\mathbf{Z}\sim \mathbf{F}$, and to $-\bar{\psi}_{\mathbf{FG}}$, a negative quantity if $\mathbf{Z}\sim \mathbf{G}$ (see \Cref{thm:3} in \Cref{Asymptotics} below). This motivates us to construct the following classifier:
\begin{eqnarray}
\label{eq:3}
    \delta_{3}(\mathbf{z})=\left\{\begin{array}{l}
1~ \text { if } \mathscr{D}_3(\mathbf{z}) > 0,\\
2~ \text { otherwise.}
\end{array}\right.
\end{eqnarray}


\section{Asymptotics under HDLSS Regime}
\label{Asymptotics}
Suppose  $\m U = (U_1,U_2,\ldots,U_d)^{\top} $ and  $\m V =(V_1,V_2,\ldots, V_d)^{\top}$ are drawn independently from $\m F$ or $\m G$. We assume that the component variables are weakly dependent. In particular, we assume the following.
\begin{assumption}
\label{ass:1}    
For any four $d$-dimensional random vectors $\mathbf{U}, \mathbf{V}, \mathbf{Q}, \mathbf{Q^*}$ having distribution $\m F$ or $\m G$, such that they are mutually independent,
\begin{itemize}
    \item[i.] $\sum_{1\leq k_1<k_2 \leq d}\operatorname{cov}(\rho_0(U_{k_1},V_{k_1};Q_{k_1}), \rho_0(U_{k_2},V_{k_2};Q^{}_{k_2}))=o(d^2);$
    \item[ii.] $ \sum_{1 \leq k_1<k_2\leq d}\operatorname{cov}(\rho_0(U_{k_1},V_{k_1};Q_{k_1}), \rho_0(U_{k_2},V_{k_2};Q^{*}_{k_2}))=o(d^2).$
\end{itemize}
\end{assumption} 
\Cref{ass:1} is trivially satisfied if the component variables of the underlying distributions are independently distributed, and it continues to hold when the components have $\rho$-mixing property.

\begin{theorem}
    \label{thm:1}
    Suppose  \cref{ass:1} is satisfied. For a test observation $\mathbf{Z} ,$ 
    \begin{itemize}
    \item[(a)] if $\mathbf{Z} \sim \mathbf{F}$, then $
\left|\mathscr{D}_1(\mathbf{Z})-\frac{1}{2} {\overline{\mathcal{W}}}^{*}_\mathbf{FG}\right| \stackrel{\mathbb{P}}{\rightarrow} 0 \text{  and }\left|\mathscr{D}_2(\mathbf{Z})-\bar{\tau}_{\mathbf{FG}}\right| \stackrel{\mathbb{P}}{\rightarrow} 0 \text{ as } d \to\infty;$
    \vspace{0.05cm}
    \item[(b)] if $\mathbf{Z} \sim \mathbf{G}$, then $
\left|\mathscr{D}_1(\mathbf{Z})+\frac{1}{2} {\overline{\mathcal{W}}}^{*}_\mathbf{FG}\right| \stackrel{\mathbb{P}}{\rightarrow} 0 \text{ and }\left|\mathscr{D}_2(\mathbf{Z})+\bar{\tau}_{\mathbf{FG}}\right| \stackrel{\mathbb{P}}{\rightarrow} 0 \text{ as } d \to\infty.$
\end{itemize}
\end{theorem}

\Cref{thm:1} states that if $\mathbf{Z} \sim \mathbf{F}$ (respectively, $\mathbf{Z} \sim \mathbf{G}$), the discriminants corresponding to
$\delta_1$ and $\delta_2$ converge in probability to a positive (respectively, negative) quantity as $d\to \infty$. This justifies our construction of the classifiers $\delta_1$ and $\delta_2$ in \eqref{eq:1} and \eqref{eq:2}, respectively.

 Now, we expect $\delta_1$ to yield an optimal performance if ${\overline{\mathcal{W}}}^{*}_\mathbf{FG}$ does not vanish with increasing dimensions. Hence, it is reasonable to assume the following:
\begin{assumption}
    \label{ass:2} $$\liminf_{d\to\infty} {\overline{\mathcal{W}}}^{*}_\mathbf{FG} > 0.$$
\end{assumption}
\Cref{ass:2} implies that the separation between $\mathbf{F}$ and $\mathbf{G}$ is asymptotically non-negligible. Observe that this assumption is satisfied if the component variables of $\mathbf{F}$ and $\mathbf{G}$ are identically distributed.
We also need the following assumption for the asymptotic analysis of $\delta_3$.
\begin{assumption}
    \label{ass:3}
    $$\liminf_{d\to\infty} \left|T_{\mathbf{F F}}-T_{\mathbf{G G}}\right|>0.$$
\end{assumption}

\begin{theorem}
    \label{thm:3}
    Suppose \cref{ass:1,ass:2,ass:3} hold true. For a test observation $\mathbf{Z}$,
\begin{itemize}
    \item[(a)] if $\mathbf{Z} \sim \mathbf{F}$, then 
    $\left|\mathscr{D}_3(\mathbf{Z})-\bar{\psi}_{\mathbf{FG}}\right| \stackrel{\mathbb{P}}{\rightarrow} 0 \text{ as } d \to\infty;$

    \item[(b)] if $\mathbf{Z} \sim \mathbf{G}$, then $\left|\mathscr{D}_3(\mathbf{Z})+\bar{\psi}_{\mathbf{FG}}\right| \stackrel{\mathbb{P}}{\rightarrow} 0 \text{ as } d \to\infty.$
\end{itemize}
\end{theorem}
\Cref{thm:3} states that if $\mathbf{Z} \sim \mathbf{F}$ (respectively, $\mathbf{Z} \sim \mathbf{G}$), the discriminant $\mathscr{D}_3$ corresponding to $\delta_3$ converges in probability to a positive (respectively, negative) quantity as $d\to \infty$, which justifies 
our construction of the classifier $\delta_3$ in \eqref{eq:3}.

\subsection{Misclassification Probabilities of $\delta_{1}, \delta_{2}$, and $\delta_{3}$ in the HDLSS asymptotic regime}
\label{sec:3.1}

We now show the convergence of the misclassification probabilities of our classifiers $\delta_{i}$ (denoted as $\Delta_{i}$), under some fairly general assumptions for $i=1,2,3$.

\begin{theorem}
\label{thm:4}
Suppose \cref{ass:1,ass:2} hold. Then,
$\Delta_{1} \stackrel{}{\rightarrow} 0$ and 
$\Delta_{2} \stackrel{}{\rightarrow} 0~ \text{ as } d \to\infty$. If, in addition, \cref{ass:3} holds, then
$\Delta_{3} \stackrel{}{\rightarrow} 0~ \text{ as } d \to\infty$. 
\end{theorem}

Observe that the asymptotic behaviors of the classifiers $\delta_1, \delta_2$, and $\delta_3$ are no longer governed by the constants $\lambda_{\mathbf{F}\mathbf{G}}, \sigma_{\mathbf{F}}$ and $\sigma_{\mathbf{G}}$. In fact, they are robust in terms of moment conditions since their behavior does not depend on the existence of any moments of $\mathbf{F}$ and $\mathbf{G}$ altogether.

\subsection{Comparison of the classifiers}
\label{sec:3.2}

Although the proposed classifiers yield perfect classification with increasing dimensions, they have some ordering among their misclassification rates under appropriate conditions. The following result describes the same.

\begin{theorem}
\label{thm:5}
Suppose \cref{ass:1,ass:2} hold. Then, 
\begin{itemize}
    \item[(a)] if $\liminf_{d\to\infty}\left(\max \{ T_{\mathbf{F F}}, T_{\mathbf{G G}}\}-T_{\mathbf{F G}}\right)>0,$ there exists $d_0'\in \mathbb N$ such that $\Delta_2 \leq \Delta_3 \leq \Delta_1$ for all $d \geq d_0',$
    \item[(b)] if $\liminf_{d\to\infty}\left(T_{\mathbf{F G}}-\max\{ T_{\mathbf{F F}}, T_{\mathbf{G G}}\}\right)>0$ and \cref{ass:3} holds, there exists $d_0'\in \mathbb N$, such that $\Delta_2 \geq \Delta_3 \geq \Delta_1$ for all $d \geq d_0'.$
\end{itemize}
\end{theorem}

\noindent\textbf{Remark.} If \cref{ass:3} is dropped from \Cref{thm:5}(b), it can still be concluded that if $\liminf_{d\to\infty}\left(T_{\mathbf{F G}}-\max \{ T_{\mathbf{F F}}, T_{\mathbf{G G}}\}\right)>0$, under \cref{ass:1,ass:2}, there exists $d_0'\in \mathbb N$ such that $\Delta_2 \geq \Delta_1$ for all $d \geq d_0'$ (see Lemma A.8(b) of the Supplementary Material).

We observe that $\delta_3$ always works `moderately' among the proposed classifiers, in the sense that its misclassification probability is neither the largest nor the smallest in both the aforementioned situations. It might be difficult to verify the conditions in \Cref{thm:5} in practice. Under such circumstances, it is more reasonable to use $\delta_3$ since it is the most `stable' among the proposed classifiers.


\section{Empirical Performance and Results}
\label{Empirical}
We examine the performance of our classifiers on a variety of simulated and real datasets, compared to several widely recognized classifiers such as GLMNET \citep{hastie2009elements}, Nearest Neighbor Random Projection (NN-RP) \citep{deegalla2006reducing}, Support Vector Machine with Linear (SVM-LIN) as well as Radial Basis Function (SVM-RBF) kernels \citep{cortes1995support}, Neural Networks (N-NET) \citep{bishop1995neural}, and $k$-Nearest Neighbor \citep{knn-paper-1,knn-paper-2} with $k=1$ (i.e., 1-NN). Additionally, the Bayes classifier is treated as a benchmark classifier in all the aforementioned simulated examples to assess the performances of the proposed classifiers, since it performs optimally when the true data distributions are known. All numerical exercises were executed on an Intel Xeon Gold 6140 CPU (2.30GHz, 2295 Mhz) using the R programming language \citep{R-lang}. Details about the packages used to implement the popular classifiers are provided in Section \ref{AppendixB} of Supplementary Material.

\subsection{Simulation Studies}

We perform our comparative study on five different simulated examples concerning different location problems as well as scale problems. In each example, we consider a binary classification problem with data simulated from two different $d$-variate distributions. Fixing the training sample size, we increase $d$ to mimic an HDLSS setting. In such situations, our proposed classifiers are expected to achieve perfect classification at higher values of $d$. We carry out our analysis for eight different values of $d$, namely, $5,10,25,$ $50,$ $100,$ $250,$ $500$, and $1000$.

\textbf{Examples 1} and \textbf{2} were already introduced in \Cref{Methodolgy}. We consider three more simulated examples as follows:
\begin{itemize}
    \item[] \textbf{Example 3:} $X_{1k} \stackrel{\text{i.i.d}}{\sim} C(0,1)$ and $Y_{1k} \stackrel{\text{i.i.d}}{\sim} C(1,1)$,

    \item[] \textbf{Example 4:} $X_{1k} \stackrel{\text{i.i.d}}{\sim} C(1,1)$ and $Y_{1k} \stackrel{\text{i.i.d}}{\sim} C(1,2)$,

    \item[] \textbf{Example 5:} $X_{1k} \stackrel{\text{i.i.d}}{\sim} \frac{9}{10} N(1,1) + \frac{1}{10} C(4,1)$ and $Y_{1k} \stackrel{\text{i.i.d}}{\sim} \frac{9}{10} N(1,2) + \frac{1}{10} C(4,1)$,
\end{itemize}
for $1\leq k \leq d$. Here, $C(\mu,\sigma^2)$ refers to Cauchy distribution with location $\mu$ and scale $\sigma$. \textbf{Example 3} and \textbf{Example 4} are location and scale problems, respectively. In \textbf{Example 5}, we consider the competing distributions to be $N(1,1)$ and $N(1,2)$ but with $10\%$ contamination from a $C(4,1)$ distribution. 

For all the examples, a training dataset was formed with a random sample of $20$ observations from each class, and a randomly generated test dataset of size $200$ ($100$ from each class) is used. The same process was repeated $100$ times independently, and all individual misclassification rates were averaged to estimate the probability of misclassification of $\delta_1, \delta_2$, and $\delta_3$ as well as the popular classifiers, which are reported in Section 2 of the Supplementary Material.

\begin{figure*}[!htb]
    \centering
    \includegraphics[width=\textwidth]{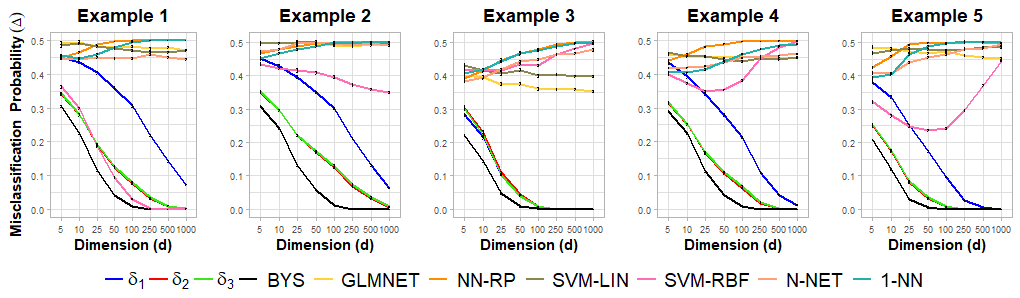}
    \caption{Average misclassification rates with errorbars for $\delta_1 , \delta_2$, and $ \delta_3,$ along with some popular classifiers for different dimensions. The Bayes classifier (assuming that the competing distributions in all examples are known) is treated as a benchmark.}
    \label{fig:comparison-sim-all}
\end{figure*}

In \textbf{Examples 2, 3,} and \textbf{4}, the competing distributions have identical first and second moments. Consequently, $\delta_0$ performs poorly in such situations. For this reason, we have dropped $\delta_0$ from further analysis. Plots of estimated misclassification probabilities of $\delta_1, \delta_2$, and $\delta_3$ along with those of the aforementioned popular classifiers are given in \Cref{fig:comparison-sim-all}. In each example, since the component variables are i.i.d., \cref{ass:1,ass:2} hold. Consequently, $\hat{T}_{IJ}$ is a consistent estimator of ${T}_{IJ}$ as $d\to \infty$ (see Lemma A.7(b) of the Supplementary Material) for $I,J\in \{\mathbf{F},\mathbf{G}\}$. Hence, we estimate ${T}_{IJ}$ by $\hat{T}_{IJ}$ to explain \Cref{fig:comparison-sim-all}. It can be observed from \Cref{fig:comparison-sim-all} that $\Delta_1,\Delta_2$ and $\Delta_3$ approach zero as $d$ increases for all the examples.

For \textbf{Examples 1, 2,} and \textbf{4}, we observe $\max \{ \hat{T}_{\mathbf{F F}}, \hat{T}_{\mathbf{G G}}\}>\hat{T}_{\mathbf{F G}}$. For these three examples, \Cref{fig:comparison-sim-all} shows that $\Delta_2 \leq \Delta_3 \leq \Delta_1$.
For \textbf{Example 3}, we observe that $\max \{ \hat{T}_{\mathbf{F F}}, \hat{T}_{\mathbf{G G}}\}<\hat{T}_{\mathbf{F G}}$ (see Table 1, Section B of the Supplementary Material). For this example, \Cref{fig:comparison-sim-all} shows that $\Delta_2 \geq \Delta_3 \geq \Delta_1$. Thus, the numerical findings are consistent with \Cref{thm:4,thm:5} (see \Cref{sec:3.1,sec:3.2}). 

\textbf{Example 5} was specially curated to validate the effectiveness of our classifiers in terms of robustness to outliers. In this example, we have considered two competing mixture distributions with contamination arising from the $C(4,1)$ distribution. All of $\delta_1$, $\delta_2$, and $\delta_3$ outperform the popular classifiers as $d$ increases. Even in the presence of contamination, all of our proposed classifiers tend to achieve perfect classification as $d$ increases.

\subsection{Implementation on Real Data}

Alongside the simulation studies, we implement our methods on several HDLSS datasets for a comprehensive performance evaluation. For each dataset, $50\%$ of the observations were selected at random to create a training set, while keeping the proportions of observations from each class consistent with those of all original datasets. The remaining observations were used to create the test set. To obtain stable estimates of the misclassification probabilities, this procedure was repeated $100$ times independently, and individual misclassification rates were averaged out to estimate the probability of misclassification of $\delta_1, \delta_2$, and $\delta_3$, as well as the popular classifiers.

Although our methods are primarily designed for binary classification, we implement a majority voting ensemble in the case of $J$-class problems with $J \geq 3$. Thus, for a dataset with $J$ different classes, we consider all $\binom{J}{2} = \frac{J(J-1)}{2}$ many unordered pairs of classes and treat them as separate binary classification problems. For each test observation, we perform classification for all of those $\binom{J}{2}$ many problems and classify the test observation to the class to which it gets assigned the maximum number of times. Ties are broken at random.

\setcounter{footnote}{2}

We conduct a thorough case study on six real HDLSS datasets, namely, GSE1577 and GSE89 from the Microarray database\footnote{\textbf{Microarray database:} Available at \url{https://file.biolab.si/biolab/supp/bi-cancer/projections/}}, Golub-1999-v2 and Gordon-2002 from the CompCancer database\footnote{\textbf{CompCancer database:} Available at \url{https://schlieplab.org/Static/Supplements/CompCancer/datasets.htm}}, and Computers and DodgerLoopDay from the UCR Time Series Classification Archive\footnote{\textbf{UCR TS Classification Archive:} Available at \url{https://www.cs.ucr.edu/~eamonn/time_series_data_2018/}} \citep{UCRArchive2018}. A brief description follows.

\begin{itemize}
    \item[$\bullet$] The \textbf{GSE1577} dataset consists of $19$ data points and $15434$ features. It is divided into $2$ classes which are T-cell lymphoblastic lymphoma (T-LL) and T-cell acute lymphoblastic leukemia (T-ALL).

    \item[$\bullet$] The \textbf{GSE89} dataset consists of $40$ data points and $5724$ features. It is divided into $3$ classes corresponding to three stages of tumor - T2+, Ta, and T1.
    
    \item[$\bullet$] The \textbf{Golub-1999-v2} dataset consists of $72$ data points and $1877$ features. It is divided into $3$ classes: Acute Myeloid Leukemia (AML), and two types of Acute Lymphoblastic Leukemia - B-cell ALL and T-cell ALL.

    \item[$\bullet$] The \textbf{Gordon-2002} dataset consists of $181$ data points and $1626$ features, divided into $2$ classes about the pathological distinction between malignant pleural mesothelioma (MPM) and adenocarcinoma (AD) of the lung.
    
    \item[$\bullet$] The \textbf{Computers} dataset contains readings on electricity consumption from $500$ households in the UK, sampled in two-minute intervals over a month. Each observation has $720$ features. The data points are categorized into two classes: `Desktop’ and ‘Laptop’.

    \item[$\bullet$] The \textbf{DodgerLoopDay} dataset consists of $158$ data points and $288$ features, divided into $7$ classes corresponding to the $7$ days of a week.
\end{itemize}
The estimated misclassification probabilities of $\delta_1, \delta_2$, and $\delta_3$, and the popular classifiers for these datasets are reported in \Cref{table:comparison-real-all}. The number of classes, data points, and features are denoted by $\operatorname{class}$, {\bf{$N$}}, and {\bf{$d$}}, respectively.

\begin{table*}[!htb]
\centering
\caption{Estimated misclassification probabilities (in \%) with standard errors (in parentheses) for $\delta_1, \delta_2$, and $\delta_3$, and popular classifiers for real datasets. For each dataset, the entries corresponding to the minimum misclassification rates are \textbf{boldfaced}.
}
\label{table1}
\renewcommand\theadfont{\bfseries}
\resizebox{\linewidth}{!}{
\begin{tabular}{lcccccccccccc}
\toprule
\multirow[b]{2}{*}{\thead{\textbf{Dataset}}}
    &   \multicolumn{3}{c}{\thead{\textbf{Description}}}
        & \multicolumn{6}{c}{\thead{\textbf{Popular Classifiers}}}
             & \multicolumn{3}{c}{\thead{\textbf{Proposed Classifiers}}}\\
\cmidrule(lr){2-4}\cmidrule(lr){5-10}\cmidrule(lr){11-13}
& \textbf{class} & \thead{$N$} & \thead{$d$} & \thead{GLM-\\NET} & \thead{NN-\\RP} & \thead{SVM-\\LIN} & \thead{SVM-\\RBF} & \thead{N-NET} & \thead{1-NN} & $\bm{\delta_1}$ & $\bm{\delta_2}$ & $\bm{\delta_3}$\\
\midrule
\addlinespace
\multirow[c]{2}{*}{GSE1577} & \multirow[c]{2}{*}{2} & \multirow[c]{2}{*}{19} & \multirow[c]{2}{*}{15434} & 6.51 & 11.59 & 6.38 & 30.06 & 33.87 & 11.12 & \textbf{6.06} & 7.33 & \textbf{6.06}\\
 &  &  &  & ~(0.66)~ & (0.44)~ & (0.35)~ & (0.36)~ & (0.98) & (0.39)~ & ~~(0.84)~ & ~(1.04) & (0.84)\\
 \addlinespace
\multirow[c]{2}{*}{GSE89} & \multirow[c]{2}{*}{3} & \multirow[c]{2}{*}{40} & \multirow[c]{2}{*}{5724} & 25.67 & 25.15 & 20.18 & 41.12 & 43.03 & 17.54 & \textbf{15.21} & 24.26 & 16.53\\
 &  &  &  & (0.47) & (0.49) & (1.05) & (0.46) & (1.92) & (0.28) & (0.80) & (0.97) & (0.96)\\
\addlinespace
\midrule
\addlinespace
\multirow[c]{2}{*}{Golub-1999-v2} & \multirow[c]{2}{*}{3} & \multirow[c]{2}{*}{72} & \multirow[c]{2}{*}{1877} & 8.78 & 15.28 & 10.55 & 33.05 & 75.26 & 8.98 & \textbf{6.89} & 9.40 & 7.49\\
 &  &  &  & (0.62) & (0.44) & (0.43) & (0.39) & (0.67) & (0.39) & (0.41) & (0.43) & (0.44)\\
 \addlinespace
\multirow[c]{2}{*}{Gordon-2002} & \multirow[c]{2}{*}{2} & \multirow[c]{2}{*}{181} & \multirow[c]{2}{*}{1626} & 2.58 & 4.11 & 1.26 & 1.84 & 11.28 & 2.69 & \textbf{0.53} & 0.54 & \textbf{0.53}\\
 &  &  &  & (0.47) & (0.12) & (0.09) & (0.09) & (2.83) & (0.10) & (0.01) & (0.01) & (0.01)\\
\addlinespace
\midrule
\addlinespace
\multirow[c]{2}{*}{Computers} & \multirow[c]{2}{*}{2} & \multirow[c]{2}{*}{500} & \multirow[c]{2}{*}{720} & 39.99 & 42.53 & 47.06 & 40.76 & 46.92 & 41.33 & 38.65 & \textbf{36.38} & 36.50\\
 &  &  &  & (0.69) & (0.62) & (0.39) & (0.27) & (0.41) & (0.57) & (0.28) & (0.21) & (0.22)\\
 \addlinespace
\multirow[c]{2}{*}{DodgerLoopDay} & \multirow[c]{2}{*}{7} & \multirow[c]{2}{*}{158} & \multirow[c]{2}{*}{288} & 55.45 & 48.72 & 39.42 & 47.03 & 71.32 & 47.38 & \textbf{37.68} & 44.73 & 42.23\\
 &  &  &  & (0.32) & (0.79) & (0.44) & (0.36) & (1.08) & (0.68) & (0.39) & (0.44) & (0.52)\\
\addlinespace
\bottomrule
\end{tabular}}
\label{table:comparison-real-all}
\end{table*}

For the datasets GSE1577, GSE89, Golub-1999-v2, Gordon-2002, and DodgerLoopDay, the estimated misclassification probabilities of our proposed classifiers are in the order $\Delta_1 \leq \Delta_3 \leq \Delta_2$, i.e., the performance of $\delta_1$ is the best for these examples and the misclassification probability of $\delta_3$ lies in between that of $\delta_1$ and $\delta_2$. To understand the relative performance of these classifiers, we computed $\hat{T}_{\mathbf{FF}}$, $\hat{T}_{\mathbf{FG}}$ and $\hat{T}_{\mathbf{GG}}$, and they satisfy $(\hat{T}_{\mathbf{FG}} - \max\{\hat{T}_{\mathbf{FF}},\hat{T}_{\mathbf{GG}}\}) > 0$. As discussed in \Cref{thm:5}, this ordering among the empirical versions of ${T}_{\mathbf{FF}}$, ${T}_{\mathbf{FG}}$ and ${T}_{\mathbf{GG}}$ is consistent with the relative ordering of the performances of $\delta_1$, $\delta_2$ and $\delta_3$. Furthermore, $\delta_1$ and $\delta_3$ performed better than all the popular classifiers. Although $\delta_2$ performed relatively worse than $\delta_1$ and $\delta_3$, it outperformed NN-RP, SVM-RBF, N-NET, and 1-NN.

For the dataset Computers, the estimated misclassification probabilities of our proposed classifiers are in the order $\Delta_2 \leq \Delta_3 \leq \Delta_1$, i.e., $\delta_2$ showed the best performance with a misclassification probability close to $36\%$. The misclassification probability of $\delta_3$ lies between that of $\delta_1$ and $\delta_2$. It turns out that $\hat{T}_{\mathbf{FF}}$, $\hat{T}_{\mathbf{FG}}$ and $\hat{T}_{\mathbf{GG}}$ satisfy $(\max\{\hat{T}_{\mathbf{FF}}, \hat{T}_{\mathbf{GG}}\} - \hat{T}_{\mathbf{FG}}) > 0$. This ordering among the empirical versions of ${T}_{\mathbf{FF}}$, ${T}_{\mathbf{FG}}$ and ${T}_{\mathbf{GG}}$ is consistent with the relative ordering of the performances of $\delta_1$, $\delta_2$ and $\delta_3$ (see \Cref{thm:5} of \Cref{sec:3.2}). All of $\delta_1$, $\delta_2$, and $\delta_3$ performed better than every popular classifier mentioned earlier.

\Cref{table:comparison-real-all} shows that $\delta_1$, $\delta_2$ and $\delta_3$ outperform widely recognized classifiers in a majority of the reported datasets, which establishes the merit of our proposed methods over the widely recognized ones. In addition, for all the reported datasets, the ordering among $\hat{T}_\mathbf{F F}$, $\hat{T}_\mathbf{F G}$ and $\hat{T}_\mathbf{G G}$ were found out to be consistent with the results stated in \Cref{thm:5}.

\section{Concluding Remarks}
In this paper, we developed some classification methods that draw good intuition from both classical and recent developments. We proved that under some general conditions, the misclassification probabilities of these classifiers steadily approach $0$ in the HDLSS asymptotic regime. The major advantages of our proposed methods are that they are free of tuning parameters, robust in terms of moment conditions, and easy to implement. Theoretical justification and comprehensive empirical studies against other well-established classification methods establish the advantages of our approach.

Nevertheless, when the competing distributions have at most $o(d)$ many different marginals, and the rest are identically distributed, \cref{ass:2,ass:3} will no longer hold. The theoretical guarantees for the optimal performance of the proposed classifiers will break down in such situations. Developing classifiers that avoid these assumptions is a fruitful avenue for further research.

\subsubsection*{Acknowledgement} 
We express our sincere gratitude to Professor Hernando Ombao (KAUST) for his generous support and insightful comments. We also thank the anonymous reviewers at ECML PKD 2023 for their comments on improving our manuscript.

\bibliographystyle{apalike}
\bibliography{references}

\begin{thebibliography}{}

\bibitem[Abdi and Williams, 2010]{abdi2010principal}
Abdi, H. and Williams, L.~J. (2010).
\newblock Principal {C}omponent {A}nalysis.
\newblock {\em Wiley Interdisciplinary Reviews: Computational Statistics}, 2(4):433--459.

\bibitem[Aggarwal et~al., 2001]{aggarwal2001surprising}
Aggarwal, C.~C., Hinneburg, A., and Keim, D.~A. (2001).
\newblock On the {S}urprising {B}ehavior of {D}istance {M}etrics in {H}igh {D}imensional {S}pace.
\newblock In {\em International Conference on Database Theory}, pages 420--434. Springer.

\bibitem[Baringhaus and Franz, 2004]{baringhaus2004new}
Baringhaus, L. and Franz, C. (2004).
\newblock On a {N}ew {M}ultivariate {T}wo-sample {T}est.
\newblock {\em Journal of Multivariate Analysis}, 88(1):190--206.

\bibitem[Beyer et~al., 1999]{beyer-et-al}
Beyer, K., Goldstein, J., Ramakrishnan, R., and Shaft, U. (1999).
\newblock When is ``{N}earest {N}eighbor'' {M}eaningful?
\newblock In Beeri, C. and Buneman, P., editors, {\em Database Theory --- ICDT'99}, pages 217--235, Berlin, Heidelberg. Springer Berlin Heidelberg.

\bibitem[Bishop et~al., 1995]{bishop1995neural}
Bishop, C.~M. et~al. (1995).
\newblock {\em Neural {N}etworks for {P}attern {R}ecognition}.
\newblock Oxford university press.

\bibitem[Biswas and Ghosh, 2014]{biswas2014nonparametric}
Biswas, M. and Ghosh, A.~K. (2014).
\newblock A nonparametric two-sample test applicable to high dimensional data.
\newblock {\em Journal of Multivariate Analysis}, 123:160--171.

\bibitem[Bradley, 2007]{bradley2007}
Bradley, R.~C. (2007).
\newblock {\em Introduction to {S}trong {M}ixing {C}onditions}.
\newblock Kendrick press.

\bibitem[Cortes and Vapnik, 1995]{cortes1995support}
Cortes, C. and Vapnik, V. (1995).
\newblock Support-vector {N}etworks.
\newblock {\em Machine learning}, 20(3):273--297.

\bibitem[Cover and Hart, 1967]{knn-paper-2}
Cover, T. and Hart, P. (1967).
\newblock Nearest {N}eighbor {P}attern {C}lassification.
\newblock {\em IEEE Transactions on Information Theory}, 13(1):21--27.

\bibitem[Dau et~al., 2018]{UCRArchive2018}
Dau, H.~A., Keogh, E., Kamgar, K., Yeh, C.-C.~M., Zhu, Y., Gharghabi, S., Ratanamahatana, C.~A., Yanping, Hu, B., Begum, N., Bagnall, A., Mueen, A., Batista, G., and Hexagon-ML (2018).
\newblock The {UCR} {T}ime {S}eries {C}lassification {A}rchive.

\bibitem[Deegalla and Bostrom, 2006]{deegalla2006reducing}
Deegalla, S. and Bostrom, H. (2006).
\newblock Reducing {H}igh-dimensional {D}ata by {P}rincipal {C}omponent {A}nalysis vs. {R}andom projection for {N}earest {N}eighbor {C}lassification.
\newblock In {\em 2006 5th International Conference on Machine Learning and Applications (ICMLA'06)}, pages 245--250. IEEE.

\bibitem[Fix and Hodges, 1989]{knn-paper-1}
Fix, E. and Hodges, J.~L. (1989).
\newblock Discriminatory {A}nalysis. {N}onparametric {D}iscrimination: {C}onsistency {P}roperties.
\newblock {\em International Statistical Review / Revue Internationale de Statistique}, 57(3):238--247.

\bibitem[Francois et~al., 2007]{franccois2007concentration}
Francois, D., Wertz, V., and Verleysen, M. (2007).
\newblock The {C}oncentration of {F}ractional {D}istances.
\newblock {\em IEEE Trans. on Knowl. and Data Eng.}, 19(7):873–886.

\bibitem[Hall et~al., 2005]{hall2005geometric}
Hall, P., Marron, J.~S., and Neeman, A. (2005).
\newblock Geometric {R}epresentation of {H}igh {D}imension, {L}ow {S}ample {S}ize {D}ata.
\newblock {\em Journal of the Royal Statistical Society: Series B (Statistical Methodology)}, 67(3):427--444.

\bibitem[Hastie et~al., 2009]{hastie2009elements}
Hastie, T., Tibshirani, R., Friedman, J.~H., and Friedman, J.~H. (2009).
\newblock {\em The {E}lements of {S}tatistical {L}earning: {D}ata {M}ining, {I}nference, and {P}rediction}, volume~2.
\newblock Springer.

\bibitem[Hinton and Salakhutdinov, 2006]{hinton2006reducing}
Hinton, G.~E. and Salakhutdinov, R.~R. (2006).
\newblock Reducing the {D}imensionality of {D}ata with {N}eural {N}etworks.
\newblock {\em science}, 313(5786):504--507.

\bibitem[Hyv{\"a}rinen and Oja, 2000]{hyvarinen2000independent}
Hyv{\"a}rinen, A. and Oja, E. (2000).
\newblock Independent {C}omponent {A}nalysis: {A}lgorithms and {A}pplications.
\newblock {\em Neural Networks}, 13(4-5):411--430.

\bibitem[Kim et~al., 2020]{10.1214/19-AOS1936}
Kim, I., Balakrishnan, S., and Wasserman, L. (2020).
\newblock {Robust {M}ultivariate {N}onparametric {T}ests via {P}rojection {A}veraging}.
\newblock {\em The Annals of Statistics}, 48(6):3417 -- 3441.

\bibitem[Li and Zhang, 2020]{pmlr-v119-li20s}
Li, Z. and Zhang, Y. (2020).
\newblock On a {P}rojective {E}nsemble {A}pproach to {T}wo {S}ample {T}est for {E}quality of {D}istributions.
\newblock In {\em Proceedings of the 37th International Conference on Machine Learning}, volume 119 of {\em Proceedings of Machine Learning Research}, pages 6020--6027. PMLR.

\bibitem[Pestov, 2013]{PESTOV20131427}
Pestov, V. (2013).
\newblock Is the k-{NN} {C}lassifier in {H}igh {D}imensions affected by the {C}urse of {D}imensionality?
\newblock {\em Computers \& Mathematics with Applications}, 65(10):1427--1437.

\bibitem[{R Core Team}, 2022]{R-lang}
{R Core Team} (2022).
\newblock {\em R: A Language and Environment for Statistical Computing}.
\newblock R Foundation for Statistical Computing, Vienna, Austria.

\bibitem[Roy et~al., 2022a]{pmlr-v151-roy22a}
Roy, S., Ray~Choudhury, J., and Dutta, S. (2022a).
\newblock On {S}ome {F}ast {A}nd {R}obust {C}lassifiers {F}or {H}igh {D}imension, {L}ow {S}ample {S}ize {D}ata.
\newblock In {\em Proceedings of The 25th International Conference on Artificial Intelligence and Statistics}, volume 151 of {\em Proceedings of Machine Learning Research}, pages 9943--9968. PMLR.

\bibitem[Roy et~al., 2022b]{roy2022generalizations}
Roy, S., Sarkar, S., Dutta, S., and Ghosh, A.~K. (2022b).
\newblock On {G}eneralizations of {S}ome {D}istance {B}ased {C}lassifiers for {HDLSS} {D}ata.
\newblock {\em Journal of Machine Learning Research}, 23(14):1--41.

\bibitem[Roy et~al., 2023]{roy2022exact}
Roy, S., Sarkar, S., Dutta, S., and Ghosh, A.~K. (2023).
\newblock On exact feature screening in ultrahigh-dimensional binary classification.

\bibitem[Sejdinovic et~al., 2013]{generalized-energy-dist}
Sejdinovic, D., Sriperumbudur, B., Gretton, A., and Fukumizu, K. (2013).
\newblock Equivalence of {D}istance-based and {RKHS}-based {S}tatistics in {H}ypothesis {T}esting.
\newblock {\em The Annals of Statistics}, 41(5):2263 -- 2291.

\bibitem[Shen et~al., 2022]{shen2022classification}
Shen, L., Er, M.~J., and Yin, Q. (2022).
\newblock Classification for {H}igh-dimension {L}ow-{S}ample {S}ize {D}ata.
\newblock {\em Pattern Recognition}, 130:108828.

\bibitem[Song et~al., 2011]{song2011fast}
Song, Q., Ni, J., and Wang, G. (2011).
\newblock A {F}ast {C}lustering-based {F}eature {S}ubset {S}election {A}lgorithm for {H}igh-dimensional {D}ata.
\newblock {\em IEEE Transactions on Knowledge and Data Engineering}, 25(1):1--14.

\bibitem[Szekely and Rizzo, 2004]{szekely-2004}
Szekely, G. and Rizzo, M. (2004).
\newblock Testing for {E}qual {D}istributions in {H}igh {D}imension.
\newblock {\em InterStat}, 5.

\bibitem[Tsukada, 2019]{Tsukada2019}
Tsukada, S.-i. (2019).
\newblock High {D}imensional {T}wo-sample {T}est {B}ased on the {I}nter-point {D}istance.
\newblock {\em Computational Statistics}, 34(2):599--615.

\bibitem[Wang et~al., 2014]{wang2014generalized}
Wang, W., Huang, Y., Wang, Y., and Wang, L. (2014).
\newblock Generalized {A}utoencoder: A {N}eural {N}etwork {F}ramework for {D}imensionality {R}eduction.
\newblock In {\em Proceedings of the IEEE Conference on Computer Vision and Pattern Recognition Workshops}, pages 490--497.

\bibitem[Yin et~al., 2020]{yin2020population}
Yin, Q., Adeli, E., Shen, L., and Shen, D. (2020).
\newblock Population-guided {L}arge {M}argin {C}lassifier for {H}igh-dimension low-sample-size {P}roblems.
\newblock {\em Pattern Recognition}, 97:107030.

\bibitem[Zou and Hastie, 2005]{zou2005regularization}
Zou, H. and Hastie, T. (2005).
\newblock Regularization and {V}ariable {S}election via the {E}lastic {N}et.
\newblock {\em Journal of the Royal Statistical Society: Series B (Statistical Methodology)}, 67(2):301--320.

\end{thebibliography}

\appendix



\newpage

\begin{center}
    \Large\textbf{Supplementary Material:\\
    Robust Classification of High-Dimensional Data\\ 
    using Data-Adaptive Energy Distance}    
\end{center}

\section{\textsc{Mathematical details \& Proofs}}
\label{AppendixA}
\begin{lemmaA}\label{lem:0.1}
    Suppose $\mathbf{U} \sim I, \mathbf{V} \sim J,$ and $\mathbf{Z} \sim K$ for $I, J, K\in \{\mathbf{F}, \mathbf{G}\}$, such that they are all independent. If assumptions 1 to 3 are satisfied, then 
$$\rho_0(\mathbf{U},\mathbf{V},\mathbf{Z})\stackrel{\mathbb{P}}{\rightarrow}\frac{1}{\pi}\cos^{-1}\left(\frac{\mu_{IJK}}{\sqrt{\mu_{IK}\mu_{JK}}}\right)  \text{~, as~}d \to \infty$$
where $\mu_{IJK}=\lim_d \frac{1}{d}\mathbb{E}((\mathbf{U}-\mathbf{Z})^{\top}(\mathbf{V}-\mathbf{Z}))$ and $~\mu_{IK}=\lim_d \frac{1}{d}\mathbb{E}(\|\mathbf{U}-\mathbf{Z}\|^2)$. 
\end{lemmaA}
\noindent\noindent\textbf{Proof of \Cref{lem:0.1}}

\noindent
Note that
\begin{align*}
\frac{1}{d}\mathbb{E}((\mathbf{U}-\mathbf{Z})^{\top}(\mathbf{V}-\mathbf{Z}))
&=\lim_d \frac{1}{d}\mathbb{E}[\mathbf{U}^{\top}\mathbf{V}-\mathbf{U}^{\top}\mathbf{Z}-\mathbf{V}^{\top}\mathbf{Z}+\mathbf{Z}^{\top}\mathbf{Z}]\\
&=\frac{1}{d}\left(\bm{\mu}_I^{\top}\bm{\mu}_J-\bm{\mu}_I^{\top}\bm{\mu_K}-\bm{\mu}_J^{\top}\bm{\mu_K}+\bm{\mu_K}^{\top}\bm{\mu_K}+\operatorname{trace}(\Sigma_K)\right)\\
&=\frac{1}{d}(\bm{\mu}_I-\bm{\mu}_K)^{\top}(\bm{\mu}_J-\bm{\mu}_K)+\frac{1}{d}\operatorname{trace}(\Sigma_K)
\end{align*}
Since $I,J,K\in \{\mathbf{F}, \mathbf{G}\}$, at least two of these take same value. Hence, using Assumption 2, we have
$$\mu_{IJK}=
\begin{cases}
\lambda_{IK}+\sigma^2_K, \text{ if } I=J\\
\sigma^2_K, \text{ otherwise. }
\end{cases}$$
Now, using Assumption 2, we shall show existence of the limiting constant $\mu_{IK}$.
\begin{align*}
\mu_{IK}&=\lim_d \frac{1}{d}\mathbb{E}[\mathbf{U}^{\top}\mathbf{U}-2\mathbf{U}^{\top}\mathbf{Z}+\mathbf{Z}^{\top}\mathbf{Z}]\\
&=\lim_d \frac{1}{d}\left(\bm{\mu}_I^{\top}\bm{\mu}_I+\operatorname{trace}(\Sigma_I)-2\bm{\mu}_I^{\top}\bm{\mu_K}+\bm{\mu_K}^{\top}\bm{\mu_K}+\operatorname{trace}(\Sigma_K)\right)\\
&=\lambda_{IK}+\sigma^2_I+\sigma^2_K.
\end{align*} 
Using Chebyshev’s inequality, we have
\begin{align*}
    &\mathbb{P}\left[\left|\frac{1}{d}(\mathbf{U}-\mathbf{Z})^{\top}(\mathbf{V}-\mathbf{Z})-\mathbb{E}\left(\frac{1}{d}(\mathbf{U}-\mathbf{Z})^{\top}(\mathbf{V}-\mathbf{Z})\right)\right|>\epsilon\right]\\
    &\leq \frac{1}{\epsilon^2}\text{var}\left(\frac{1}{d}\sum_{i=1}^n({U}_i-{Z}_i)({V}_i-{Z}_i)\right)\\
  &=\frac{1}{\epsilon^2 d^2}\bigg[\sum_{i=1}^d \text{var}(({U}_i-{Z}_i)({V}_i-{Z}_i))+\sum_{1\leq i<j\leq d}\operatorname{cov}(({U}_i-{Z}_i)({V}_i-{Z}_i),({U}_j-{Z}_j)({V}_j-{Z}_j))\bigg]\\ 
  &\leq \frac{1}{\epsilon^2 d^2}\bigg[\sum_{i=1}^d \mathbb{E}(({U}_i-{Z}_i)^2({V}_i-{Z}_i)^2)+\sum_{1\leq i<j\leq d}\operatorname{cov}(({U}_i-{Z}_i)({V}_i-{Z}_i),({U}_j-{Z}_j)({V}_j-{Z}_j))\bigg]\\ 
  & \to 0 ~\text{ ,  as } d \to \infty~~ [\text{Using }  \text{ assumptions 1 and 3}] .\\
\end{align*}
So, as $d \to \infty$,
\begin{eqnarray*}
 \left|\frac{1}{d}(\mathbf{U}-\mathbf{Z})^{\top}(\mathbf{V}-\mathbf{Z})-\mu_{IJK}\right| &\leq\left|\displaystyle\frac{1}{d}(\mathbf{U}-\mathbf{Z})^{\top}(\mathbf{V}-\mathbf{Z})-\mathbb{E}\left(\frac{1}{d}(\mathbf{U}-\mathbf{Z})^{\top}(\mathbf{V}-\mathbf{Z})\right)\right|\\
 &+\left|\mathbb{E}\left(\displaystyle\frac{1}{d}(\mathbf{U}-\mathbf{Z})^{\top}(\mathbf{V}-\mathbf{Z})\right)-\mu_{IJK}\right|\stackrel{\mathbb{P}}{\rightarrow}0.   
\end{eqnarray*}
Since $\mu_{IJK}$ is independent of $d$, we can write it as
$$\frac{1}{d}(\mathbf{U}-\mathbf{Z})^{\top}(\mathbf{V}-\mathbf{Z})\stackrel{\mathbb{P}}{\rightarrow}\mu_{IJK},  ~\text{ ,  as } d \to \infty. $$
Similarly, we have
\begin{equation*}
   \frac{1}{d}\|\mathbf{U}-\mathbf{Z}\|^2\stackrel{\mathbb{P}}{\rightarrow} \mu_{IK}~\text{ 
 and  }~~ \frac{1}{d}\|\mathbf{V}-\mathbf{Z}\|^2\stackrel{\mathbb{P}}{\rightarrow}\mu_{JK}~\text{,  as } d \to \infty.
\end{equation*}
Using the continuous mapping theorem repeatedly, we get
\begin{equation*}
\cos(\pi\cdot\rho_0(\mathbf{U},\mathbf{V},\mathbf{Z}))=\frac{\frac{1}{d}(\mathbf{U}-\mathbf{Z})^{\top}(\mathbf{V}-\mathbf{Z})}{\frac{1}{\sqrt{d}}\|\mathbf{U}-\mathbf{Z}\|.\frac{1}{\sqrt{d}}\|\mathbf{V}-\mathbf{Z}\|}\stackrel{\mathbb{P}}{\rightarrow}\frac{\mu_{IJK}}{\sqrt{\mu_{IK}\mu_{JK}}} ~\text{ ,  as } d \to \infty.
\end{equation*}
Therefore, we have
$$\rho_0(\mathbf{U},\mathbf{V},\mathbf{Z})\stackrel{\mathbb{P}}{\rightarrow}\frac{1}{\pi}\cos^{-1}\left(\frac{\mu_{IJK}}{\sqrt{\mu_{IK}\mu_{JK}}}\right)  \text{~, as~}d \to \infty.$$
\qed
\begin{lemmaA}\label{lem:0.2} 
Suppose, assumptions 1 to 3 are satisfied. Then, for $\mathbf{U} \sim I, \mathbf{V} \sim J$; $I, J\in \{\mathbf{F}, \mathbf{G}\} \text{~, as~}d \to \infty$,
$$\hat{\rho}(\mathbf{U},\mathbf{V})\stackrel{\mathbb{P}}{\rightarrow}\frac{1}{\pi(m+n)}\left(m\cos^{-1}\left(\frac{\mu_{IJ\mathbf{F}}}{\sqrt{\mu_{I\mathbf{F}}\mu_{J\mathbf{F}}}}\right)+n\cos^{-1}\left(\frac{\mu_{IJ\mathbf{G}}}{\sqrt{\mu_{I\mathbf{G}}\mu_{J\mathbf{G}}}}\right)\right).$$
In particular,
    $\hat{\rho}(\mathbf{X_i},\mathbf{X_j})\stackrel{\mathbb{P}}{\rightarrow}\theta_{\mathbf{FF}},~ \hat{\rho}(\mathbf{Y_i},\mathbf{Y_j})\stackrel{\mathbb{P}}{\rightarrow}\theta_{\mathbf{GG}},~
    \hat{\rho}(\mathbf{X_i},\mathbf{Y_j})\stackrel{\mathbb{P}}{\rightarrow}\theta_{\mathbf{FG}}  \text{~, as~}d \to \infty,$ where 
    \begin{equation*}
    \theta_{\mathbf{FF}}=\frac{1}{\pi(m+n)}\left(\frac{m\pi}{3}+n\cos^{-1}\left(\frac{\lambda_{\mathbf{FF}}+\lambda_{\mathbf{GG}}-2\lambda_{\mathbf{FG}}+\sigma^2_\mathbf{G}}{\lambda_{\mathbf{FF}}+\lambda_{\mathbf{GG}}-2\lambda_{\mathbf{FG}}+\sigma^2_\mathbf{G}+\sigma^2_\mathbf{F}}\right)\right),
\end{equation*}
\begin{equation*}
    \theta_{\mathbf{GG}}=\frac{1}{\pi(m+n)}\left(m\cos^{-1}\left(\frac{\lambda_{\mathbf{FF}}+\lambda_{\mathbf{GG}}-2\lambda_{\mathbf{FG}}+\sigma^2_\mathbf{F}}{\lambda_{\mathbf{FF}}+\lambda_{\mathbf{GG}}-2\lambda_{\mathbf{FG}}+\sigma^2_\mathbf{G}+\sigma^2_\mathbf{F}}\right)+\frac{n\pi}{3}\right) \text{, and }
\end{equation*}
\begin{eqnarray*}
    \theta_{\mathbf{F G}}&=\displaystyle\frac{1}{2}-\displaystyle\frac{1}{2\pi(m+n)}\bigg[m\cos^{-1}\left(\frac{\lambda_{\mathbf{FF}}+\lambda_{\mathbf{GG}}-2\lambda_{\mathbf{FG}}+\sigma^2_\mathbf{G}}{\lambda_{\mathbf{FF}}+\lambda_{\mathbf{GG}}-2\lambda_{\mathbf{FG}}+\sigma^2_\mathbf{G}+\sigma^2_\mathbf{F}}\right)\\
&\hspace{1.2cm}+n\cos^{-1}\left(\displaystyle\frac{\lambda_{\mathbf{FF}}+\lambda_{\mathbf{GG}}-2\lambda_{\mathbf{FG}}+\sigma^2_\mathbf{F}}{\lambda_{\mathbf{FF}}+\lambda_{\mathbf{GG}}-2\lambda_{\mathbf{FG}}+\sigma^2_\mathbf{G}+\sigma^2_\mathbf{F}}\right)\bigg] ~.
\end{eqnarray*}
\end{lemmaA}
\noindent\textbf{Proof of \Cref{lem:0.2}} From \Cref{lem:0.1} and definition of $\hat{\rho}$, we see that
\begin{equation*}
    \hat{\rho}(\mathbf{U},\mathbf{V})\stackrel{\mathbb{P}}{\rightarrow}\frac{1}{\pi(m+n)}\left(m\cos^{-1}\left(\frac{\mu_{IJ\mathbf{F}}}{\sqrt{\mu_{I\mathbf{F}}\mu_{J\mathbf{F}}}}\right)+n\cos^{-1}\left(\frac{\mu_{IJ\mathbf{G}}}{\sqrt{\mu_{I\mathbf{G}}\mu_{J\mathbf{G}}}}\right)\right), 
\end{equation*}
$\text{as~}d \to \infty$.
Therefore, as $d\to \infty$,
\begin{align*}
    \hat{\rho}(\mathbf{X}_i,\mathbf{X}_j)\stackrel{\mathbb{P}}{\rightarrow}&\frac{m\cos^{-1}\left(\frac{\mu_{FFF}}{\mu_{\mathbf{FF}}}\right)+n\cos^{-1}\left(\frac{\mu_{FFG}}{\mu_{\mathbf{FG}}}\right)}{\pi(m+n)}\\
    &=\frac{1}{\pi(m+n)}\left(\frac{m\pi}{3}+n\cos^{-1}\left(\frac{\lambda_{\mathbf{FF}}+\lambda_{\mathbf{GG}}-2\lambda_{\mathbf{FG}}+\sigma_G}{\lambda_{\mathbf{FF}}+\lambda_{\mathbf{GG}}-2\lambda_{\mathbf{FG}}+\sigma_G+\sigma_F}\right)\right),
\end{align*}
\begin{align*}
\hat{\rho}(\mathbf{Y}_i,\mathbf{Y}_j)\stackrel{\mathbb{P}}{\rightarrow}&\frac{m\cos^{-1}\left(\frac{\mu_{\mathbf{GGF}}}{\mu_{\mathbf{GF}}}\right)+n\cos^{-1}\left(\frac{\mu_{GGG}}{\mu_{\mathbf{GG}}}\right)}{\pi(m+n)}\\ &=\frac{1}{\pi(m+n)}\left(m\cos^{-1}\left(\frac{\lambda_{\mathbf{FF}}+\lambda_{\mathbf{GG}}-2\lambda_{\mathbf{FG}}+\sigma^2_\mathbf{F}}{\lambda_{\mathbf{FF}}+\lambda_{\mathbf{GG}}-2\lambda_{\mathbf{FG}}+\sigma^2_\mathbf{G}+\sigma^2_\mathbf{F}}\right)+\frac{n\pi}{3}\right),
\end{align*}
\begin{flalign*}
 \text{ and }&\hat{\rho}(\mathbf{X}_i,\mathbf{Y}_j)\stackrel{\mathbb{P}}{\rightarrow}\frac{1}{\pi(m+n)}\left(m\cos^{-1}\left(\frac{\mu_{\mathbf{FGF}}}{\sqrt{\mu_{\mathbf{FF}}\mu_{\mathbf{GF}}}}\right)+n\cos^{-1}\left(\frac{\mu_{\mathbf{FGG}}}{\sqrt{\mu_{\mathbf{FG}}\mu_{\mathbf{GG}}}}\right)\right) \cdot&&
 \end{flalign*}
 Note that
\begin{align*}&\frac{m\cos^{-1}\left(\frac{\mu_{\mathbf{FGF}}}{\sqrt{\mu_{\mathbf{FF}}\mu_{\mathbf{GF}}}}\right)+n\cos^{-1}\left(\frac{\mu_{\mathbf{FGG}}}{\sqrt{\mu_{\mathbf{FG}}\mu_{\mathbf{GG}}}}\right)}{\pi(m+n)}\\ =&\frac{m\cos^{-1}\left(\frac{\sqrt{\sigma^2_\mathbf{F}}}{\sqrt{2(\lambda_{\mathbf{FF}}+\lambda_{\mathbf{GG}}-2\lambda_{\mathbf{FG}}+\sigma^2_\mathbf{G}+\sigma^2_\mathbf{F})}}\right)+n\cos^{-1}\left(\frac{\sqrt{\sigma^2_\mathbf{G}}}{\sqrt{2(\lambda_{\mathbf{FF}}+\lambda_{\mathbf{GG}}-2\lambda_{\mathbf{FG}}+\sigma^2_\mathbf{G}+\sigma^2_\mathbf{F})}}\right)}{\pi(m+n)} \\ 
=&\frac{\pi(m+n)-m\cos^{-1}\left(\frac{\lambda_{\mathbf{FF}}+\lambda_{\mathbf{GG}}-2\lambda_{\mathbf{FG}}+\sigma^2_\mathbf{G}}{\lambda_{\mathbf{FF}}+\lambda_{\mathbf{GG}}-2\lambda_{\mathbf{FG}}+\sigma^2_\mathbf{G}+\sigma^2_\mathbf{F}}\right)-n\cos^{-1}\left(\frac{\lambda_{\mathbf{FF}}+\lambda_{\mathbf{GG}}-2\lambda_{\mathbf{FG}}+\sigma^2_\mathbf{F}}{\lambda_{\mathbf{FF}}+\lambda_{\mathbf{GG}}-2\lambda_{\mathbf{FG}}+\sigma^2_\mathbf{G}+\sigma^2_\mathbf{F}}\right)}{2\pi(m+n)}.
\end{align*}
\qed

\noindent We now define $$t_{\mathbf{F G}}=\mathbb{E}\left[{{\rho}}\left(\mathbf{X}_{1}, \mathbf{Y}_{1}\right)\right] \text{ and } \hat{t}_{\mathbf{F G}}=\frac{1}{mn} \sum_{i, j} \hat{{\rho}}\left(\mathbf{X}_{i}, \mathbf{Y}_{j}\right).$$

\begin{lemmaA}
\label{lem:0.3}
   Suppose, assumptions 1 to 3 are satisfied. Then, as $d \to \infty$,
\begin{align*}
\mathcal{W}_{\mathbf{FG}}^{*}\to2\theta_{\mathbf{FG}}-\theta_{\mathbf{FF}}-\theta_{\mathbf{GG}}=& \frac{2}{3}-\frac{1}{\pi}\bigg[\cos^{-1}\left(\frac{\lambda_{\mathbf{FF}}+\lambda_{\mathbf{GG}}-2\lambda_{\mathbf{FG}}+\sigma^2_\mathbf{G}}{\lambda_{\mathbf{FF}}+\lambda_{\mathbf{GG}}-2\lambda_{\mathbf{FG}}+\sigma^2_\mathbf{G}+\sigma^2_\mathbf{F}}\right)\\
&\qquad\qquad+\cos^{-1}\left(\frac{\lambda_{\mathbf{FF}}+\lambda_{\mathbf{GG}}-2\lambda_{\mathbf{FG}}+\sigma^2_\mathbf{F}}{\lambda_{\mathbf{FF}}+\lambda_{\mathbf{GG}}-2\lambda_{\mathbf{FG}}+\sigma^2_\mathbf{G}+\sigma^2_\mathbf{F}}\right)\bigg] .  
\end{align*}
\end{lemmaA}
\noindent\noindent\textbf{Proof of \Cref{lem:0.3}}
\begin{align*}
&\mathbb{E}[\hat{{\rho}}\left({\mathbf{X}}_{i}, {\mathbf{X}}_{j}\right)] - t_{\mathbf{F F}}\\
&=\mathbb{E}[\hat{{\rho}}\left({\mathbf{X}}_{i}, {\mathbf{X}}_{j}\right)] - \mathbb{E}\left[{{\rho}}\left({\mathbf{X}}_{i}, {\mathbf{X}}_{j}\right)\right]\\
&=\frac{m}{m+n} \mathbb{E}\left[\rho_0(\mathbf{\mathbf{X}}_{i},\mathbf{X}_{j};{\mathbf{X}}_{k})\right] + \frac{n}{m+n} \mathbb{E}\left[\rho_0(\mathbf{X}_{i},\mathbf{X}_{j};{\mathbf{Y}}_{k})\right] - \mathbb{E}\left[{{\rho}}\left({\mathbf{X}}_{i}, {\mathbf{X}}_{j}\right)\right]\\
&=\frac{m}{m+n} \mathbb{E}\left[\rho_0(\mathbf{X}_{i},\mathbf{X}_{j};{\mathbf{X}}_{k})\right] + \frac{n}{m+n} \mathbb{E}\left[\rho_0(\mathbf{X}_{i},\mathbf{X}_{j};{\mathbf{Y}}_{k})\right] - \mathbb{E}\left[\mathbb{E}_{\mathbf{Q}}[{{\rho}_0}\left({\mathbf{X}}_{i}, {\mathbf{X}}_{j}; \mathbf{Q}\right)\right]]\\
&\hspace{7cm}  \text{ where } \mathbf{Q} \sim \frac{m}{m+n} \mathbf{F} + \frac{n}{m+n} \mathbf{G}\\
&=\frac{m}{m+n} \mathbb{E}\left[\rho_0(\mathbf{X}_{i},\mathbf{X}_{j};{\mathbf{X}}_{3})\right] + \frac{n}{m+n} \mathbb{E}\left[\rho_0(\mathbf{X}_{i},\mathbf{X}_{j};{\mathbf{Y}}_{3})\right] - \mathbb{E}\left[{{\rho}}_0\left({\mathbf{X}}_{i}, {\mathbf{X}}_{j}; \mathbf{Q}\right)\right]\\
&\begin{aligned}
&=\frac{m}{m+n} \mathbb{E}\left[\rho_0(\mathbf{X}_{i},\mathbf{X}_{j};{\mathbf{X}}_{3})\right] + \frac{n}{m+n} \mathbb{E}\left[\rho_0(\mathbf{X}_{i},\mathbf{X}_{j};{\mathbf{Y}}_{3})\right]\\
&\qquad - \frac{m}{m+n} \mathbb{E}\left[{{\rho}}\left({\mathbf{X}}_{i}, {\mathbf{X}}_{j}; \mathbf{Q}\right)|\mathbf{Q} \sim \mathbf{F}\right]-\frac{n}{m+n}\mathbb{E}\left[{{\rho}}\left({\mathbf{X}}_{i}, {\mathbf{X}}_{j}; \mathbf{Q}\right)|\mathbf{Q} \sim \mathbf{G}\right]
\end{aligned}\\
&= 0.
\end{align*}
Since, $\hat{\rho}(\mathbf{X_i},\mathbf{X_j})\stackrel{\mathbb{P}}{\rightarrow}\theta_{\mathbf{FF}}$ and $\hat{\rho}$ is a bounded function, using the Dominated Convergence Theorem, we have
\begin{align*}
   t_{\mathbf{FF}}= \mathbb{E}[\hat{\rho}(\mathbf{X_i},\mathbf{X_j})]\to\theta_{\mathbf{FF}} \text{~, as~}d \to \infty.
\end{align*}
Similarly, we can show that $t_{\mathbf{GG}}\to \theta_{\mathbf{GG}} $ and $t_{\mathbf{FG}}\to \theta_{\mathbf{FG}} $, as $d\to \infty$.
Hence, $\mathcal{W}_{\mathbf{FG}}^{*}\to 2\theta_{\mathbf{FG}}-\theta_{\mathbf{FF}}-\theta_{\mathbf{GG}} \text{~, as~}d \to \infty$. Substituting the values of $\theta_{\mathbf{FG}},\theta_{\mathbf{FF}}$ and $\theta_{\mathbf{GG}}$, we get 
\begin{align*}
\mathcal{W}_{\mathbf{FG}}^{*}\to & \frac{2}{3}-\frac{1}{\pi}\bigg[\cos^{-1}\left(\frac{\lambda_{\mathbf{FF}}+\lambda_{\mathbf{GG}}-2\lambda_{\mathbf{FG}}+\sigma^2_\mathbf{G}}{\lambda_{\mathbf{FF}}+\lambda_{\mathbf{GG}}-2\lambda_{\mathbf{FG}}+\sigma^2_\mathbf{G}+\sigma^2_\mathbf{F}}\right)\\
&+\cos^{-1}\left(\frac{\lambda_{\mathbf{FF}}+\lambda_{\mathbf{GG}}-2\lambda_{\mathbf{FG}}+\sigma^2_\mathbf{F}}{\lambda_{\mathbf{FF}}+\lambda_{\mathbf{GG}}-2\lambda_{\mathbf{FG}}+\sigma^2_\mathbf{G}+\sigma^2_\mathbf{F}}\right)\bigg] \text{~, as~}d \to \infty .  
\end{align*}
\qed

\begin{lemmaA}
\label{lem:0.4}Suppose, assumptions 1 to 3 are satisfied. Then,
\begin{itemize}
\item[(a)] i. $\hat{t}_{\mathbf{F F}}\stackrel{\mathbb{P}}{\rightarrow} \theta_{\mathbf{F F}}  \text{~, as~}d \to \infty ~;$

ii. $\hat{t}_{\mathbf{G G}}\stackrel{\mathbb{P}}{\rightarrow} \theta_{\mathbf{G G}}   \text{~, as~}d \to \infty ~;$

iii. $\hat{t}_{\mathbf{F G}} \stackrel{\mathbb{P}}{\rightarrow} \theta_{\mathbf{F G}}  \text{~, as~}d \to \infty .$

\item[(b)] i. if $\mathbf{Z} \sim \mathbf{F}$, then $\hat{t}_{\mathbf{F}}(\mathbf{Z}) \stackrel{\mathbb{P}}{\rightarrow} \theta_{\mathbf{F F}}$ and $\hat{t}_{\mathbf{G}}(\mathbf{Z})\stackrel{\mathbb{P}}{\rightarrow} \theta_{\mathbf{F G}}  \text{~, as~}d \to \infty ~;$

ii. if $\mathbf{Z} \sim \mathbf{G}$, then $\hat{t}_{\mathbf{F}}(\mathbf{Z})\stackrel{\mathbb{P}}{\rightarrow} \theta_{\mathbf{F G}} $ and $\hat{t}_{\mathbf{G}}(\mathbf{Z}) \stackrel{\mathbb{P}}{\rightarrow} \theta_{\mathbf{G G}} \text{~, as~}d \to \infty ~.$
\end{itemize}
\end{lemmaA}
\noindent\noindent\textbf{Proof of \Cref{lem:0.4}}
\begin{itemize}
    \item[(a)]Since $\hat{\rho}(\mathbf{X_i},\mathbf{X_j})\stackrel{\mathbb{P}}{\rightarrow}\theta_{\mathbf{FF}}$, 
$$\hat{t}_{\mathbf{F F}}=\frac{1}{m(m-1)} \sum_{i \neq j} \hat{{\rho}}\left(\mathbf{X}_{i}, \mathbf{X}_{j}\right)\stackrel{\mathbb{P}}{\rightarrow}\theta_{\mathbf{FF}}\text{~, as~}d \to \infty.$$
Similarly, $\hat{t}_{\mathbf{G G}}\stackrel{\mathbb{P}}{\rightarrow} \theta_{\mathbf{G G}}$
and $\hat{t}_{\mathbf{F G}} \stackrel{\mathbb{P}}{\rightarrow} \theta_{\mathbf{F G}}  \text{~, as~}d \to \infty$.

\vspace{0.3cm}\item[(b)] If $\mathbf{Z} \sim \mathbf{F}$, $\hat{\rho}(\mathbf{X_i},\mathbf{Z})\stackrel{\mathbb{P}}{\rightarrow}\theta_{\mathbf{FF}}$ and $\hat{\rho}(\mathbf{Y_i},\mathbf{Z})\stackrel{\mathbb{P}}{\rightarrow}\theta_{\mathbf{FG}} \text{~, as~}d \to \infty$. 
So, $$\hat{t}_{\mathbf{F}}(\mathbf{Z})=\frac{1}{n} \sum_{i} \hat{{\rho}}\left(\mathbf{X}_{i}, \mathbf{Z}\right)\stackrel{\mathbb{P}}{\rightarrow}\theta_{\mathbf{FF}}\text{  and  }\hat{t}_{\mathbf{G}}(\mathbf{Z})=\frac{1}{n} \sum_{i} \hat{{\rho}}\left(\mathbf{Y}_{i}, \mathbf{Z}\right)\stackrel{\mathbb{P}}{\rightarrow}\theta_{\mathbf{FG}},$$ as $d \to \infty$.
Similarly, if $\mathbf{Z} \sim \mathbf{G}$, $\hat{t}_{\mathbf{F}}(\mathbf{Z})\stackrel{\mathbb{P}}{\rightarrow} \theta_{\mathbf{F G}} $ and $\hat{t}_{\mathbf{G}}(\mathbf{Z}) \stackrel{\mathbb{P}}{\rightarrow} \theta_{\mathbf{G G}} \text{~, as~}d \to \infty$.
\end{itemize}
\qed

\noindent\textbf{{Proof of Theorem 1}}\\

From \Cref{lem:0.3}, we have $\theta^{*}_{\mathbf{FG}}=\lim_d\mathcal{W}^{*}_{\mathbf{FG}}=2\theta_{\mathbf{FG}}-\theta_{\mathbf{FF}}-\theta_{\mathbf{GG}}$.
Given $\mathbf{Z} \sim \mathbf{F}$,
\begin{align*}
    &\left|l_{\mathbf{G}}(\mathbf{Z})-l_{\mathbf{F}}(\mathbf{Z})-\frac{1}{2} \theta^{*}_{\mathbf{FG}}\right|\\
    &=\left|(\hat{t}_{\mathbf{G}}(\mathbf{Z})-\theta_{\mathbf{F G}})-(\hat{t}_{\mathbf{F}}(\mathbf{Z})- \theta_{\mathbf{F F}} )+\frac{1}{2}(\hat{t}_{\mathbf{F F}}-\theta_{\mathbf{F F}})-\frac{1}{2}(\hat{t}_{\mathbf{G G}}-\theta_{\mathbf{G G}}) \right|\\
    &\leq |\hat{t}_{\mathbf{G}}(\mathbf{Z})-\theta_{\mathbf{F G}}|+|\hat{t}_{\mathbf{F}}(\mathbf{Z})- \theta_{\mathbf{F F}}|+\frac{1}{2}|\hat{t}_{\mathbf{F F}}-\theta_{\mathbf{F F}}|+\frac{1}{2}|\hat{t}_{\mathbf{G G}}-\theta_{\mathbf{G G}}|\\
    &\stackrel{\mathbb{P}}{\rightarrow}0,  \text{~, as~}d \to \infty \text{ [By using } \Cref{lem:0.4}].
\end{align*}
Therefore, if $\mathbf{Z} \sim \mathbf{F}$, $l_{\mathbf{G}}(\mathbf{Z})-l_{\mathbf{F}}(\mathbf{Z})\stackrel{\mathbb{P}}{\rightarrow}\frac{1}{2}\theta^{*}_{\mathbf{FG}}  ~ \text{, as } d \to\infty~.$\\

\noindent
Part (b) can be shown in an exactly similar way.
\qed

\begin{lemmaA}
\label{lem:0.6}
$2\theta_{\mathbf{FG}}-\theta_{\mathbf{FF}}-\theta_{\mathbf{GG}}=0$ if and only if $\sigma^2_\mathbf{G}=\sigma^2_\mathbf{F}$ and $\lambda_{\mathbf{FF}}+\lambda_{\mathbf{GG}}-2\lambda_{\mathbf{FG}}=0$.
\end{lemmaA}
\noindent\textbf{Proof of \Cref{lem:0.6}}
\begin{align*}
2\theta_{\mathbf{FG}}-\theta_{\mathbf{FF}}-\theta_{\mathbf{GG}}=& \frac{2}{3}-\frac{1}{\pi}\bigg[\cos^{-1}\left(\frac{\lambda_{\mathbf{FF}}+\lambda_{\mathbf{GG}}-2\lambda_{\mathbf{FG}}+\sigma^2_\mathbf{G}}{\lambda_{\mathbf{FF}}+\lambda_{\mathbf{GG}}-2\lambda_{\mathbf{FG}}+\sigma^2_\mathbf{G}+\sigma^2_\mathbf{F}}\right)\\
&\qquad +\cos^{-1}\left(\frac{\lambda_{\mathbf{FF}}+\lambda_{\mathbf{GG}}-2\lambda_{\mathbf{FG}}+\sigma_{\mathbf{F}}}{\lambda_{\mathbf{FF}}+\lambda_{\mathbf{GG}}-2\lambda_{\mathbf{FG}}+\sigma^2_\mathbf{G}+\sigma^2_\mathbf{F}}\right)\bigg].
\end{align*}
Let, $x=\displaystyle\frac{\lambda_{\mathbf{FF}}+\lambda_{\mathbf{GG}}-2\lambda_{\mathbf{FG}}}{\sigma^2_\mathbf{G}+\sigma^2_\mathbf{F}}$ and $\alpha=\displaystyle\frac{\sigma^2_\mathbf{F}}{\sigma^2_\mathbf{G}+\sigma^2_\mathbf{F}}$. Note that $x\geq 0$ and $\alpha \in (0,1)$.

We need to solve $\cos^{-1}\left(\frac{x+\alpha}{x+1}\right)+\cos^{-1}\left(\frac{x+1-\alpha}{x+1}\right)=\frac{2\pi}{3}$, for $x\geq 0$ and $\alpha \in (0,1)$, which is equivalent to solve
\begin{align*}
   & \cos^{-1}\left(\frac{x+\alpha}{x+1}\cdot\frac{x+1-\alpha}{x+1}-\sqrt{\left(1-\left(\frac{x+\alpha}{x+1}\right)^2\right)\left(1-\left(\frac{x+1-\alpha}{x+1}\right)^2\right)}\right)=\frac{2\pi}{3}\\
    &\implies \frac{x+\alpha}{x+1}\cdot\frac{x+1-\alpha}{x+1}-\sqrt{\left(1-\left(\frac{x+\alpha}{x+1}\right)^2\right)\left(1-\left(\frac{x+1-\alpha}{x+1}\right)^2\right)}=-\frac{1}{2}\\
   &\implies x^2+\alpha(1-\alpha)+x-\sqrt{\alpha(1-\alpha)(4x^2+2+6x+\alpha(1-\alpha))}=-\frac{1}{2}(x+1)^2\\
   &\implies \left(\frac{1}{2}(x+1)(3x+1)+\beta\right)^2=\beta(4x^2+2+6x+\beta) \text{ ; substituting, }  \beta=\alpha(1-\alpha)\\
   &\implies \beta=\frac{(3x+1)^2}{4}\geq \frac{1}{4} \text{ since } x\geq 0.
\end{align*}
But, since $\alpha\in (0,1)$, by AM-GM inequality, $\beta=\alpha(1-\alpha)\leq\frac{1}{4}$, equality holds iff $\alpha=\frac{1}{2}.$ So, we must have $\beta=\frac{1}{4}$ which will imply $\alpha=\frac{1}{2}$ and $x=0$, i.e, $\sigma_G=\sigma_F$ and $\lambda_{\mathbf{FF}}+\lambda_{\mathbf{GG}}-2\lambda_{\mathbf{FG}}=0$.

Also, $\sigma_{\mathbf{G}}=\sigma_{\mathbf{F}}$ and $\lambda_{\mathbf{FF}}+\lambda_{\mathbf{GG}}-2\lambda_{\mathbf{FG}}=0$ implies $2\theta_{\mathbf{FG}}-\theta_{\mathbf{FF}}-\theta_{\mathbf{GG}}=0$.

\qed

\noindent\textbf{Proof of Theorem 2}\\\\
The misclassification probability of the classifier $\delta_{0}$ can be written as
\begin{align*}
\Delta_0  
&=\mathbb{P}\left[\delta_{0}(\mathbf{Z})=2, \mathbf{Z} \sim \mathbf{F}\right]+\mathbb{P}\left[\delta_{0}(\mathbf{Z})=1, \mathbf{Z} \sim \mathbf{G}\right] \\
&=\frac{m}{m+n} \mathbb{P}\left[\delta_{0}(\mathbf{Z})=2 \mid \mathbf{Z} \sim \mathbf{F}\right]+\frac{n}{m+n} \mathbb{P}\left[\delta_{0}(\mathbf{Z})=1 \mid \mathbf{Z} \sim \mathbf{G}\right] \\
&=\frac{m}{m+n} \mathbb{P}\left[l_{\mathbf{G}}(\mathbf{Z})-l_{\mathbf{F}}(\mathbf{Z}) \leq 0 \mid \mathbf{Z} \sim \mathbf{F}\right]+\frac{n}{m+n}
\mathbb{P}\left[l_{\mathbf{G}}(\mathbf{Z})-l_{\mathbf{F}}(\mathbf{Z})>0 \mid \mathbf{Z} \sim \mathbf{G}\right].
\end{align*}
By \Cref{lem:0.6}, we know that if either $\displaystyle\lim_{d \to \infty} \tfrac{1}{d}\|\bm{\mu}_\mathbf{F}-\bm{\mu}_\mathbf{G}\|^2\neq0~$ 
   or $\sigma^2_\mathbf{F}\neq\sigma^2_\mathbf{G}~$ holds, 
$2\theta_{\mathbf{FG}}-\theta_{\mathbf{FF}}-\theta_{\mathbf{GG}}>0$.
We can choose $\epsilon>0$ such that $\epsilon<\frac{1}{2} (2\theta_{\mathbf{FG}}-\theta_{\mathbf{FF}}-\theta_{\mathbf{GG}})$. Therefore, we have:
$$
\begin{aligned}
&\mathbb{P}\left[l_{\mathbf{G}}(\mathbf{Z})-l_{\mathbf{F}}(\mathbf{Z}) \leq 0 \mid \mathbf{Z} \sim \mathbf{F}\right] \\
 &\leq \mathbb{P}\left[l_{\mathbf{G}}(\mathbf{Z})-l_{\mathbf{F}}(\mathbf{Z}) \leq \frac{1}{2} (2\theta_{\mathbf{FG}}-\theta_{\mathbf{FF}}-\theta_{\mathbf{GG}})-\epsilon \mid \mathbf{Z} \sim \mathbf{F}\right] \\
 &\leq \mathbb{P}\left[l_{\mathbf{G}}(\mathbf{Z})-l_{\mathbf{F}}(\mathbf{Z})-\frac{1}{2} (2\theta_{\mathbf{FG}}-\theta_{\mathbf{FF}}-\theta_{\mathbf{GG}}) \leq-\epsilon \mid \mathbf{Z} \sim \mathbf{F}\right] \\
&\leq \mathbb{P}\left[\left|l_{\mathbf{G}}(\mathbf{Z})-l_{\mathbf{F}}(\mathbf{Z})-\frac{1}{2} (2\theta_{\mathbf{FG}}-\theta_{\mathbf{FF}}-\theta_{\mathbf{GG}})\right|>\epsilon \mid \mathbf{Z} \sim \mathbf{F}\right] ~\to 0 \text{~, as~}d \to \infty~.\\
&\hspace{9cm}\text{[Using Theorem 1]}
\end{aligned}
$$
Similarly, one can show that $$\mathbb{P}\left[l_{\mathbf{G}}(\mathbf{Z})-l_{\mathbf{F}}(\mathbf{Z})> 0 \mid \mathbf{Z} \sim \mathbf{G}\right] \to 0 \text{ , as } d \to \infty.$$

\noindent Thus, we conclude that $\Delta_0 \to 0$ as $d \to \infty$.

\qed

\begin{lemmaA}
\label{lem:1}
It follows from Assumption 4 that  \begin{equation*}
    \sum_{1\leq k_1<k_2\leq d}\operatorname{cov}(\hat{\rho}(U_{k_1}, V_{k_1}), \hat{\rho}(U_{k_2} , V_{k_2} )) = o(d^2).
\end{equation*}   
\end{lemmaA}
\noindent\textbf{Proof of \Cref{lem:1}}

Let, $\mathcal{X}=\{\m X_1,\cdots,\m X_m\}$ and $\mathcal{Y}=\{\m Y_1,\cdots,\m Y_n\}$.
   Then, 
 \begin{align*}
     \operatorname{cov}(\hat{\rho}(U_{k_1}, V_{k_1}), \hat{\rho}(U_{k_2} , V_{k_2} ))
     &=\frac{1}{(m+n)^2}\sum_{\m Q,\m Q^* \in \mathcal{X} \cup \mathcal{Y}} \operatorname{cov}(\rho_0(U_{k_1},V_{k_1};Q_{k_1}), \rho_0(U_{k_2},V_{k_2};Q^{*}_{k_2})).
 \end{align*}

Hence, it follows that 
\begin{align*}
    &\sum_{1\leq k_1<k_2\leq d}\operatorname{cov}(\hat{\rho}(U_{k_1}, V_{k_1}), \hat{\rho}(U_{k_2} , V_{k_2} ))\\
    &=\frac{1}{(m+n)^2}\sum_{1\leq k_1<k_2\leq d}\sum_{\m Q,\m Q^* \in \mathcal{X} \cup \mathcal{Y}} \operatorname{cov}(\rho_0(U_{k_1},V_{k_1};Q_{k_1}), \rho_0(U_{k_2},V_{k_2};Q^{*}_{k_2}))\\
    &=\frac{1}{(m+n)^2}\sum_{\m Q,\m Q^* \in \mathcal{X} \cup \mathcal{Y}} \sum_{1\leq k_1<k_2\leq d}\operatorname{cov}(\rho_0(U_{k_1},V_{k_1};Q_{k_1}), \rho_0(U_{k_2},V_{k_2};Q^{*}_{k_2}))\\ &\hspace{5cm}\text{  [Since the sum over } \m Q,\m Q^* \text{ is finite sum.]}\\
    &= o(d^2)  \text{ [By using Assumption 4]. }
\end{align*} \qed

We now define $$T_{\mathbf{F G}}=\mathbb{E}\left[{\bar{\rho}}\left(\mathbf{X}_{1}, \mathbf{Y}_{1}\right)\right] \text{ ~and~ } \hat{T}_{\mathbf{F G}}=\frac{1}{n m} \sum_{i, j} \hat{\bar{\rho}}\left(\mathbf{X}_{i}, \mathbf{Y}_{j}\right).$$

\begin{lemmaA}
   \label{lem:A1}
   Suppose Assumption 4 is satisfied. Then,
   \begin{itemize}
\item[(a)] Irrespective of whether $\mathbf{U}, \mathbf{V}$ are coming from $\mathbf{F}$ and/or $\mathbf{G}$,
$$
\hat{\bar{\rho}}(\mathbf{U}, \mathbf{V})-\mathbb{E}[\hat{\rho}(\mathbf{U}, \mathbf{V})] \stackrel{\mathbb{P}}{\rightarrow} 0 \text{ ,  as } d \to\infty
$$

\item[(b)] i. $\hat{T}_{\mathbf{F F}}-T_{\mathbf{F F}} \stackrel{\mathbb{P}}{\rightarrow} 0 \text{ ,  as } d \to\infty ~;$

ii. $\hat{T}_{\mathbf{G G}}-T_{\mathbf{G G}} \stackrel{\mathbb{P}}{\rightarrow} 0 \text{ ,  as } d \to\infty ~;$

iii. $\hat{T}_{\mathbf{F G}}-T_{\mathbf{F G}} \stackrel{\mathbb{P}}{\rightarrow} 0 \text{ ,  as } d \to\infty ~.$

\item[(c)] i. If $\mathbf{Z} \sim \mathbf{F}$, then as $d \to \infty$, $\hat{T}_{\mathbf{F}}(\mathbf{Z})-T_{\mathbf{F F}} \stackrel{\mathbb{P}}{\rightarrow} 0$ and $\hat{T}_{\mathbf{G}}(\mathbf{Z})-T_{\mathbf{F G}} \stackrel{\mathbb{P}}{\rightarrow} 0 ~;$

ii. If $\mathbf{Z} \sim \mathbf{G}$, then as $d \to \infty$, $\hat{T}_{\mathbf{F}}(\mathbf{Z})-T_{\mathbf{F G}} \stackrel{\mathbb{P}}{\rightarrow} 0$ and $\hat{T}_{\mathbf{G}}(\mathbf{Z})-T_{\mathbf{G G}} \stackrel{\mathbb{P}}{\rightarrow} 0~.$
  \end{itemize} 
\end{lemmaA}

\noindent\textbf{Proof of \Cref{lem:A1}}
\begin{itemize}
\item[\textbf{(a)}]
For any $\epsilon> 0$, using \textsl{Chebyshev's Inequality}, we have,
$$\mathbb{P}[|\hat{\bar{\rho}}\left(\mathbf{U}, \mathbf{V}\right)-\mathbb{E}[\hat{\bar{\rho}}\left(\mathbf{U}, \mathbf{V}\right)]|>\epsilon] \leq \frac{\text{var}(\hat{\bar{\rho}}\left(\mathbf{U}, \mathbf{V}\right))}{\epsilon^2} \cdot$$
Note that $\rho_0 \in [-1,1]$. Thus, $\hat{\rho}$ being a convex combination of some $\rho_0$, lies between $[-1,1]$. Also, since $\rho(a,b) = \mathbb{E}[\rho_0(a,b)]$, we have $\hat{{\rho}} \in [-1,1]$. So, $\mathbb{E}({\hat{\rho}^2}\left({U_i}, {V_i}\right))\leq 1$.
\begin{align*}
&\mathbb{P}[|\hat{\bar{\rho}}\left(\mathbf{U}, \mathbf{V}\right)-\mathbb{E}[\hat{\bar{\rho}}\left(\mathbf{U}, \mathbf{V}\right)]|>\epsilon] \\
& \leq \frac{1}{\epsilon^2}\text{var}(\frac{1}{d}\sum_{i=1}^d{\hat{\rho}}\left({U_i},{V_i}\right))\\
  &=\frac{1}{\epsilon^2 d^2}\left[\sum_{i=1}^d \text{var}({\hat{\rho}}\left({U_i}, {V_i}\right))+\sum_{1\leq i<j\leq d}\operatorname{cov}(\hat{\rho}(U_{i}, V_{i}), \hat{\rho}(U_{j} , V_{j} ))\right]\\
  &\leqslant \frac{1}{\epsilon^2 d^2}\left[\sum_{i=1}^d \mathbb{E}({\hat{\rho}^2}\left({U_i}, {V_i}\right))+\sum_{1\leq i<j\leq d}\operatorname{cov}(\hat{\rho}(U_{i}, V_{i}), \hat{\rho}(U_{j} , V_{j} ))\right]\\
  &\leqslant \frac{1}{\epsilon^2 d^2}\left[d + o(d^2)\right]~\to0 ~\text{ ,  as } d \to\infty.
  \end{align*}  
  The last assertion holds good due to \Cref{lem:1} and the fact that $\mathbb{E}({\hat{\rho}^2}\left({U_i}, {V_i}\right))\leq 1$.
Therefore, we have: $$|\hat{\bar{\rho}}\left(\mathbf{U}, \mathbf{V}\right)-\mathbb{E}[\hat{\bar{\rho}}\left(\mathbf{U}, \mathbf{V}\right)]| \stackrel{\mathbb{P}}{\rightarrow} 0 \text{, as } d\to \infty.$$
    
\item[\textbf{(b)}]
    Once we have proved part \textbf{(a)}, we have the following: \begin{align*}
      \hat{T}_{\mathbf{F F}}-{T}_{\mathbf{F F}} &= \frac{1}{m(m-1)} \sum_{i \neq j} \Big[\hat{\bar{\rho}}\left(\mathbf{X}_{i}, \mathbf{X}_{j}\right) - \mathbb{E}[\hat{\bar{\rho}}\left(\mathbf{X}_{i}, \mathbf{X}_{j}\right)] + \mathbb{E}[{\bar{\rho}}\left(\mathbf{X}_{i}, \mathbf{X}_{j}\right)] - \mathbb{E}\left[\hat{\bar{\rho}}\left(\mathbf{X}_{i}, \mathbf{X}_{j}\right)\right]\Big]\\
      &= \frac{1}{m(m-1)} \sum_{i \neq j} \Big[\hat{\bar{\rho}}\left(\mathbf{X}_{i}, \mathbf{X}_{j}\right) - \mathbb{E}[\hat{\bar{\rho}}\left(\mathbf{X}_{i}, \mathbf{X}_{j}\right)]\Big] \\
  &\qquad+ \frac{1}{m(m-1)} \sum_{i \neq j} \Big[\mathbb{E}[\hat{\bar{\rho}}\left(\mathbf{X}_{i}, \mathbf{X}_{j}\right)] - \mathbb{E}\left[{\bar{\rho}}\left(\mathbf{X}_{i}, \mathbf{X}_{j}\right)\right]\Big].
  \end{align*}
  
  We can show that as $d \to \infty$, the first summand goes to $0$ in probability using part (a). Now,
  \begin{align*}
      & \sum_{i \neq j} \Big[\mathbb{E}[\hat{\bar{\rho}}\left(\mathbf{X}_{i}, \mathbf{X}_{j}\right)] - \mathbb{E}\left[{\bar{\rho}}\left(\mathbf{X}_{i}, \mathbf{X}_{j}\right)\right]\Big]\\
      &= \frac{1}{d}\sum_{k = 1}^d \Big[\mathbb{E}[\hat{{\rho}}\left({X}_{1k}, {X}_{2k}\right)] - \mathbb{E}\left[{{\rho}}\left({X}_{1k}, {X}_{2k}\right)\right]\Big]\\
      &= \frac{1}{d}\sum_{k = 1}^d \bigg[\frac{m}{m+n} \mathbb{E}\left[\rho_0(X_{1k},X_{2k};{X}_{3k})\right] + \frac{n}{m+n} \mathbb{E}\left[\rho_0(X_{1k},X_{2k};{Y}_{3k})\right] -\mathbb{E}\left[{{\rho}}\left({X}_{1k}, {X}_{2k}\right)\right]\bigg]\\
      &= \frac{1}{d}\sum_{k = 1}^d \bigg[\frac{m}{m+n} \mathbb{E}\left[\rho_0(X_{1k},X_{2k};{X}_{3k})\right] \\
      & \qquad\qquad+ \frac{n}{m+n} \mathbb{E}\left[\rho_0(X_{1k},X_{2k};{Y}_{3k})\right] - \mathbb{E}\left[E_{\mathbf{Q}}[{{\rho}_0}\left({X}_{1k}, {X}_{2k}; Q_k\right)\right]]\bigg]\\
      &\hspace{5cm}\text{ where } \mathbf{Q} \sim \frac{m}{m+n} \mathbf{F} + \frac{n}{m+n} \mathbf{G}\\
      &= \frac{1}{d}\sum_{k = 1}^d \bigg[\frac{m}{m+n} \mathbb{E}\left[\rho_0(X_{1k},X_{2k};{X}_{3k})\right] + \frac{n}{m+n} \mathbb{E}\left[\rho_0(X_{1k},X_{2k};{Y}_{3k})\right] - \mathbb{E}\left[{{\rho}}_0\left({X}_{1k}, {X}_{2k}; {Q}_k\right)\right]\bigg]\\
      &\begin{aligned}
      &= \frac{1}{d}\sum_{k = 1}^d \bigg[\frac{m}{m+n} \mathbb{E}\left[\rho_0(X_{1k},X_{2k};{X}_{3k})\right] + \frac{n}{m+n} \mathbb{E}\left[\rho_0(X_{1k},X_{2k};{Y}_{3k})\right] \\
      &\hspace{2cm} - \frac{m}{m+n} \mathbb{E}\left[{{\rho}}\left({X}_{1k}, {X}_{2k}; Q_k\right)|\mathbf{Q} \sim \mathbf{F}\right]-\frac{n}{m+n}\mathbb{E}\left[{{\rho}}\left({X}_{1k}, {X}_{2k}; Q_k\right)|\mathbf{Q} \sim \mathbf{G}\right]\bigg]
      \end{aligned}\\
      &= 0.
  \end{align*}
  
    \item[ii.] The proof is the same as above.
    \item[iii.] The proof is the same as above. 
    
  \item[\textbf{(c)}]  Once again, we shall use the result from part \textbf{(a)}. 
  \end{itemize}
  \begin{itemize}
  \item[i.] If $\mathbf{Z} \sim \mathbf{F}$, in that case, $T_{\mathbf{F F}}=\mathbb{E}\left[{\bar{\rho}}\left(\mathbf{X}_{1}, \mathbf{X}_{2}\right)\right]=\mathbb{E}[{\bar{\rho}}\left(\mathbf{X_i}, \mathbf{Z}\right)]$. So, \begin{align*}
      \hat{T}_{\mathbf{F}}(\mathbf{Z})-T_{\mathbf{F F}}
      &= \frac{1}{m}\sum_{i}(\hat{\bar{\rho}}\left(\mathbf{X_i}, \mathbf{Z}\right)-{\mathbb{E}[{\bar{\rho}}\left(\mathbf{X_i}, \mathbf{Z}\right)]})\\
      &=\frac{1}{m}\sum_{i}\big[(\hat{\bar{\rho}}\left(\mathbf{X_i}, \mathbf{Z}\right)-\mathbb{E}[\hat{\bar{\rho}}\left(\mathbf{X_i}, \mathbf{Z}\right)])+(\mathbb{E}[\hat{\bar{\rho}}\left(\mathbf{X_i},\mathbf{Z}\right)]-{\mathbb{E}[{\bar{\rho}}\left(\mathbf{X_i}, \mathbf{Z}\right)]})\big].
  \end{align*}
\begin{align*}
 &\frac{1}{m}\sum_{i}(\mathbb{E}[\hat{\bar{\rho}}\left(\mathbf{X_i},\mathbf{Z}\right)]-{\mathbb{E}[{\bar{\rho}}\left(\mathbf{X_i}, \mathbf{Z}\right)]})\\
 &= \mathbb{E}[\hat{\bar{\rho}}\left(\mathbf{X_1},\mathbf{Z}\right)]-{\mathbb{E}[{\bar{\rho}}\left(\mathbf{X_1}, \mathbf{Z}\right)]}\\
 &=\frac{1}{d}\sum_{k = 1}^d \bigg[\mathbb{E}[\hat{{\rho}}\left(\mathbf{X_1},\mathbf{Z}\right)]-{\mathbb{E}[{{\rho}}\left(\mathbf{X_1}, \mathbf{Z}\right)]}\bigg]\\
 &=\frac{1}{d}\sum_{k = 1}^d \bigg[\frac{m}{m+n} \mathbb{E}\left[\rho_0(X_{1k},Z_k;{X}_{2k})\right] + \frac{n}{m+n} \mathbb{E}\left[\rho_0(X_{1k},Z_{k};{Y}_{2k})\right]- \mathbb{E}\left[{{\rho}}\left({X}_{1k}, {Z}_{k}\right)\right]\bigg]\\
 &=\frac{1}{d}\sum_{k = 1}^d \bigg[\frac{m}{m+n} \mathbb{E}\left[\rho_0(X_{1k},Z_{k};{X}_{2k})\right] + \frac{n}{m+n} \mathbb{E}\left[\rho_0(X_{1k},Z_{k};{Y}_{2k})\right] -\\
      &\qquad\qquad\qquad \mathbb{E}\left[E_{{Q}_k}[{{\rho}_0}\left({X}_{1k}, {Z}_{k}; {Q}_k\right)\right]]\bigg] \text{, where } \mathbf{Q} \sim \frac{m}{m+n} \mathbf{F} + \frac{n}{m+n} \mathbf{G}\\
 &=\frac{1}{d}\sum_{k = 1}^d \bigg[\frac{m}{m+n} \mathbb{E}\left[\rho_0(X_{1k},Z_{k};{X}_{2k})\right] + \frac{n}{m+n} \mathbb{E}\left[\rho_0(X_{1k},Z_{k};{Y}_{2k})\right]- \mathbb{E}\left[{{\rho}}_0\left({X}_{1k}, {Z}_{k};{Q}_k\right)\right]\bigg]\\
 &\begin{aligned}
      &=\frac{1}{d} \sum_{k = 1}^d \bigg[\frac{m}{m+n} \mathbb{E}\left[\rho_0(X_{1k},Z_{k};{X}_{2k})\right] + \frac{n}{m+n} \mathbb{E}\left[\rho_0(X_{1k},Z_{k};{Y}_{2k})\right] \\
      &\qquad - \frac{m}{m+n} \mathbb{E}\left[{{\rho}}\left({X}_{1k}, {Z}_{k}; {Q}_k\right)|\mathbf{Q} \sim \mathbf{F}\right]-\frac{n}{m+n} \mathbb{E}\left[{{\rho}}\left({X}_{1k}, {Z}_{k}; {Q}_k\right)|\mathbf{Q} \sim \mathbf{G}\right]\bigg]
      \end{aligned}\\
&= 0
\qed
\end{align*} 
\end{itemize}

\noindent\textbf{Proof of Theorem 3}\\\\
We shall prove part (a) only, and the next part will follow analogously.
  \begin{align*}
  |\mathscr{D}_1(\mathbf{Z})-\frac{1}{2}{\overline{\mathcal{W}}}^{*}_\mathbf{FG}|
  &= \bigg|(\hat{T}_{\mathbf{G}}(\mathbf{Z})-{T}_{\mathbf{F G}})-(\hat{T}_{\mathbf{F}}(\mathbf{Z})-{T}_{\mathbf{F F}})-\frac{1}{2}(\hat{T}_{\mathbf{G G}}-{T}_{\mathbf{G G}})+\frac{1}{2}(\hat{T}_{\mathbf{F F}}-{T}_{\mathbf{F F}})\bigg|\\
  &\leq |\hat{T}_{\mathbf{G}}(\mathbf{Z})-{T}_{\mathbf{F G}}|+|\hat{T}_{\mathbf{F}}(\mathbf{Z})-{T}_{\mathbf{F F}}|+\frac{1}{2}|\hat{T}_{\mathbf{G G}}-{T}_{\mathbf{G G}}|+\frac{1}{2}|\hat{T}_{\mathbf{F F}}-{T}_{\mathbf{F F}}|\\
&\stackrel{\mathbb{P}}{\rightarrow} 0 \text{,  as } d \to \infty, \text{ given }\mathbf{Z} \sim \mathbf{F}.
  \end{align*}
Here, the last assertion follows from \Cref{lem:A1}. Now,
\begin{align*}
&\bigg|S(\mathbf{Z})-\frac{1}{2}\left(T_{\mathbf{F F}}-T_{\mathbf{G G}}\right)\bigg|\\
&=\bigg|\left(\hat{T}_{\mathbf{F}}(\mathbf{Z})-\hat{T}_{\mathbf{F F}}\right)+\left(\hat{T}_{\mathbf{G}}(\mathbf{Z})-\hat{T}_{\mathbf{F G}}\right)+\frac{1}{2}\left(\hat{T}_{\mathbf{F F}}-T_{\mathbf{F F}}\right)-\frac{1}{2}\left(\hat{T}_{\mathbf{G G}}-T_{\mathbf{G G}}\right)\bigg|\\
&\leq |\hat{T}_{\mathbf{F}}(\mathbf{Z})-\hat{T}_{\mathbf{F F}}|+|\hat{T}_{\mathbf{G}}(\mathbf{Z})-\hat{T}_{\mathbf{F G}}|+\frac{1}{2}|\hat{T}_{\mathbf{F F}}-T_{\mathbf{F F}}|+\frac{1}{2}|\hat{T}_{\mathbf{G G}}-T_{\mathbf{G G}}|.
\end{align*}
From here, it follows from \Cref{lem:A1} that
\begin{equation}
\label{eq:4}
\bigg|S(\mathbf{Z})-\frac{1}{2}\left(T_{\mathbf{F F}}-T_{\mathbf{G G}}\right)\bigg|  \stackrel{\mathbb{P}}{\rightarrow} 0 \text{,  as } d \to \infty, \text{ given }\mathbf{Z} \sim \mathbf{F}.
\end{equation}
Also, from theorem 1, given $\mathbf{Z} \sim \mathbf{F}$,

$$
\bigg|\mathscr{D}_1(\mathbf{Z})-\frac{1}{2} {\overline{\mathcal{W}}}^{*}_\mathbf{FG} \bigg|\stackrel{\mathbb{P}}{\rightarrow} 0 \text{,  as } d \to \infty.
$$
Next, we have: \begin{align*}
    |\hat{\overline{\mathcal{W}}}^{*}_\mathbf{FG}-{\overline{\mathcal{W}}}^{*}_\mathbf{FG}|&=\bigg|2\left(\hat{T}_{\mathbf{F G}}-T_{\mathbf{F G}}\right)-\left(\hat{T}_{\mathbf{F F}}-T_{\mathbf{F F}}\right)-\left(\hat{T}_{\mathbf{G G}}-T_{\mathbf{G G}}\right)\bigg| \\
    &\leq 2|\hat{T}_{\mathbf{F G}}-T_{\mathbf{F G}}|+|\hat{T}_{\mathbf{F F}}-T_{\mathbf{F F}}|+|\hat{T}_{\mathbf{G G}}-T_{\mathbf{G G}}| \\
    &\stackrel{\mathbb{P}}{\rightarrow} 0 \text{,  as } d \to \infty.
\end{align*}
And also:
\begin{align*}
\bigg|S_{\mathbf{F G}}-\left(T_{\mathbf{F F}}-T_{\mathbf{G G}}\right)\bigg|&=\bigg|\left(\hat{T}_{\mathbf{F F}}-T_{\mathbf{F F}}\right)-\left(\hat{T}_{\mathbf{G G}}-T_{\mathbf{G G}}\right)\bigg|\\
&\leq |\hat{T}_{\mathbf{F F}}-T_{\mathbf{F F}}|+|\hat{T}_{\mathbf{G G}}-T_{\mathbf{G G}}| \stackrel{\mathbb{P}}{\rightarrow} 0\text{,  as } d \to \infty.
\end{align*}

\noindent
Combining the 4 results stated above, we can see that given $\mathbf{Z} \sim \mathbf{F}$,

$$
\left|\frac{1}{2} \hat{\overline{\mathcal{W}}}^{*}_\mathbf{FG}\mathscr{D}_1(\mathbf{Z})+\frac{1}{2} S_{\mathbf{F G}} \cdot S(\mathbf{Z})-\bar{\tau}_{\mathbf{FG}}\right| \stackrel{\mathbb{P}}{\rightarrow} 0 \text{,  as } d \to \infty.
$$
\qed

\noindent\textbf{Proof of Theorem 4}\\\\
We shall prove part (a) only, and the next part will follow analogously.
We have already shown that given $\mathbf{Z} \sim \mathbf{F}$, 
$$\left|\mathscr{D}_1(\mathbf{Z})-\frac{1}{2}\overline{\mathcal{W}}_{\mathbf{FG}}^{*}\right|
    \stackrel{\mathbb{P}}{\rightarrow} 0 \text{,  as } d \to\infty.$$
Let, $\liminf_{d} \overline{\mathcal{W}}_{\mathbf{FG}}^{*}= \delta$. By $\delta > 0$. So, there exists some $d_{0} \in \mathbb{N}$, such that ${\overline{\mathcal{W}}}_{\mathbf{FG}}^{*} > \frac{\delta}{2}, $ for all $d\geq d_0.$\\
We have,  $\mathbb{P}\left[\left|\mathscr{D}_1(\mathbf{Z})-\frac{1}{2}\overline{\mathcal{W}}_{\mathbf{FG}}^{*}\right|<\frac{\delta}{4}\mid \mathbf{Z} \sim \mathbf{F}\right]\to 1 \text{,  as } d \to\infty$. For all $d\geq d_0$, we have
\begin{align*}
    \mathbb{P}\left[ \mathscr{D}_1(\mathbf{Z}) > 0\mid \mathbf{Z} \sim \mathbf{F}\right] 
    &\geq \mathbb{P}\left[ \mathscr{D}_1(\mathbf{Z}) > \frac{1}{2}\overline{\mathcal{W}}_{\mathbf{FG}}^{*}-\frac{\delta}{4}\mid \mathbf{Z} \sim \mathbf{F}\right]\\
    &\geq \mathbb{P}\left[| \mathscr{D}_1(\mathbf{Z})- \frac{1}{2}\overline{\mathcal{W}}_{\mathbf{FG}}^{*}|<\frac{\delta}{4}\mid \mathbf{Z} \sim \mathbf{F}\right]\\
    &\implies \mathbb{P}\left[ \mathscr{D}_1(\mathbf{Z}) > 0\mid \mathbf{Z} \sim \mathbf{F}\right] \to 1 \text{,  as } d \to\infty.
\end{align*}

\noindent
Given $\mathbf{Z} \sim \mathbf{F}$,
$$ \operatorname{sign}\left(\mathscr{D}_1(\mathbf{Z})\right) \stackrel{\mathbb{P}}{\rightarrow} 1 \text{ ,  as } d \to\infty$$
Also, as we have seen before: $$\hat{\overline{\mathcal{W}}}_{\mathbf{FG}}^{*} - \overline{\mathcal{W}}_{\mathbf{FG}}^{*} = 2\left(\hat{T}_{\mathbf{F G}}-{T}_{\mathbf{F G}}\right) - \left(\hat{T}_{\mathbf{F F}}-{T}_{\mathbf{F F}}\right) - \left(\hat{T}_{\mathbf{G G}}-{T}_{\mathbf{G G}}\right) \stackrel{\mathbb{P}}{\rightarrow} 0\text{ ,  as } d \to\infty.$$
Hence, we have 
\begin{equation}
\label{eq:thm3.1}
   \frac{1}{2} \hat{\overline{\mathcal{W}}}_{\mathbf{FG}}^{*} \operatorname{sign}\left(\mathscr{D}_1(\mathbf{Z})\right) -\frac{1}{2} \overline{\mathcal{W}}_{\mathbf{FG}}^{*}\stackrel{\mathbb{P}}{\rightarrow}0 \text{ ,  as } d \to\infty.
\end{equation}
Now, Assumption 6 implies, for any fixed $\epsilon >0$, we can choose some $d_1$, such that $\left|T_{\mathbf{F F}}-T_{\mathbf{G G}}\right| > \epsilon$, $ \forall d\geq d_1$. From \eqref{eq:4},
\begin{equation}
\label{eq:thm3.2}
 \mathbb{P}\left[~\left|S(\mathbf{Z})-\frac{1}{2}\left({T}_{\mathbf{F F}}-{T}_{\mathbf{G G}}\right)\right|\leq \epsilon/4 \mid \mathbf{Z} \sim \mathbf{F}\right]
\to1   .
\end{equation}
So, if $\text{sign}({T}_{\mathbf{F F}}-{T}_{\mathbf{G G}})=1$, we have, $T_{\mathbf{F F}}-T_{\mathbf{G G}} > \epsilon$, $ \forall d\geq d_0$. Then, \\
$\left|S(\mathbf{Z})-\frac{1}{2}\left({T}_{\mathbf{F F}}-{T}_{\mathbf{G G}}\right)\right|\leq \epsilon/4$ implies that $S(\mathbf{Z})>\frac{3\epsilon}{4}$ 
and so, $\text{sign}(S(\mathbf{Z}))=1$\\
Similarly, if $\text{sign}({T}_{\mathbf{F F}}-{T}_{\mathbf{G G}})=-1$, we can show that $\left|S(\mathbf{Z})-\frac{1}{2}\left({T}_{\mathbf{F F}}-{T}_{\mathbf{G G}}\right)\right|\leq \epsilon/4$ implies $\text{sign}(S(\mathbf{Z}))=-1$. Hence, for all $d\geq d_1$,
  \begin{align}
  \label{eq:thm3.3}
    &\mathbb{P}\bigg[\text{sign}(S(\mathbf{Z}))-\text{sign}({T}_{\mathbf{F F}}-{T}_{\mathbf{G G}})= 0\mid \mathbf{Z} \sim \mathbf{F}\bigg] \geq \nonumber \\
    &\hspace{2cm}\mathbb{P}\left[~\left|S(\mathbf{Z})-\frac{1}{2}\left({T}_{\mathbf{F F}}-{T}_{\mathbf{G G}}\right)\right|\leq \epsilon/4 \mid \mathbf{Z} \sim \mathbf{F}\right] .\stepcounter{equation}\tag{\theequation}
  \end{align}
  From \eqref{eq:thm3.2} and \eqref{eq:thm3.3} we get, $ \mathbb{P}\left[\text{sign}(S(\mathbf{Z}))-\text{sign}({T}_{\mathbf{F F}}-{T}_{\mathbf{G G}})= 0 \mid \mathbf{Z} \sim \mathbf{F}\right]
\to1.$\\
$$\implies \text{sign}(S(\mathbf{Z}))-\text{sign}({T}_{\mathbf{F F}}-{T}_{\mathbf{G G}})\stackrel{\mathbb{P}}{\rightarrow}0.$$
And also, $$\hat S_{\mathbf{F G}} - \left(T_{\mathbf{F F}}-T_{\mathbf{G G}}\right) = \left(\hat{T}_{\mathbf{F F}} - T_{\mathbf{F F}}\right)-\left(\hat{T}_{\mathbf{G G}} - T_{\mathbf{GG}}\right) \stackrel{\mathbb{P}}{\rightarrow} 0\text{ ,  as } d \to\infty.$$
  Hence, given $\mathbf{Z} \sim \mathbf{F}$,
\begin{equation}
\label{eq:thm3.4}
  \frac{1}{2} \hat S_{\mathbf{F G}} \operatorname{sign}(S(\mathbf{Z}))-\frac{1}{2}\left|T_{\mathbf{F F}}-T_{\mathbf{G G}}\right|\stackrel{\mathbb{P}}{\rightarrow} 0\text{ ,  as } d \to\infty  .
\end{equation}
From \eqref{eq:thm3.1} and \eqref{eq:thm3.4}, we obtain that given $\mathbf{Z} \sim \mathbf{F}$, 
\begin{equation*}
    \left|\frac{1}{2} \hat{\overline{\mathcal{W}}}_{\mathbf{FG}}^{*} \operatorname{sign}\left(\mathscr{D}_1(\mathbf{Z})\right)+\frac{1}{2} \hat S_{\mathbf{F G}} \operatorname{sign}(S(\mathbf{Z}))-\bar{\psi}_{\mathbf{FG}}\right|
\stackrel{\mathbb{P}}{\rightarrow} 0 \text{ ,  as } d \to\infty.
\end{equation*}\qed\\
\noindent\textbf{Proof of Theorem 5}
\begin{itemize}
    \item[\textbf{(1)~}] The misclassification probability of the classifier $\delta_{1}$ can be written as
$$
\begin{aligned}
\Delta_1  &=\mathbb{P}\left[\delta_{1}(\mathbf{Z})=2, \mathbf{Z} \sim \mathbf{F}\right]+\mathbb{P}\left[\delta_{1}(\mathbf{Z})=1, \mathbf{Z} \sim \mathbf{G}\right] \\
&=\frac{m}{m+n} \mathbb{P}\left[\delta_{1}(\mathbf{Z})=2 \mid \mathbf{Z} \sim \mathbf{F}\right]+\frac{n}{m+n} \mathbb{P}\left[\delta_{1}(\mathbf{Z})=1 \mid \mathbf{Z} \sim \mathbf{G}\right] \\
&=\frac{m}{m+n} \mathbb{P}\left[\mathscr{D}_1(\mathbf{Z}) \leq 0 \mid \mathbf{Z} \sim \mathbf{F}\right]+\frac{n}{m+n}
\mathbb{P}\left[\mathscr{D}_1(\mathbf{Z})>0 \mid \mathbf{Z} \sim \mathbf{G}\right].
\end{aligned}
$$

Since  $\liminf_{d} \overline{\mathcal{W}}_{d}^{*} > 0$ (Assumption 5), we can choose $\epsilon>0$ such that $\epsilon<\overline{\mathcal{W}}^{*}$ for all $d \geq d_{0}$ for some $d_{0} \in \mathbb{N}$. Therefore, we have:
$$
\begin{aligned}
\mathbb{P}\left[\mathscr{D}_1(\mathbf{Z}) \leq 0 \mid \mathbf{Z} \sim \mathbf{F}\right] 
& \leq \mathbb{P}\left[\mathscr{D}_1(\mathbf{Z}) \leq \overline{\mathcal{W}}_{d}^{*}-\epsilon \mid \mathbf{Z} \sim \mathbf{F}\right] \\
& \leq \mathbb{P}\left[\mathscr{D}_1(\mathbf{Z})-\overline{\mathcal{W}}_{d}^{*} \leq-\epsilon \mid \mathbf{Z} \sim \mathbf{F}\right] \\
& \leq \mathbb{P}\left[\left|\mathscr{D}_1(\mathbf{Z})-\overline{\mathcal{W}}_{d}^{*}\right|>\epsilon \mid \mathbf{Z} \sim \mathbf{F}\right] ~\to0 \text{~, as~}d \to \infty.
\end{aligned}
$$
Similarly, one can show that $$\mathbb{P}\left[\mathscr{D}_1(\mathbf{Z})> 0 \mid \mathbf{Z} \sim \mathbf{G}\right] \to0 \text{ , as } d \to \infty.$$

Thus, we conclude that $\Delta_1 \to 0$ as $d \to \infty$.

\item[\textbf{(2)~}] The misclassification probability of the classifier $\delta_{2}$ can be written as
$$
\begin{aligned}
\Delta_2  &=\mathbb{P}\left[\delta_{2}(\mathbf{Z})=2, \mathbf{Z} \sim \mathbf{F}\right]+\mathbb{P}\left[\delta_{2}(\mathbf{Z})=1, \mathbf{Z} \sim \mathbf{G}\right] \\
&=\frac{m}{m+n} \mathbb{P}\left[\delta_{2}(\mathbf{Z})=2 \mid \mathbf{Z} \sim \mathbf{F}\right]+\frac{n}{m+n} \mathbb{P}\left[\delta_{2}(\mathbf{Z})=1 \mid \mathbf{Z} \sim \mathbf{G}\right] \\
&=\frac{m}{m+n} \mathbb{P}\left[\hat{\overline{\mathcal{W}}}^{*}_\mathbf{FG}\mathscr{D}_1(\mathbf{Z}) + \hat S_{\mathbf{F G}}\cdot S(\mathbf{Z}) \leq 0 \mid \mathbf{Z} \sim \mathbf{F}\right]\\
& \qquad +\frac{n}{m+n}
\mathbb{P}\left[\hat{\overline{\mathcal{W}}}^{*}_\mathbf{FG}\mathscr{D}_1(\mathbf{Z}) + \hat S_{\mathbf{F G}}\cdot S(\mathbf{Z}) > 0 \mid \mathbf{Z} \sim \mathbf{G}\right].
\end{aligned}
$$
Assumption 5 implies, we have: $$\liminf_{d} \bar{\tau} = \liminf_{d} \left[\frac{1}{2}{\overline{\mathcal{W}}}^{*2}_{\mathbf{FG}}+\frac{1}{2}\left(T_{\mathbf{F F}}-T_{\mathbf{G G}}\right)^2\right] \geq \frac{1}{2}\liminf_{d} {\overline{\mathcal{W}}}^{*2}_\mathbf{FG}> 0.$$

From here onward, we can proceed exactly similar to the proof of part \textbf{(1)}, and we can show that $\Delta_2 \to 0$ as $d \to \infty$. 

\item[\textbf{(3)~}] The misclassification probability of the classifier $\delta_3$ can be written as
\begin{align*}
\Delta_3  &=\mathbb{P}\left[\delta_{3}(\mathbf{Z})=2, \mathbf{Z} \sim \mathbf{F}\right]+\mathbb{P}\left[\delta_{3}(\mathbf{Z})=1, \mathbf{Z} \sim \mathbf{G}\right] \\
&=\frac{m}{m+n} \mathbb{P}\left[\delta_{3}(\mathbf{Z})=2 \mid \mathbf{Z} \sim \mathbf{F}\right]+\frac{n}{m+n} \mathbb{P}\left[\delta_{3}(\mathbf{Z})=1 \mid \mathbf{Z} \sim \mathbf{G}\right].\\
&\begin{aligned}
&=\frac{m}{m+n} \mathbb{P}\left[\hat{\overline{\mathcal{W}}}^{*}_\mathbf{FG} \operatorname{sign}\mathscr{D}_1(\mathbf{Z})+\hat S_{\mathbf{F G}} \operatorname{sign}S(\mathbf{Z}) \leq 0 \mid \mathbf{Z} \sim \mathbf{F}\right]\\
& \qquad +\frac{n}{m+n}
\mathbb{P}\left[\hat{\overline{\mathcal{W}}}^{*}_\mathbf{FG}\operatorname{sign}\mathscr{D}_1(\mathbf{Z})+\hat S_{\mathbf{F G}} \operatorname{sign}S(\mathbf{Z})> 0 \mid \mathbf{Z} \sim \mathbf{G}\right].
\end{aligned}
\end{align*}
Assumption 5 implies, we have: $$\liminf_{d} \bar{\psi} = \liminf_{d} \left[\frac{1}{2}{{\overline{\mathcal{W}}}^{*}_\mathbf{FG}}+\frac{1}{2}|T_{\mathbf{F F}}-T_{\mathbf{G G}}|\right] \geq \frac{1}{2}\liminf_{d} {{\overline{\mathcal{W}}}^{*}_\mathbf{FG}}> 0.$$
The argument for rest of the proof is similar to what has been shown for part \textbf{(1)}. Finally, we shall conclude that $\Delta_3 \to 0$ as $d \to \infty$. 
\end{itemize}
\qed

\begin{lemmaA}
    \label{lem:A2}
 Suppose  assumptions 6 and 8 are satisfied.
\begin{itemize}
    \item[(a)] If $\liminf_d\left(\max \{ T_{\mathbf{F F}}, T_{\mathbf{G G}}\}-T_{\mathbf{F G}}\right)>0$
there exists $d_0'\in \mathbb N$, 
such that  $\Delta_2 \leq \Delta_1$ for all $d \geq d_0'$.
    \item[(b)] If $\liminf_d\left(T_{\mathbf{F G}}-\max \{ T_{\mathbf{F F}}, T_{\mathbf{F G}}\}\right)>0$, 
there exists $d_0'\in \mathbb N$, such that $\Delta_2 \geq \Delta_1$ for all $d \geq d_0'$.
\end{itemize}
\end{lemmaA}

\noindent\textbf{Proof of \Cref{lem:A2}}
\begin{itemize}
\item [(a)] Since, ${\overline{\mathcal{W}}}^{*}_{\mathbf{FG}}-\hat{\overline{\mathcal{W}}}^{*}_\mathbf{FG}\stackrel{\mathbb{P}}{\rightarrow} 0$ as $d \rightarrow \infty$. Therefore, for any $\epsilon_{0}>0$ and $\delta>0$, there exists a $d_0$ such that for all $d \geq d_0$,
\begin{align*}
\mathbb{P}\left[\left|\hat{\overline{\mathcal{W}}}^{*}_{\mathbf{FG}}-{\overline{\mathcal{W}}}^{*}_\mathbf{FG}\right|>\delta \right]<\epsilon_{0}
&\implies \mathbb{P}\left[\hat{\overline{\mathcal{W}}}^{*}_{\mathbf{FG}}-{\overline{\mathcal{W}}}^{*}_\mathbf{FG} <-\delta \right]<\epsilon_{0} \\
&\implies \mathbb{P}\left[\hat{\overline{\mathcal{W}}}^{*}_{\mathbf{FG}}<{\overline{\mathcal{W}}}^{*}_\mathbf{FG}-\delta\right]<\epsilon_{0}.
\end{align*}

Using Assumption 5, we have $\lambda_0=\liminf_d {\overline{\mathcal{W}}}^{*}_\mathbf{FG} >0 $. Hence, for any $0<\delta<\lambda_{0}$, 
\begin{equation}
\label{eq:w}
\mathbb{P}[\hat{\overline{\mathcal{W}}}^{*}_{\mathbf{FG}}<0] < \epsilon_0,  \text { for all } d\geq d_0. 
\end{equation}
 WLOG, we assume that ${T}_{\mathbf{F F}}>{T}_{\mathbf{G G}}$. 
 Since, $\liminf_d\left(\max \{ T_{\mathbf{F F}}, T_{\mathbf{G G}}\}-T_{\mathbf{F G}}\right)>0$,
 there exists $d' \in \mathbb N$ such that $  T_{\mathbf{F F}} > T_{\mathbf{F G}} $ for all $d\geq d^{'}$.
 Also, ${\overline{\mathcal{W}}}^{*}_\mathbf{FG}=2{T}_{\mathbf{F G}}-{T}_{\mathbf{F F}}-{T}_{\mathbf{G G}}>0$.
 So, we have, ${T}_{\mathbf{F G}}>{T}_{\mathbf{G G}}$ for all $d\geq d^{'}$. Hence,
$$
{\overline{\mathcal{W}}}^{*}_\mathbf{FG}=\left({T}_{\mathbf{F G}}-{T}_{\mathbf{F F}}\right)+\left({T}_{\mathbf{F G}}-{T}_{\mathbf{G G}}\right) < {T}_{\mathbf{F F}}-{T}_{\mathbf{G G}}= S_{\mathbf{F G}} 
$$
Since $\hat{\overline{\mathcal{W}}}^{*}_{\mathbf{FG}}-\hat S_{\mathbf{F G}}-\left({\overline{\mathcal{W}}}^{*}_\mathbf{FG}-S_{\mathbf{F G}}\right)\stackrel{\mathbb{P}}{\rightarrow} 0$ as $d \rightarrow \infty$ and $\liminf_d\left(S_{\mathbf{F G}}-{\overline{\mathcal{W}}}^{*}_\mathbf{FG}\right)=\liminf_d 2\left({T}_{\mathbf{F F}}-{T}_{\mathbf{F G}}\right)>0$, in a similar way, we can show that for every $\epsilon_1>1$, there exists a $d_1$, such that 
\begin{equation}
\label{eq:w-s}
\mathbb{P}[\hat{\overline{\mathcal{W}}}^{*}_{\mathbf{FG}}>\hat S_{\mathbf{F G}}] < \epsilon_1 \text { for all } d \geq d_1
\end{equation}
Now, it follows from \eqref{eq:4} that for $\left.\mathbf{Z} \sim \mathbf{F}, \mid S(\mathbf{Z})-\frac{1}{2}\left(T_{\mathbf{F F}}-T_{\mathbf{G G}}\right)\right\} \mid \stackrel{\mathbb{P}}{\rightarrow} 0$ as $d \rightarrow \infty$. 
We have already assumed that ${T}_{\mathbf{F F}}>{T}_{\mathbf{G G}}$ . Define $\lambda_{0}=\liminf _{d} \frac{1}{2}\left({T}_{\mathbf{FF}}-{T}_{\mathbf{\mathbf{G G}}}\right)$. \\\\
Since ${\overline{\mathcal{W}}}^{*}_\mathbf{FG} < {T}_{\mathbf{FF}}-{T}_{\mathbf{GG}}$, $\liminf _{d} \left({T}_{\mathbf{FF}}-{T}_{\mathbf{GG}}\right) \geq \liminf _{d}{\overline{\mathcal{W}}}^{*}_\mathbf{FG}>0 $, by Assumption 5. Hence, we have $\lambda_0>0$.
Following similar arguments, one can show that for any $\epsilon_2>0$, there exists $d_2$, such that
\begin{equation}
\label{eq:sz}
 \mathbb{P}[{S}(\mathbf{Z}) < 0 \mid \mathbf{Z} \sim \mathbf{F}]<\epsilon_{2} \text { for all } d \geq d_2   
\end{equation}
Recall that
$$\Delta_{1}=\frac{m}{m+n} \mathbb{P}\left[\mathscr{D}_1(\mathbf{Z}) \leq 0 \mid \mathbf{Z} \sim \mathbf{F}\right]+\frac{n}{m+n} \mathbb{P}\left[\mathscr{D}_1(\mathbf{Z})>0 \mid \mathbf{Z} \sim \mathbf{G}\right]$$
and
$$
\begin{aligned}
\Delta_2 &=\frac{m}{m+n} \mathbb{P}\left[\hat{\overline{\mathcal{W}}}^{*}_\mathbf{FG}\mathscr{D}_1(\mathbf{Z}) + \hat S_{\mathbf{F G}}\cdot S(\mathbf{Z}) \leq 0 \mid \mathbf{Z} \sim \mathbf{F}\right]\\
& \qquad +\frac{n}{m+n}
\mathbb{P}\left[\hat{\overline{\mathcal{W}}}^{*}_\mathbf{FG}\mathscr{D}_1(\mathbf{Z}) + \hat S_{\mathbf{F G}}\cdot S(\mathbf{Z}) > 0 \mid \mathbf{Z} \sim \mathbf{G}\right]
\end{aligned}
$$
It follows that 
\begingroup
\allowdisplaybreaks
\begin{align*}
&\mathbb{P}\left[\hat{\overline{\mathcal{W}}}^{*}_{\mathbf{FG}}\mathscr{D}_1(\mathbf{Z})+\hat S_{\mathbf{F G}} S(\mathbf{Z}) \leq 0 \mid \mathbf{Z} \sim \mathbf{F}\right] \\
&=\mathbb{P}\left[\hat{\overline{\mathcal{W}}}^{*}_{\mathbf{FG}}\mathscr{D}_1(\mathbf{Z})+\hat S_{\mathbf{F G}} S(\mathbf{Z}) \leq 0, S(\mathbf{Z}) \geq 0 \mid \mathbf{Z} \sim \mathbf{F}\right]\\
& \qquad +
\mathbb{P}\left[\hat{\overline{\mathcal{W}}}^{*}_{\mathbf{FG}}\mathscr{D}_1(\mathbf{Z})+\hat S_{\mathbf{F G}} S(\mathbf{Z}) > 0, S(\mathbf{Z}) < 0 \mid \mathbf{Z} \sim \mathbf{F}\right] \\
&\leq \mathbb{P}\left[\hat{\overline{\mathcal{W}}}^{*}_{\mathbf{FG}}\mathscr{D}_1(\mathbf{Z})+\hat S_{\mathbf{F G}} S(\mathbf{Z})  \leq 0, S(\mathbf{Z})\geq 0, \hat{\overline{\mathcal{W}}}^{*}_{\mathbf{FG}}\leq \hat S_{\mathbf{F G}} \mid \mathbf{Z} \sim \mathbf{F}\right]\\
&\qquad +\mathbb{P}\left[\hat{\overline{\mathcal{W}}}^{*}_{\mathbf{FG}}\mathscr{D}_1(\mathbf{Z})+\hat S_{\mathbf{F G}} S(\mathbf{Z})  \leq 0, S(\mathbf{Z})\geq 0, \hat{\overline{\mathcal{W}}}^{*}_{\mathbf{FG}}>\hat S_{\mathbf{F G}} \mid \mathbf{Z} \sim \mathbf{F}\right] + \mathbb{P}[S(\mathbf{Z})<0 \mid \mathbf{Z} \sim \mathbf{F}] \\
&\leq \mathbb{P}\left[\hat{\overline{\mathcal{W}}}^{*}_{\mathbf{FG}}\left\{\mathscr{D}_1(\mathbf{Z})+S(\mathbf{Z})\right\} \leq 0, S(\mathbf{Z})\geq 0 \mid \mathbf{Z} \sim \mathbf{F}\right] +\mathbb{P}\left[  \hat{\overline{\mathcal{W}}}^{*}_{\mathbf{FG}}>\hat S_{\mathbf{F G}} \right]+ \epsilon_2, \text { for all } d \geq d_2\\
&\leq \mathbb{P}\left[\hat{\overline{\mathcal{W}}}^{*}_{\mathbf{FG}}\left\{\mathscr{D}_1(\mathbf{Z})+S(\mathbf{Z})\right\} \leq 0, S(\mathbf{Z})\geq 0,\hat{\overline{\mathcal{W}}}^{*}_{\mathbf{FG}}\geq 0 \mid \mathbf{Z} \sim \mathbf{F}\right]\\
& \qquad  + \mathbb{P}\left[\hat{\overline{\mathcal{W}}}^{*}_{\mathbf{FG}}< 0 \right] + \epsilon_1 + \epsilon_2, \text { for all } d \geq \max\{d_1,d_2\}\\
&\leq \mathbb{P}\left[\mathscr{D}_1(\mathbf{Z})+S(\mathbf{Z}) \leq 0, S(\mathbf{Z}) \geq 0  \mid \mathbf{Z} \sim \mathbf{F}\right] + \epsilon_0 +\epsilon_1+ \epsilon_2, \text { for all } d \geq \max\{d_0,d_1,d_2\}\\
&\leq \mathbb{P}\left[\mathscr{D}_1(\mathbf{Z}) \leq 0  \mid \mathbf{Z} \sim \mathbf{F}\right] + \epsilon_0 +\epsilon_1+ \epsilon_2, \text { for all } d \geq \max\{d_{0},d_1,d_2\}
\end{align*}
\endgroup

Similarly, one can show that, for all $d \geq \max\{d_{0},d_1,d_2\}$
\begin{equation*}
\mathbb{P}\left[\hat{\overline{\mathcal{W}}}^{*}_{\mathbf{FG}}\mathscr{D}_1(\mathbf{Z})+\hat S_{\mathbf{F G}} S(\mathbf{Z})>0 \mid \mathbf{Z} \sim \mathbf{G}\right]\leq \mathbb{P}\left[\mathscr{D}_1(\mathbf{Z})>0 \mid \mathbf{Z} \sim \mathbf{G}\right] + \epsilon_0 +\epsilon_1 +\epsilon_2.
\end{equation*}
Adding the two inequalities, we obtain
$$
\Delta_{2} \leq \Delta_{1}+\epsilon_0 +\epsilon_1+\epsilon_2 \text { for all } d \geq \max\{d_{0},d_1,d_2\},
$$
Let, $d_{0}^{\prime}=\max \left\{ d_{0},d_1,d_2\right\}$. Since, $\epsilon_{0},\epsilon_{1},\epsilon_{2}>0$ are arbitrary, we have $\Delta_{2} \leq \Delta_{1}$ for all $d \geq d_{0}^{\prime}$.

\item[(b)] We have already proved  that for every $\epsilon_1>0$, there exists a $d_1$, such that $\mathbb{P}[\hat{\overline{\mathcal{W}}}^{*}_{\mathbf{FG}}<0] < \epsilon_1$,  for all $d\geq d_1$.
$$
\begin{aligned}
&\mathbb{P}\left[\mathscr{D}_1(\mathbf{Z}) \leq 0 \mid \mathbf{Z} \sim \mathbf{F}\right]\\
&=\mathbb{P}\left[\mathscr{D}_1(\mathbf{Z}) \leq 0 ,\hat{\overline{\mathcal{W}}}^{*}_{\mathbf{FG}}\geq 0\mid \mathbf{Z} \sim \mathbf{F}\right]+\mathbb{P}\left[\mathscr{D}_1(\mathbf{Z}) \leq 0,\hat{\overline{\mathcal{W}}}^{*}_{\mathbf{FG}}< 0 \mid \mathbf{Z}\sim \mathbf{F}\right]\\
&\leq \mathbb{P}\left[\hat{\overline{\mathcal{W}}}^{*}_{\mathbf{FG}}\mathscr{D}_1(\mathbf{Z}) \leq 0 \mid \mathbf{Z} \sim \mathbf{F}\right]+\mathbb{P}\left[\hat{\overline{\mathcal{W}}}^{*}_{\mathbf{FG}}< 0 \mid \mathbf{Z}\sim \mathbf{F}\right]\\
&\leq \mathbb{P}\left[\hat{\overline{\mathcal{W}}}^{*}_{\mathbf{FG}}\mathscr{D}_1(\mathbf{Z})\leq 0 ,\hat S_{\mathbf{F G}}S(\mathbf{Z})\leq 0\mid \mathbf{Z} \sim \mathbf{F}\right]\\
& \qquad +\mathbb{P}\left[\hat{\overline{\mathcal{W}}}^{*}_{\mathbf{FG}}\mathscr{D}_1(\mathbf{Z}) \leq 0,\hat S_{\mathbf{F G}}S(\mathbf{Z})> 0 \mid \mathbf{Z}\sim \mathbf{F}\right]+\epsilon_1\\
&\leq \mathbb{P}\left[\hat{\overline{\mathcal{W}}}^{*}_{\mathbf{FG}}\mathscr{D}_1(\mathbf{Z})+\hat S_{\mathbf{F G}}S(\mathbf{Z})\leq 0\mid \mathbf{Z} \sim \mathbf{F}\right]\\
& \qquad +\mathbb{P}\left[\hat{\overline{\mathcal{W}}}^{*}_{\mathbf{FG}}\mathscr{D}_1(\mathbf{Z})-\hat S_{\mathbf{F G}}S(\mathbf{Z})< 0\mid \mathbf{Z} \sim \mathbf{F}\right]+\epsilon_1\\
\end{aligned}
$$
Now, we know, if $\mathbf{Z} \sim \mathbf{F}$,$$\left|\mathscr{D}_1(\mathbf{Z})-\frac{1}{2}\overline{\mathcal{W}}_{\mathbf{F G}}^{*}\right|
    \stackrel{\mathbb{P}}{\rightarrow} 0 \text{ ,  as } d \to\infty$$
$$\left|S(\mathbf{Z})-\frac{1}{2}\left(T_{\mathbf{F F}}-T_{\mathbf{G G}}\right)\right| \stackrel{\mathbb{P}}{\rightarrow} 0 \text{ ,  as } d \to\infty$$
Also, we know,$$\left|\hat S_{\mathbf{F G}}-\left(T_{\mathbf{F F}}-T_{\mathbf{G G}}\right)\right| \stackrel{\mathbb{P}}{\rightarrow} 0 \text{ ,  as } d \to\infty$$
$$\left|\hat{\overline{\mathcal{W}}}^{*}_{\mathbf{FG}}-{\overline{\mathcal{W}}}^{*}_\mathbf{FG}\right| \stackrel{\mathbb{P}}{\rightarrow} 0 \text{ ,  as } d \to\infty$$
Combining these, we get that given $\mathbf{Z} \sim \mathbf{F}$, 
$$\left|\hat{\overline{\mathcal{W}}}^{*}_{\mathbf{FG}}\mathscr{D}_1(\mathbf{Z})-\hat S_{\mathbf{F G}}S(\mathbf{Z})-\frac{1}{2}\left({\overline{\mathcal{W}}}^{*2}_\mathbf{FG}-\left(T_{\mathbf{F F}}-T_{\mathbf{G G}}\right)^2\right)\right|\stackrel{\mathbb{P}}{\rightarrow} 0 \text{ ,  as } d \to\infty$$
Now, if there exists a $d^{'} \in \mathbb N$ such that $ T_{\mathbf{F G}} > \max \{ T_{\mathbf{F F}}, T_{\mathbf{F G}}\}$ for all $d\geq d^{'}$, \\
$$\begin{aligned}
{\overline{\mathcal{W}}}^{*2}_\mathbf{FG}-\left(T_{\mathbf{F F}}-T_{\mathbf{G G}}\right)^2
&={\overline{\mathcal{W}}}^{*}_\mathbf{FG}-\left(T_{\mathbf{F F}}-T_{\mathbf{G G}}\right))\cdot({\overline{\mathcal{W}}}^{*}_\mathbf{FG}+\left(T_{\mathbf{F F}}-T_{\mathbf{G G}}\right))\\
&=4(T_{\mathbf{F G}}-T_{\mathbf{F F}})(T_{\mathbf{F G}}-T_{\mathbf{G G}})\\
& > 4(T_{\mathbf{F G}}-\max \{ T_{\mathbf{F F}}, T_{\mathbf{G G}}\})^2\\
& > 0 ,\text{ for all } d\geq d^{'}
\end{aligned}$$
\begin{align*}
    &\liminf_d\left(T_{\mathbf{F G}}-\max \{ T_{\mathbf{F F}}, T_{\mathbf{G G}}\}\right)>0 \\
    &\implies \liminf_d\left(T_{\mathbf{F G}}-\max \{ T_{\mathbf{F F}}, T_{\mathbf{G G}}\}\right)^2>0\\
    &\implies \liminf_d \left({\overline{\mathcal{W}}}^{*2}_\mathbf{FG}-\left(T_{\mathbf{F F}}-T_{\mathbf{G G}}\right)^2\right)>4 \cdot \liminf_d \left(T_{\mathbf{F G}}-\max \{ T_{\mathbf{F F}}, T_{\mathbf{G G}}\}\right)^2>0
\end{align*}
Using this, we can show that for every $\epsilon_2>0$, there exists a $d_0'$, such that $$\mathbb{P}[\hat{\overline{\mathcal{W}}}^{*}_{\mathbf{FG}}\mathscr{D}_1(\mathbf{Z})-\hat S_{\mathbf{F G}}S(\mathbf{Z})<0 \mid \mathbf{Z} \sim \mathbf{F}] < \epsilon_2, \text{ for all } d\geq d_0'$$
 So for all $d\geq d_0'$, we have,
$$\mathbb{P}\left[\mathscr{D}_1(\mathbf{Z}) \leq 0 \mid \mathbf{Z} \sim \mathbf{F}\right]\leq \mathbb{P}\left[\hat{\overline{\mathcal{W}}}^{*}_{\mathbf{FG}}\mathscr{D}_1(\mathbf{Z})+\hat S_{\mathbf{F G}}S(\mathbf{Z})\leq 0\mid \mathbf{Z} \sim \mathbf{F}\right]+\epsilon_1+\epsilon_2.$$
Since $\epsilon_1,\epsilon_2>0$ are arbitrary, we can say that, for all $d\geq d_0'$,
$$\mathbb{P}\left[\mathscr{D}_1(\mathbf{Z}) \leq 0 \mid \mathbf{Z} \sim \mathbf{F}\right]\leq \mathbb{P}\left[\hat{\overline{\mathcal{W}}}^{*}_{\mathbf{FG}}\mathscr{D}_1(\mathbf{Z})+\hat S_{\mathbf{F G}}S(\mathbf{Z})\leq 0\mid \mathbf{Z} \sim \mathbf{F}\right]$$
Similarly, one can show that, for all $d \geq d_0'$
$$\mathbb{P}\left[\mathscr{D}_1(\mathbf{Z})>0 \mid \mathbf{Z} \sim \mathbf{G}\right]\leq \mathbb{P}\left[\hat{\overline{\mathcal{W}}}^{*}_{\mathbf{FG}}\mathscr{D}_1(\mathbf{Z})+\hat S_{\mathbf{F G}} S(\mathbf{Z})>0 \mid \mathbf{Z} \sim \mathbf{G}\right]$$
Adding the last two inequalities, we obtain 
$\Delta_2 \geq \Delta_1$ for all $d \geq d_0'$.\qed
\end{itemize}
\begin{lemmaA}
    \label{lem:A3}
 Suppose  assumptions 6,8 and 9 are satisfied.
\begin{itemize}
    \item[(a)] If $\liminf_d\left(\max \{ T_{\mathbf{F F}}, T_{\mathbf{G G}}\}-T_{\mathbf{F G}}\right)>0$
there exists $d_0'\in \mathbb N$, 
such that $\Delta_3 \leq \Delta_1$ for all $d \geq d_0'$.

    \item[(b)] Suppose  Assumption 6 holds in addition. If $\liminf_d\left(T_{\mathbf{F G}}-\max \{ T_{\mathbf{F F}}, T_{\mathbf{F G}}\}\right)>0$, 
there exists $d_0'\in \mathbb N$, such that $\Delta_3 \geq \Delta_1$ for all $d \geq d_0'$.
\end{itemize}
\end{lemmaA}

\noindent\textbf{Proof of \Cref{lem:A3}}
\begin{itemize}
 \item[(a)]   
WLOG, we assume that ${T}_{\mathbf{F F}}>{T}_{\mathbf{G G}}$.

We have shown in  \eqref{eq:w}, \eqref{eq:w-s} and \eqref{eq:sz} that for every $\epsilon_0>0$, there exists a $d_0$, such that $\mathbb{P}[\hat{\overline{\mathcal{W}}}^{*}_{\mathbf{FG}}<0] < \epsilon_0$  for all $d\geq d_0$, for every $\epsilon_1>0$, there exists a $d_1$, such that $\mathbb{P}[\hat{\overline{\mathcal{W}}}^{*}_{\mathbf{FG}}> \hat S_{\mathbf{F G}}] < \epsilon_1$,  for all $d\geq d_1$ and for any $\epsilon_2>0$, there exists $d_2$, such that
$\mathbb{P}[{S}(\mathbf{Z}) < 0 \mid \mathbf{Z} \sim \mathbf{F}]<\epsilon_{2} \text { for all } d \geq d_2$ respectively. It follows that
\begin{align*}
&\mathbb{P}\left[\hat{\overline{\mathcal{W}}}_{\mathbf{FG}}^{*} \operatorname{sign}L_(\mathbf{Z})+\hat S_{\mathbf{F G}}  \operatorname{sign}S(\mathbf{Z}) \leq 0 \mid \mathbf{Z} \sim \mathbf{F}\right] \\
&=\mathbb{P}\left[\hat{\overline{\mathcal{W}}}_{\mathbf{FG}}^{*} \operatorname{sign}\mathscr{D}_1(\mathbf{Z})+\hat S_{\mathbf{F G}} \operatorname{sign}S(\mathbf{Z})\leq 0, S(\mathbf{Z}) \geq 0 \mid \mathbf{Z} \sim \mathbf{F}\right]\\
& \qquad +
\mathbb{P}\left[\hat{\overline{\mathcal{W}}}_{\mathbf{FG}}^{*} \operatorname{sign}\mathscr{D}_1(\mathbf{Z})+\hat S_{\mathbf{F G}}  \operatorname{sign}S(\mathbf{Z}) > 0, S(\mathbf{Z}) < 0, \mid \mathbf{Z} \sim \mathbf{F}\right] \\
&\leq \mathbb{P}\left[\hat{\overline{\mathcal{W}}}_{\mathbf{FG}}^{*} \operatorname{sign}\mathscr{D}_1(\mathbf{Z})+\hat S_{\mathbf{F G}} \operatorname{sign}S(\mathbf{Z})\leq 0,  S(\mathbf{Z})\geq 0, \hat{\overline{\mathcal{W}}}^{*}_{\mathbf{FG}}\leq \hat S_{\mathbf{F G}}\mid \mathbf{Z} \sim \mathbf{F}\right]\\
&\qquad + \mathbb{P}[\hat{\overline{\mathcal{W}}}^{*}_{\mathbf{FG}}> \hat S_{\mathbf{F G}}]+ \mathbb{P}[S(\mathbf{Z})<0 \mid \mathbf{Z} \sim \mathbf{F}]  \\
&\leq \mathbb{P}\left[\hat{\overline{\mathcal{W}}}_{\mathbf{FG}}^{*}\left\{ \operatorname{sign}\mathscr{D}_1(\mathbf{Z})+ \operatorname{sign}S(\mathbf{Z})\right\} \leq 0, S(\mathbf{Z})\geq 0 \mid \mathbf{Z} \sim \mathbf{F}\right] + \epsilon_1+ \epsilon_2 \\  &\hspace{7cm}\text { for all } d \geq \max\{d_{1},d_2\} \\
&\leq \mathbb{P}\left[\hat{\overline{\mathcal{W}}}_{\mathbf{FG}}^{*}\left\{ \operatorname{sign}\mathscr{D}_1(\mathbf{Z})+ \operatorname{sign}S(\mathbf{Z})\right\} \leq 0, S(\mathbf{Z})\geq 0, \hat{\overline{\mathcal{W}}}_{\mathbf{FG}}^{*}\geq 0 \mid \mathbf{Z} \sim \mathbf{F}\right] \\
&\quad+\mathbb{P}[\hat{\overline{\mathcal{W}}}^{*}_{\mathbf{FG}}<0]+ \epsilon_1+ \epsilon_2 \\
&\leq \mathbb{P}\left[ \operatorname{sign}\mathscr{D}_1(\mathbf{Z})+ \operatorname{sign}S(\mathbf{Z}) \leq 0, S(\mathbf{Z}) \geq 0  \mid \mathbf{Z} \sim \mathbf{F}\right] + \epsilon_0+\epsilon_1+\epsilon_2\\
&\leq \mathbb{P}\left[ \operatorname{sign}\mathscr{D}_1(\mathbf{Z})\leq 0 \mid \mathbf{Z} \sim \mathbf{F}\right]+ \epsilon_0 +\epsilon_1 +\epsilon_2 \\
&= \mathbb{P}\left[\mathscr{D}_1(\mathbf{Z})\leq 0 \mid \mathbf{Z} \sim \mathbf{F}\right] + \epsilon_0 +\epsilon_1 + \epsilon_2
\end{align*}
Similarly, one can show that, for all $d \geq \max\{d_{0},d_1,d_2,d^{'}\}$
\begin{equation*}
\mathbb{P}\left[\hat{\overline{\mathcal{W}}}^{*}_{\mathbf{FG}} \operatorname{sign}\mathscr{D}_1(\mathbf{Z})+\hat S_{\mathbf{F G}} \operatorname{sign}( S(\mathbf{Z}))>0 \mid \mathbf{Z} \sim \mathbf{G}\right]\leq \mathbb{P}\left[\mathscr{D}_1(\mathbf{Z})>0 \mid \mathbf{Z} \sim \mathbf{G}\right]+\epsilon_0 +\epsilon_1+\epsilon_2
\end{equation*}

Combining the two inequalities, we obtain
$$
\Delta_{3} \leq \Delta_{1}+ \epsilon_0 +\epsilon_1+\epsilon_2 \text { for all } d \geq \max\{d_{0},d_1,d_2,d^{'}\}
$$
Let, $d_{0}^{\prime}=\max \left\{ d_{0},d_1,d_2,d^{'}\right\}$. Since, $\epsilon_{0},\epsilon_{1},\epsilon_{2}>0$ are arbitrary, we have $\Delta_{3} \leq \Delta_{1}$ for all $d \geq d_{0}^{\prime}$.

\item [(b)] In \eqref{eq:w}, we have shown that for every $\epsilon_0>0$, there exists a $d_0$, such that $\mathbb{P}[\hat{\overline{\mathcal{W}}}^{*}_{\mathbf{FG}}<0] < \epsilon_0$,  for all $d\geq d_0$.
\begin{align*}
&\mathbb{P}\left[\mathscr{D}_1(\mathbf{Z}) \leq 0 \mid \mathbf{Z} \sim \mathbf{F}\right]\\
&=\mathbb{P}\left[ \operatorname{sign}\mathscr{D}_1(\mathbf{Z}) \leq 0 \mid \mathbf{Z} \sim \mathbf{F}\right]\\
&=\mathbb{P}\left[ \operatorname{sign}\mathscr{D}_1(\mathbf{Z}) \leq 0 ,\hat{\overline{\mathcal{W}}}^{*}_{\mathbf{FG}}\geq 0\mid \mathbf{Z} \sim \mathbf{F}\right]+\mathbb{P}\left[ \operatorname{sign}\mathscr{D}_1(\mathbf{Z}) \leq 0,\hat{\overline{\mathcal{W}}}^{*}_{\mathbf{FG}}< 0 \mid \mathbf{Z}\sim \mathbf{F}\right]\\
&\leq \mathbb{P}\left[\hat{\overline{\mathcal{W}}}^{*}_{\mathbf{FG}} \operatorname{sign}\mathscr{D}_1(\mathbf{Z}) \leq 0 \mid \mathbf{Z} \sim \mathbf{F}\right]+\mathbb{P}\left[\hat{\overline{\mathcal{W}}}^{*}_{\mathbf{FG}}< 0 \mid \mathbf{Z}\sim \mathbf{F}\right]\\
&\leq \mathbb{P}\left[\hat{\overline{\mathcal{W}}}^{*}_{\mathbf{FG}} \operatorname{sign}\mathscr{D}_1(\mathbf{Z})\leq 0 ,\hat S_{\mathbf{F G}} \operatorname{sign}S(\mathbf{Z})\leq 0\mid \mathbf{Z} \sim \mathbf{F}\right]\\
&\quad+\mathbb{P}\left[\hat{\overline{\mathcal{W}}}^{*}_{\mathbf{FG}} \operatorname{sign}\mathscr{D}_1(\mathbf{Z}) \leq 0,\hat S_{\mathbf{F G}} \operatorname{sign}S(\mathbf{Z})> 0 \mid \mathbf{Z}\sim \mathbf{F}\right]+\epsilon_0\\
&\leq \mathbb{P}\left[\hat{\overline{\mathcal{W}}}^{*}_{\mathbf{FG}} \operatorname{sign}\mathscr{D}_1(\mathbf{Z})+\hat S_{\mathbf{F G}} \operatorname{sign}S(\mathbf{Z})\leq 0\mid \mathbf{Z} \sim \mathbf{F}\right]\\
&\quad+\mathbb{P}\left[\hat{\overline{\mathcal{W}}}^{*}_{\mathbf{FG}} \operatorname{sign}\mathscr{D}_1(\mathbf{Z})-\hat S_{\mathbf{F G}} \operatorname{sign}S(\mathbf{Z})< 0\mid \mathbf{Z} \sim \mathbf{F}\right]+\epsilon_0,\text{ for all }d\geq d_0
\end{align*}

In the proof of Theorem 4, we have shown that given $\mathbf{Z} \sim \mathbf{F}$,
$$ \hat S_{\mathbf{F G}} \operatorname{sign}S(\mathbf{Z})-\left|T_{\mathbf{F F}}-T_{\mathbf{G G}}\right|\stackrel{\mathbb{P}}{\rightarrow} 0\text{ ,  as } d \to\infty$$
$$\hat{\overline{\mathcal{W}}}^{*}_{\mathbf{FG}} \operatorname{sign}\mathscr{D}_1(\mathbf{Z})-{\overline{\mathcal{W}}}^{*}_\mathbf{FG}\stackrel{\mathbb{P}}{\rightarrow} 0\text{ ,  as } d \to\infty$$
Hence, $\text{ as } d \to \infty$,
$$\left|\hat{\overline{\mathcal{W}}}^{*}_{\mathbf{FG}} \operatorname{sign}\mathscr{D}_1(\mathbf{Z})-\hat S_{\mathbf{F G}} \operatorname{sign}S(\mathbf{Z})-\left({\overline{\mathcal{W}}}^{*}_\mathbf{FG}-\left|T_{\mathbf{F F}}-T_{\mathbf{G G}}\right|\right)\right|\stackrel{\mathbb{P}}{\rightarrow} 0.$$
Now, 
\begin{align*}
 &{\overline{\mathcal{W}}}^{*}_\mathbf{FG}-\left|T_{\mathbf{F F}}-T_{\mathbf{G G}}\right|\\
&=2T_{\mathbf{F G}}-T_{\mathbf{F F}}-T_{\mathbf{G G}}-\left(\max \{ T_{\mathbf{F F}}, T_{\mathbf{G G}}\}-\min \{ T_{\mathbf{F F}}, T_{\mathbf{G G}}\}\right)\\
&=2(T_{\mathbf{F G}}-\max \{ T_{\mathbf{F F}}, T_{\mathbf{G G}}\}).   
\end{align*}

So, $\liminf_d \left({\overline{\mathcal{W}}}^{*}_\mathbf{FG}-\left|T_{\mathbf{F F}}-T_{\mathbf{G G}}\right|\right)>0$ and using this, one can show that for every $\epsilon_1>0$, there exists a $d_0'$, such that $\mathbb{P}[\hat{\overline{\mathcal{W}}}^{*}_{\mathbf{FG}} \operatorname{sign}\mathscr{D}_1(\mathbf{Z})-\hat S_{\mathbf{F G}} \operatorname{sign}S(\mathbf{Z})<0] < \epsilon_1$,  for all $d\geq d_0'$.\\\\
Therefore, for all $d\geq \max\{d_0,d_0'\}$, we have,
\begin{align*}
 \mathbb{P}\left[\mathscr{D}_1(\mathbf{Z}) \leq 0 \mid \mathbf{Z} \sim \mathbf{F}\right]&\leq \mathbb{P}\left[\hat{\overline{\mathcal{W}}}^{*}_{\mathbf{FG}} \operatorname{sign}\mathscr{D}_1(\mathbf{Z})-\hat S_{\mathbf{F G}} \operatorname{sign}S(\mathbf{Z})\leq 0\mid \mathbf{Z} \sim \mathbf{F}\right]\\
&\quad+\epsilon_1+\epsilon_0   
\end{align*}
Since $\epsilon_1,\epsilon_0>0$ are arbitrary, we can say that, for all $d\geq \max\{d_0,d_0'\}$,
$$\mathbb{P}\left[\mathscr{D}_1(\mathbf{Z}) \leq 0 \mid \mathbf{Z} \sim \mathbf{F}\right]\leq \mathbb{P}\left[\hat{\overline{\mathcal{W}}}^{*}_{\mathbf{FG}} \operatorname{sign}\mathscr{D}_1(\mathbf{Z})-\hat S_{\mathbf{F G}} \operatorname{sign}S(\mathbf{Z})\leq 0\mid \mathbf{Z} \sim \mathbf{F}\right]$$
Similarly, one can show that, for all $d\geq \max\{d_0,d_0'\}$
$$\mathbb{P}\left[\mathscr{D}_1(\mathbf{Z})>0 \mid \mathbf{Z} \sim \mathbf{G}\right]\leq \mathbb{P}\left[\hat{\overline{\mathcal{W}}}^{*}_{\mathbf{FG}} \operatorname{sign}\mathscr{D}_1(\mathbf{Z})-\hat S_{\mathbf{F G}} \operatorname{sign}S(\mathbf{Z})>0 \mid \mathbf{Z} \sim \mathbf{G}\right]$$
Adding the two inequalities, we obtain 
$\Delta_3 \geq \Delta_1$ for all $d\geq \max\{d_0,d_0'\}$.
\end{itemize}
\qed
\begin{lemmaA}
    \label{lem:A4}
    Suppose  assumptions 6,8 and 9 are satisfied.
\begin{itemize}
    \item[(a)] If $\liminf_d\left(\max \{ T_{\mathbf{F F}}, T_{\mathbf{G G}}\}-T_{\mathbf{F G}}\right)>0$, there exists $d_0'\in \mathbb N$, such that $\Delta_2 \leq \Delta_3$ for all $d \geq d_0'$.
    \item[(b)] Suppose  Assumption 6 holds in addition. If $\liminf_d\left(T_{\mathbf{F G}}-\max \{ T_{\mathbf{F F}}, T_{\mathbf{F G}}\}\right)>0$, 
there exists $d_0'\in \mathbb N$, such that $\Delta_3 \leq \Delta_2$ for all $d \geq d_0'$.
\end{itemize}
\end{lemmaA}
\noindent\textbf{Proof of \Cref{lem:A4}}
\begin{itemize}
    \item[(a)] WLOG, we assume that ${T}_{\mathbf{F F}}>{T}_{\mathbf{G G}}$. We have shown in  \eqref{eq:w} and \eqref{eq:w-s} that for every $\epsilon_0>0$, there exists a $d_0$, such that $\mathbb{P}[\hat{\overline{\mathcal{W}}}^{*}_{\mathbf{FG}}<0] < \epsilon_0$  for all $d\geq d_0$, for every $\epsilon_1>0$, there exists a $d_1$, such that $\mathbb{P}[\hat{\overline{\mathcal{W}}}^{*}_{\mathbf{FG}}> \hat S_{\mathbf{F G}}] < \epsilon_1$.
\begin{align*}
&\mathbb{P}\left[\hat{\overline{\mathcal{W}}}^{*}_{\mathbf{FG}}\mathscr{D}_1(\mathbf{Z})+\hat S_{\mathbf{F G}} S(\mathbf{Z}) \leq 0 \mid \mathbf{Z} \sim \mathbf{F}\right] \\
 &\leq \mathbb{P}\left[\hat{\overline{\mathcal{W}}}^{*}_{\mathbf{FG}}\mathscr{D}_1(\mathbf{Z})+\hat S_{\mathbf{F G}} S(\mathbf{Z}) \leq 0, \hat{\overline{\mathcal{W}}}^{*}_{\mathbf{FG}} \geq 0 \mid \mathbf{Z} \sim \mathbf{F}\right]+\mathbb{P}[\hat{\overline{\mathcal{W}}}^{*}_{\mathbf{FG}}<0]\\
& \leq \mathbb{P}\left[\hat{\overline{\mathcal{W}}}^{*}_{\mathbf{FG}}\mathscr{D}_1(\mathbf{Z})+\hat S_{\mathbf{F G}} S(\mathbf{Z}) \leq 0, 0 \leq \hat{\overline{\mathcal{W}}}^{*}_{\mathbf{FG}} \leq \hat S_{\mathbf{F G}} \mid \mathbf{Z} \sim \mathbf{F}\right]\\
&\quad+\mathbb{P}\left[\hat{\overline{\mathcal{W}}}^{*}_{\mathbf{FG}} > \hat S_{\mathbf{F G}}\right]+\epsilon_0\\
& \leq \mathbb{P}\left[\hat{\overline{\mathcal{W}}}^{*}_{\mathbf{FG}}\mathscr{D}_1(\mathbf{Z})+\hat S_{\mathbf{F G}} S(\mathbf{Z}) \leq 0, 0 \leq \hat{\overline{\mathcal{W}}}^{*}_{\mathbf{FG}} \leq \hat S_{\mathbf{F G}},S(\mathbf{Z}) < 0 \mid \mathbf{Z} \sim \mathbf{F}\right]\\
&\quad+\mathbb{P}\left[\hat{\overline{\mathcal{W}}}^{*}_{\mathbf{FG}}\mathscr{D}_1(\mathbf{Z})+\hat S_{\mathbf{F G}} S(\mathbf{Z}) \leq 0, 0 \leq \hat{\overline{\mathcal{W}}}^{*}_{\mathbf{FG}} \leq \hat S_{\mathbf{F G}},S(\mathbf{Z}) \geq 0 \mid \mathbf{Z} \sim \mathbf{F}\right]+\epsilon_0+\epsilon_1 
\end{align*}
Now, we see that $0 \leq \hat{\overline{\mathcal{W}}}^{*}_{\mathbf{FG}} \leq \hat S_{\mathbf{F G}}$ implies that $ \hat{\overline{\mathcal{W}}}^{*}_{\mathbf{FG}}\operatorname{sign}\mathscr{D}_1(\mathbf{Z}) - \hat S_{\mathbf{F G}}\leq 0$. Therefore, 
\begin{align*}
&\mathbb{P}\left[\hat{\overline{\mathcal{W}}}^{*}_{\mathbf{FG}}\mathscr{D}_1(\mathbf{Z})+\hat S_{\mathbf{F G}} S(\mathbf{Z}) \leq 0, 0 \leq\hat{\overline{\mathcal{W}}}^{*}_{\mathbf{FG}} \leq \hat S_{\mathbf{F G}},S(\mathbf{Z}) < 0 \mid \mathbf{Z} \sim \mathbf{F}\right]  \\
&=\mathbb{P}\bigg[\hat{\overline{\mathcal{W}}}^{*}_{\mathbf{FG}}\mathscr{D}_1(\mathbf{Z})+\hat S_{\mathbf{F G}} S(\mathbf{Z}) \leq 0, 0 \leq \hat{\overline{\mathcal{W}}}^{*}_{\mathbf{FG}} \leq \hat S_{\mathbf{F G}},\operatorname{sign}S(\mathbf{Z})=-1,\\
&\qquad\qquad\hat{\overline{\mathcal{W}}}^{*}_{\mathbf{FG}}\operatorname{sign}\mathscr{D}_1(\mathbf{Z}) - \hat S_{\mathbf{F G}}\leq 0 \mid \mathbf{Z} \sim \mathbf{F}\bigg]\\
  &\leq \mathbb{P}\left[\hat{\overline{\mathcal{W}}}^{*}_{\mathbf{FG}}\operatorname{sign}\mathscr{D}_1(\mathbf{Z}) + \hat S_{\mathbf{F G}}\operatorname{sign}S(\mathbf{Z})\leq 0 \mid \mathbf{Z} \sim \mathbf{F}\right]
\end{align*}
Now,
 \begin{align*}
&\mathbb{P}\left[\hat{\overline{\mathcal{W}}}^{*}_{\mathbf{FG}}\mathscr{D}_1(\mathbf{Z})+\hat S_{\mathbf{F G}} S(\mathbf{Z}) \leq 0, 0 \leq \hat{\overline{\mathcal{W}}}^{*}_{\mathbf{FG}} \leq \hat S_{\mathbf{F G}},S(\mathbf{Z}) \geq 0 \mid \mathbf{Z} \sim \mathbf{F}\right]\\ 
 &=\mathbb{P}\left[\mathscr{D}_1(\mathbf{Z})\leq -\frac{\hat S_{\mathbf{F G}}}{\hat{\overline{\mathcal{W}}}^{*}_{\mathbf{FG}}} S(\mathbf{Z}) , 0 \leq \hat{\overline{\mathcal{W}}}^{*}_{\mathbf{FG}} \leq \hat S_{\mathbf{F G}},S(\mathbf{Z}) \geq 0 \mid \mathbf{Z} \sim \mathbf{F}\right]\\ 
  &=\mathbb{P}\left[\mathscr{D}_1(\mathbf{Z})\leq -S(\mathbf{Z})\mid \mathbf{Z} \sim \mathbf{F}\right]\\
   &=\mathbb{P}\left[2T_{\mathbf{G}}(\mathbf{Z})-\hat T_{\mathbf{F G}}-\hat T_{\mathbf{G G}}\leq 0\mid \mathbf{Z} \sim \mathbf{F}\right]\\
 \end{align*}
 Given $\mathbf{Z} \sim \mathbf{F}$,
 \begin{equation*}
     \left|2T_{\mathbf{G}}(\mathbf{Z})-\hat T_{\mathbf{F G}}-\hat T_{\mathbf{G G}}-(T_{\mathbf{F G}}-T_{\mathbf{G G}})\right| \stackrel{\mathbb{P}}{\rightarrow}0, \text{ ,  as } d \to \infty
 \end{equation*}
 $$\liminf_d(T_{\mathbf{F G}}-T_{\mathbf{G G}})=\liminf_d{\left(\hat{\overline{\mathcal{W}}}^{*}_{\mathbf{FG}}+(T_{\mathbf{F F}}-T_{\mathbf{F G}})\right)}>0$$
 Hence, for any $\epsilon_2>0$, there exists $d_2$, such that
 $$\mathbb{P}\left[2T_{\mathbf{G}}(\mathbf{Z})-\hat T_{\mathbf{F G}}-\hat T_{\mathbf{G G}}\leq 0\mid \mathbf{Z} \sim \mathbf{F}\right]<\epsilon_2, \text{ for all } d\geq d_2$$

 Combining all these, we get that for $d\geq \max\{d_0,d_1,d_2\}= d_0'$ (let),
 \begin{align*}
&\mathbb{P}\left[\hat{\overline{\mathcal{W}}}^{*}_{\mathbf{FG}}\mathscr{D}_1(\mathbf{Z})+\hat S_{\mathbf{F G}} S(\mathbf{Z}) \leq 0 \mid \mathbf{Z} \sim \mathbf{F}\right]\\
&\quad
 \leq \mathbb{P}\left[\hat{\overline{\mathcal{W}}}^{*}_{\mathbf{FG}}\operatorname{sign}\mathscr{D}_1(\mathbf{Z}) + \hat S_{\mathbf{F G}}\operatorname{sign}S(\mathbf{Z})\leq 0 \mid \mathbf{Z} \sim \mathbf{F}\right]+\epsilon_0+\epsilon_1+\epsilon_2
 \end{align*}
Since $\epsilon_0,\epsilon_1,\epsilon_2>0$ are arbitrary, we can say that, for all $d\geq d_0'$,
\begin{align*}
 &\mathbb{P}\left[\hat{\overline{\mathcal{W}}}^{*}_{\mathbf{FG}}\mathscr{D}_1(\mathbf{Z})+\hat S_{\mathbf{F G}} S(\mathbf{Z}) \leq 0 \mid \mathbf{Z} \sim \mathbf{F}\right]
 \leq \mathbb{P}\left[\hat{\overline{\mathcal{W}}}^{*}_{\mathbf{FG}}\operatorname{sign}\mathscr{D}_1(\mathbf{Z}) + \hat S_{\mathbf{F G}}\operatorname{sign}S(\mathbf{Z})\leq 0 \mid \mathbf{Z} \sim \mathbf{F}\right].
 \end{align*}
 Similarly, one can show that, for all $d \geq d_0'$,
 \begin{align*}
&\mathbb{P}\left[\hat{\overline{\mathcal{W}}}^{*}_{\mathbf{FG}}\mathscr{D}_1(\mathbf{Z})+\hat S_{\mathbf{F G}} S(\mathbf{Z}) > 0 \mid \mathbf{Z} \sim \mathbf{G}\right]\leq \mathbb{P}\left[\hat{\overline{\mathcal{W}}}^{*}_{\mathbf{FG}}\operatorname{sign}\mathscr{D}_1(\mathbf{Z}) + \hat S_{\mathbf{F G}}\operatorname{sign}S(\mathbf{Z})>0 \mid \mathbf{Z} \sim \mathbf{G}\right].
 \end{align*}
 Combing the above two inequalities, we obtain 
$\Delta_2 \leq \Delta_3$ for all $d \geq d_0'$.
\item[(b)]
 WLOG, assume that ${T}_{\mathbf{F F}}>{T}_{\mathbf{G G}}$. W e know that $\hat S_{\mathbf{F G}} \stackrel{\mathbb{P}}{\rightarrow} {T}_{\mathbf{F F}}-{T}_{\mathbf{G G}}$, as $d \to \infty$. Then Assumption 6 implies that
for every $\epsilon_0>0$, there exists a $d_0$, such that $$\mathbb{P}[\hat S_{\mathbf{F G}}\leq0] < \epsilon_0 \text { for all } d \geq d_0$$

Since, $\liminf_d\left(T_{\mathbf{F G}}-\max \{ T_{\mathbf{F F}}, T_{\mathbf{F G}}\}\right)>0$,
 there exists $d' \in \mathbb N$ such that $ T_{\mathbf{F G}}  > T_{\mathbf{F F}} $ for all $d\geq d^{'}$. Hence, for all $d\geq d^{'}$,
$$
\hat{\overline{\mathcal{W}}}^{*}_\mathbf{FG}=\left({T}_{\mathbf{F G}}-{T}_{\mathbf{F F}}\right)+\left({T}_{\mathbf{F G}}-{T}_{\mathbf{G G}}\right) > {T}_{\mathbf{F F}}-{T}_{\mathbf{G G}}= S_{\mathbf{F G}} 
$$
Since $\hat{\overline{\mathcal{W}}}^{*}_{\mathbf{FG}}-\hat S_{\mathbf{F G}}-\left({\overline{\mathcal{W}}}^{*}_\mathbf{FG}-S_{\mathbf{F G}}\right)\stackrel{\mathbb{P}}{\rightarrow} 0$ as $d \rightarrow \infty$. 

Also, since $\liminf_d\left({\overline{\mathcal{W}}}^{*}_\mathbf{FG}-S_{\mathbf{F G}}\right)=\liminf_d 2\left({T}_{\mathbf{F G}}-{T}_{\mathbf{F F}}\right)>0$, hence, we can show that for every $\epsilon_1>0$, there exists a $d_1$ such that $$\mathbb{P}[\hat{\overline{\mathcal{W}}}^{*}_{\mathbf{FG}}\leq \hat S_{\mathbf{F G}}] < \epsilon_1 \text { for all } d \geq d_1.$$
It follows that
\begin{align*}
&\mathbb{P}\left[\hat{\overline{\mathcal{W}}}^{*}_{\mathbf{FG}} \operatorname{sign}\mathscr{D}_1(\mathbf{Z})+\hat S_{\mathbf{F G}}  \operatorname{sign}S(\mathbf{Z}) \leq 0 \mid \mathbf{Z} \sim \mathbf{F}\right] \\
&\leq\mathbb{P}\left[\hat{\overline{\mathcal{W}}}^{*}_{\mathbf{FG}} \operatorname{sign}\mathscr{D}_1(\mathbf{Z})+\hat S_{\mathbf{F G}}  \operatorname{sign}S(\mathbf{Z}) \leq 0, \hat S_{\mathbf{F G}}> 0 \mid \mathbf{Z} \sim \mathbf{F}\right]+\mathbb{P}[\hat S_{\mathbf{F G}}\leq0]\\
&=\mathbb{P}\left[\hat{\overline{\mathcal{W}}}^{*}_{\mathbf{FG}} \operatorname{sign}\mathscr{D}_1(\mathbf{Z})+\hat S_{\mathbf{F G}}  \operatorname{sign}S(\mathbf{Z}) \leq 0, \hat{\overline{\mathcal{W}}}^{*}_{\mathbf{FG}} >\hat S_{\mathbf{F G}}> 0 \mid \mathbf{Z} \sim \mathbf{F}\right]+\mathbb{P}[\hat{\overline{\mathcal{W}}}^{*}_{\mathbf{FG}}\leq \hat S_{\mathbf{F G}}]+\epsilon_0\\
&\leq \mathbb{P}\bigg[\hat{\overline{\mathcal{W}}}^{*}_{\mathbf{FG}} \operatorname{sign}\mathscr{D}_1(\mathbf{Z})+\hat S_{\mathbf{F G}}  \operatorname{sign}S(\mathbf{Z}) \leq 0,\hat{\overline{\mathcal{W}}}^{*}_{\mathbf{FG}}\mathscr{D}_1(\mathbf{Z})+\hat S_{\mathbf{F G}} S(\mathbf{Z}) \leq 0,\hat{\overline{\mathcal{W}}}^{*}_{\mathbf{FG}} > \hat S_{\mathbf{F G}}> 0 \mid \mathbf{Z} \sim \mathbf{F}\bigg]\\
&\quad + \mathbb{P}\bigg[\hat{\overline{\mathcal{W}}}^{*}_{\mathbf{FG}} \operatorname{sign}\mathscr{D}_1(\mathbf{Z})+\hat S_{\mathbf{F G}}  \operatorname{sign}S(\mathbf{Z}) \leq 0, \hat{\overline{\mathcal{W}}}^{*}_{\mathbf{FG}} > \hat S_{\mathbf{F G}}> 0, \\
&\qquad \qquad\hat{\overline{\mathcal{W}}}^{*}_{\mathbf{FG}}\mathscr{D}_1(\mathbf{Z})+\hat S_{\mathbf{F G}} S(\mathbf{Z}) > 0 \mid \mathbf{Z} \sim \mathbf{F}\bigg]+\epsilon_0+\epsilon_1\\
&\leq \mathbb{P}\left[ \hat{\overline{\mathcal{W}}}^{*}_{\mathbf{FG}}\mathscr{D}_1(\mathbf{Z})+\hat S_{\mathbf{F G}} S(\mathbf{Z}) \leq 0 \mid \mathbf{Z} \sim \mathbf{F}\right]+\epsilon_0+\epsilon_1+p_0
\end{align*}
where $p_0=\mathbb{P}(A(\mathbf{Z}))$ with
\begin{align*}
    A(\mathbf{Z})=&\{\hat{\overline{\mathcal{W}}}^{*}_{\mathbf{FG}} \operatorname{sign}\mathscr{D}_1(\mathbf{Z})+\hat S_{\mathbf{F G}}  \operatorname{sign}S(\mathbf{Z}) \leq 0, \\&\qquad \hat{\overline{\mathcal{W}}}^{*}_{\mathbf{FG}}\mathscr{D}_1(\mathbf{Z})+\hat S_{\mathbf{F G}} S(\mathbf{Z}) > 0, \hat{\overline{\mathcal{W}}}^{*}_{\mathbf{FG}} >\hat S_{\mathbf{F G}}> 0 \mid \mathbf{Z} \sim \mathbf{F}\}
\end{align*}
We consider the four mutually disjoint and exclusive events here,
\begin{itemize}
    \item[1.]$L_{\mathbf{G}}(\mathbf{Z})>L_{\mathbf{F}}(\mathbf{Z}), S(\mathbf{Z})>0  \implies A(\mathbf{Z})$ cannot occur.
    \item[2.]$L_{\mathbf{G}}(\mathbf{Z})>L_{\mathbf{F}}(\mathbf{Z}), S(\mathbf{Z})<0 \implies A(\mathbf{Z})$ cannot occur.
    \item[3.]$L_{\mathbf{G}}(\mathbf{Z})<L_{\mathbf{F}}(\mathbf{Z}), S(\mathbf{Z})>0 $
    \item[4.]$L_{\mathbf{G}}(\mathbf{Z})<L_{\mathbf{F}}(\mathbf{Z}), S(\mathbf{Z})<0, \implies A(\mathbf{Z})$ cannot occur.
\end{itemize}
Hence, $A(\mathbf{Z})$ implies $L_{\mathbf{G}}(\mathbf{Z})<L_{\mathbf{F}}(\mathbf{Z}), S(\mathbf{Z})>0 $. So,
\begin{align*}
p_0 & = \mathbb{P}[\hat{\overline{\mathcal{W}}}^{*}_{\mathbf{FG}} \operatorname{sign}\mathscr{D}_1(\mathbf{Z})+\hat S_{\mathbf{F G}}  \operatorname{sign}S(\mathbf{Z}) \leq 0, 
\hat{\overline{\mathcal{W}}}^{*}_{\mathbf{FG}}\mathscr{D}_1(\mathbf{Z})+\hat S_{\mathbf{F G}} S(\mathbf{Z}) > 0,  \\
&\quad\hat{\overline{\mathcal{W}}}^{*}_{\mathbf{FG}} > \hat S_{\mathbf{F G}}> 0, L_{\mathbf{G}}(\mathbf{Z})<L_{\mathbf{F}}(\mathbf{Z}), S(\mathbf{Z})>0 \mid \mathbf{Z} \sim \mathbf{F}]\\
& = \mathbb{P}[\hat{\overline{\mathcal{W}}}^{*}_{\mathbf{FG}} > \hat S_{\mathbf{F G}}> 0,L_{\mathbf{G}}(\mathbf{Z})<L_{\mathbf{F}}(\mathbf{Z}), S(\mathbf{Z})>0, \hat{\overline{\mathcal{W}}}^{*}_{\mathbf{FG}}\mathscr{D}_1(\mathbf{Z})+\hat S_{\mathbf{F G}} S(\mathbf{Z}) > 0 \mid \mathbf{Z} \sim \mathbf{F}] \\
&\leq\mathbb{P}\left[S(\mathbf{Z})-(\mathscr{D}_1(\mathbf{Z}))>0\mid \mathbf{Z} \sim \mathbf{F}\right]\\
&=\mathbb{P}\left[2L_{\mathbf{F}}(\mathbf{Z})-T_{\mathbf{F G}}>0\mid \mathbf{Z} \sim \mathbf{F}\right]\\
&=\mathbb{P}\left[T_{\mathbf{F F}}+T_{\mathbf{F G}}-2T_{\mathbf{F}}(\mathbf{Z})<0\mid \mathbf{Z} \sim \mathbf{F}\right]
\end{align*}
Now, given $\mathbf{Z} \sim \mathbf{F}$, 
$$T_{\mathbf{F F}}+T_{\mathbf{F G}}-2T_{\mathbf{F}}(\mathbf{Z})-( T_{\mathbf{F G}}-T_{\mathbf{F F}})\stackrel{\mathbb{P}}{\rightarrow}0.$$
Now, $\liminf_d(T_{\mathbf{F G}}-T_{\mathbf{F F}})>0$, because we have assumed that $T_{\mathbf{F F}}=\max\{T_{\mathbf{F F}},T_{\mathbf{G G}}\}$. So,
for every $\epsilon_2>0$, there exists a $d_2$ such that $$\mathbb{P}\left[T_{\mathbf{F F}}+T_{\mathbf{F G}}-2T_{\mathbf{F}}(\mathbf{Z})<0\mid \mathbf{Z} \sim \mathbf{F}\right] < \epsilon_2 \text { for all } d \geq d_2.$$
Hence, $p_0\leq\epsilon_2 \text { for all } d \geq d_2.$ And
for all d $\geq \max\{d_0,d_1,d_2\},$
\begin{align*}
&\mathbb{P}\left[\hat{\overline{\mathcal{W}}}^{*}_{\mathbf{FG}} \operatorname{sign}\mathscr{D}_1(\mathbf{Z})+\hat S_{\mathbf{F G}}  \operatorname{sign}S(\mathbf{Z}) \leq 0 \mid \mathbf{Z} \sim \mathbf{F}\right]\\
&\quad \leq \mathbb{P}\left[ \hat{\overline{\mathcal{W}}}^{*}_{\mathbf{FG}}\mathscr{D}_1(\mathbf{Z})+\hat S_{\mathbf{F G}} S(\mathbf{Z}) \leq 0 \mid \mathbf{Z} \sim \mathbf{F}\right]+\epsilon_0+\epsilon_1+\epsilon_2
\end{align*}
Let, $d_0'=\max\{d_0,d_1,d_2\}$. Since $\epsilon_0,\epsilon_1,\epsilon_2>0$ are arbitrary, we can say that, for all $d\geq d_0'$,
\begin{align*}
&\mathbb{P}\left[\hat{\overline{\mathcal{W}}}_{\mathbf{F}\mathbf{G}}^{*}\operatorname{sign}\mathscr{D}_1(\mathbf{Z}) + \hat S_{\mathbf{F G}}\operatorname{sign}S(\mathbf{Z})\leq 0 \mid \mathbf{Z} \sim \mathbf{F}\right] \\
&\quad\leq\mathbb{P}\left[\hat{\overline{\mathcal{W}}}^{*}_{\mathbf{FG}}\mathscr{D}_1(\mathbf{Z})+\hat S_{\mathbf{F G}} S(\mathbf{Z}) \leq 0 \mid \mathbf{Z} \sim \mathbf{F}\right]
 \end{align*}
 Similarly, one can show that, for all $d \geq d_0'$,
 \begin{align*}
&\mathbb{P}\left[\hat{\overline{\mathcal{W}}}_{\mathbf{F}\mathbf{G}}^{*}\operatorname{sign}\mathscr{D}_1(\mathbf{Z}) + \hat S_{\mathbf{F G}}\operatorname{sign}S(\mathbf{Z})>0 \mid \mathbf{Z} \sim \mathbf{G}\right] \\
&\quad
\leq \mathbb{P}\left[\hat{\overline{\mathcal{W}}}^{*}_{\mathbf{FG}}\mathscr{D}_1(\mathbf{Z})+\hat S_{\mathbf{F G}} S(\mathbf{Z}) > 0 \mid \mathbf{Z} \sim \mathbf{G}\right]
 \end{align*}
 Combing the above two inequalities, we obtain 
$\Delta_3 \leq \Delta_2$ for all $d \geq d_0'$.
\end{itemize}
\qed

\noindent\textbf{Proof of Theorem 6}\\
The proof of this theorem follows directly from \Cref{lem:A2}, \Cref{lem:A3} and \Cref{lem:A4}.

\section{\textsc{Additional numerical results \& Details}}
\label{AppendixB}

\subsection{\textsc{Simulation details}}
\begin{itemize}
 \item[$\bullet$]\textbf{GLMNET:} The {\tt R}-package {\tt glmnet} is used for implementing GLMNET. The tuning parameter $\alpha$ in the elastic-net penalty term is fixed at the default value $1$. The weight $\lambda$ of the penalty term is chosen via cross-validation using the function {\tt cv.glmnet} with default values of its parameters.
 \item[$\bullet$] \textbf{NN-RP:} The function {\tt classify} from the package {\tt RandPro} is used with default values of its parameters.
 \item[$\bullet$] \textbf{SVM:} The {\tt R} package {\tt e1071} is used for implementing SVM with linear (SVM-LIN) and radial basis function (SVF-RBF) kernels. For the RBF kernel, i.e., $K_\theta(\mathbf{x},\mathbf{y})=\mathrm{exp}\{-\theta\|\mathbf{x}-\mathbf{y}\|^2\}$, the default value of the tuning parameter $\theta$ was chosen, i.e., $\theta=\frac{1}{d}$.
 \item[$\bullet$] \textbf{N-NET:} We used the {\tt nnet} function from the {\tt R} package {\tt nnet} to fit a single-hidden-layer artificial neural network with default parameters. The number of units in the hidden layer was allowed to vary from $1$ up to $10$. Among them, the one with the minimum misclassification rate was reported as N-NET.
 \item[$\bullet$] \textbf{1-NN:} The {\tt knn1} function from the {\tt R}-package {\tt class} is used for implementing the 1-nearest neighbor classifier.
\end{itemize}

\subsection{\textsc{Tables}}

\begin{table}[!htb]
\centering
\caption{Estimated values of $T_\mathbf{FF}, T_\mathbf{FG}$ and $T_\mathbf{GG}$ with standard errors (in parentheses) for $\delta_1$, $\delta_2$, $\delta_3$, for simulated examples at $d = 1000$}
\label{table-T}
\resizebox{0.8\columnwidth}{!}{%
\begin{tabular}{ccccc}
\toprule
\textbf{Example} & ~~~~~~$\hat{T}_\mathbf{FF}$~~~~~~ & ~~~~~~$\hat{T}_\mathbf{FG}$~~~~~~ & ~~~~~~$\hat{T}_\mathbf{GG}$~~~~~~ & $\hat{T}_\mathbf{FG} > \max(\hat{T}_\mathbf{FF},\hat{T}_\mathbf{GG})$\\
\midrule
 & 0.28362 & 0.31839 & 0.3461 & \\
\multirow{-2}{*}{\centering\arraybackslash 1} & (0.00008) & (0.000023) & (0.000064) & \multirow{-2}{*}{\centering\arraybackslash FALSE}\\
\cmidrule{1-5}
 & 0.34206 & 0.31841 & 0.2876 & \\
\multirow{-2}{*}{\centering\arraybackslash 2} & (0.00007) & (0.000025) & (0.00008) & \multirow{-2}{*}{\centering\arraybackslash FALSE}\\
\cmidrule{1-5}
 & 0.3004 & 0.33211 & 0.30041 & \\
\multirow{-2}{*}{\centering\arraybackslash 3} & (0.000112) & (0.000058) & (0.000104) & \multirow{-2}{*}{\centering\arraybackslash TRUE}\\
\cmidrule{1-5}
 & 0.27954 & 0.31944 & 0.34796 & \\
\multirow{-2}{*}{\centering\arraybackslash 4} & (0.00009) & (0.000024) & (0.000077) & \multirow{-2}{*}{\centering\arraybackslash FALSE}\\
\cmidrule{1-5}
& 0.26462 & 0.32142	& 0.35871 & \\

\multirow{-2}{*}{\centering\arraybackslash 5} & (0.000078) & (0.000032) & (0.000076) & \multirow{-2}{*}{\centering\arraybackslash TRUE}\\
\bottomrule
\end{tabular}
}
\end{table}

\newpage

\begin{table}[!htb]
\centering
\caption{Estimated misclassification probabilities (in percentages) with standard errors (in parentheses) for $\delta_1$, $\delta_2$, $\delta_3$, and popular classifiers \textbf{Simulated Example 1}.}
\label{table-Ex1-updated}
\resizebox{\linewidth}{!}{
\begin{tabular}{ccccccccccc}
\toprule
\multirow[b]{2}{*}{\thead{\textbf{d}}}
        & \multicolumn{7}{c}{\thead{\textbf{Popular Classifiers}}}
            & \multicolumn{3}{c}{\thead{\textbf{Proposed Classifiers}}}\\
\cmidrule(lr){2-8}\cmidrule(lr){9-11}

& \textbf{Bayes} & \textbf{GLMNET} & \textbf{NN-RP} & \textbf{SVM-LIN} & \textbf{SVM-RBF} & \textbf{N-NET} & \textbf{1-NN} & $\bm{\delta_1}$ & $\bm{\delta_2}$ & $\bm{\delta_3}$\\
\midrule
 & 30.36 & 49.30 & 45.10 & 48.49 & 36.29 & 44.76 & 45.68 & 45.23 & 33.66 & 33.62\\

\multirow{-2}{*}{\centering\arraybackslash 5} & (0.0031) & (0.0038) & (0.0030) & (0.0037) & (0.0039) & (0.0042) & (0.0038) & (0.0039) & (0.0032) & (0.0033)\\
\cmidrule{1-11}
 & 22.4 & 49.64 & 46.51 & 49.21 & 29.90 & 44.51 & 44.74 & 43.74 & 26.83 & 26.71\\

\multirow{-2}{*}{\centering\arraybackslash 10} & (0.0027) & (0.0035) & (0.0021) & (0.0033) & (0.0040) & (0.0042) & (0.0035) & (0.0041) & (0.0033) & (0.0032)\\
\cmidrule{1-11}
 & 11.44 & 48.36 & 48.90 & 48.30 & 18.53 & 45.12 & 45.89 & 40.62 & 16.24 & 16.36\\

\multirow{-2}{*}{\centering\arraybackslash 25} & (0.0023) & (0.0033) & (0.0011) & (0.0038) & (0.0032) & (0.0039) & (0.0024) & (0.0036) & (0.0028) & (0.0028)\\
\cmidrule{1-11}
 & 4.16 & 48.05 & 49.93 & 47.76 & 9.45 & 44.82 & 47.98 & 35.98 & 8.04 & 7.96\\

\multirow{-2}{*}{\centering\arraybackslash 50} & (0.0014) & (0.0038) & (0.0000) & (0.0036) & (0.0024) & (0.0041) & (0.0013) & (0.0038) & (0.0020) & (0.0020)\\
\cmidrule{1-11}
 & 0.73 & 48.33 & 50.00 & 47.08 & 2.90 & 44.78 & 49.64 & 30.61 & 2.38 & 2.40\\

\multirow{-2}{*}{\centering\arraybackslash 100} & (0.0000) & (0.0037) & (0.0000) & (0.0032) & (0.0013) & (0.0041) & (0.0000) & (0.0038) & (0.0013) & (0.0013)\\
\cmidrule{1-11}
 & 0.01 & 47.60 & 50.00 & 46.50 & 0.13 & 45.92 & 49.99 & 21.39 & 0.10 & 0.11\\

\multirow{-2}{*}{\centering\arraybackslash 250} & (0.0000) & (0.0038) & (0.0000) & (0.0028) & (0.0000) & (0.0035) & (1e-04) & (0.0032) & (0.0000) & (0.0000)\\
\cmidrule{1-11}
 & 0.00 & 47.95 & 50.00 & 46.48 & 0.00 & 45.08 & 50.00 & 13.30 & 0.00 & 0.00\\

\multirow{-2}{*}{\centering\arraybackslash 500} & (0.0000) & (0.0037) & (0.0000) & (0.0025) & (0.0000) & (0.0038) & (0.0000) & (0.0024) & (0.0000) & (0.0000)\\
\cmidrule{1-11}
 & 0.00 & 47.14 & 50.00 & 47.11 & 0.00 & 44.46 & 50.00 & 5.74 & 0.00 & 0.00\\

\multirow{-2}{*}{\centering\arraybackslash 1000} & (0.0000) & (0.0038) & (0.0000) & (0.0017) & (0.0000) & (0.0040) & (0.0000) & (0.0017) & (0.0000) & (0.0000)\\
\bottomrule
\end{tabular}}
\end{table}

\begin{table}[!htb]
\centering
\caption{Estimated misclassification probabilities (in percentages) with standard errors (in parentheses) for $\delta_1$, $\delta_2$, $\delta_3$, and popular classifiers \textbf{Simulated Example 2}.}
\label{table-Ex2-updated}
\resizebox{\linewidth}{!}{
\begin{tabular}{ccccccccccc}
\toprule
\multirow[b]{2}{*}{\thead{\textbf{d}}}
        & \multicolumn{7}{c}{\thead{\textbf{Popular Classifiers}}}
            & \multicolumn{3}{c}{\thead{\textbf{Proposed Classifiers}}}\\
\cmidrule(lr){2-8}\cmidrule(lr){9-11}

& \textbf{Bayes} & \textbf{GLMNET} & \textbf{NN-RP} & \textbf{SVM-LIN} & \textbf{SVM-RBF} & \textbf{N-NET} & \textbf{1-NN} & $\bm{\delta_1}$ & $\bm{\delta_2}$ & $\bm{\delta_3}$\\
\midrule
 & 30.64 & 49.98 & 46.78 & 49.58 & 43.48 & 47.24 & 45.15 & 45.18 & 34.79 & 35.05\\

\multirow{-2}{*}{\centering\arraybackslash 5} & (0.0030) & (0.0039) & (0.0003) & (0.0039) & (0.0051) & (0.0042) & (0.0039) & (0.0045) & (0.0032) & (0.0041)\\
\cmidrule{1-11}
 & 24.17 & 49.60 & 47.84 & 49.66 & 42.19 & 48.01 & 46.56 & 42.91 & 28.73 & 28.61\\

\multirow{-2}{*}{\centering\arraybackslash 10} & (0.0031) & (0.0035) & (0.0029) & (0.0035) & (0.0042) & (0.0035) & (0.0031) & (0.0036) & (0.0032) & (0.0033)\\
\cmidrule{1-11}
 & 13.05 & 49.94 & 48.91 & 50.04 & 41.54 & 49.63 & 47.84 & 39.53 & 19.22 & 19.40\\

\multirow{-2}{*}{\centering\arraybackslash 25} & (0.0024) & (0.0037) & (0.0020) & (0.0039) & (0.0052) & (0.0036) & (0.0034) & (0.0045) & (0.0030) & (0.0029)\\
\cmidrule{1-11}
 & 5.72 & 49.48 & 49.65 & 50.07 & 40.86 & 49.54 & 48.68 & 35.29 & 11.34 & 11.48\\

\multirow{-2}{*}{\centering\arraybackslash 50} & (0.0016) & (0.0038) & (0.0016) & (0.0040) & (0.0041) & (0.0038) & (0.0027) & (0.0039) & (0.0021) & (0.0020)\\
\cmidrule{1-11}
 & 1.20 & 49.01 & 50.00 & 49.40 & 39.50 & 49.56 & 49.84 & 30.14 & 4.09 & 4.15\\

\multirow{-2}{*}{\centering\arraybackslash 100} & (0.0000) & (0.0031) & (0.0012) & (0.0034) & (0.0044) & (0.0029) & (0.0019) & (0.0034) & (0.0015) & (0.0015)\\
\cmidrule{1-11}
 & 0.00 & 48.80 & 50.15 & 49.28 & 37.24 & 49.12 & 49.63 & 20.71 & 0.31 & 0.30\\

\multirow{-2}{*}{\centering\arraybackslash 250} & (0.0000) & (0.0037) & (0.0011) & (0.0032) & (0.0033) & (0.0037) & (0.0014) & (0.0031) & (0.0000)& (0.0000)\\
\cmidrule{1-11}
 & 0.00 & 49.25 & 50.08 & 49.73 & 35.91 & 49.54 & 50.19 & 12.66 & 0.00 & 0.00\\

\multirow{-2}{*}{\centering\arraybackslash 500} & (0.0000) & (0.0034) & (0.0000) & (0.0035) & (0.0033) & (0.0033) & (0.0000) & (0.0022) & (0.0000) & (0.0000)\\
\cmidrule{1-11}
 & 0.00 & 49.06 & 49.96 & 49.60 & 34.84 & 49.06 & 50.06 & 5.19 & 0.00 & 0.00\\

\multirow{-2}{*}{\centering\arraybackslash 1000} & (0.0000) & (0.0033) & (0.0000) & (0.0032) & (0.0030) & (0.0033) & (0.0000) & (0.0018) & (0.0000) & (0.0000)\\
\bottomrule
\end{tabular}}
\end{table}

\begin{table}[!htb]
\centering
\caption{Estimated misclassification probabilities (in percentages) with standard errors (in parentheses) for $\delta_1$, $\delta_2$, $\delta_3$, and popular classifiers \textbf{Simulated Example 3}.}
\label{table-Ex3-updated}
\resizebox{\linewidth}{!}{
\begin{tabular}{ccccccccccc}
\toprule
\multirow[b]{2}{*}{\thead{\textbf{d}}}
        & \multicolumn{7}{c}{\thead{\textbf{Popular Classifiers}}}
            & \multicolumn{3}{c}{\thead{\textbf{Proposed Classifiers}}}\\
\cmidrule(lr){2-8}\cmidrule(lr){9-11}

& \textbf{Bayes} & \textbf{GLMNET} & \textbf{NN-RP} & \textbf{SVM-LIN} & \textbf{SVM-RBF} & \textbf{N-NET} & \textbf{1-NN} & $\bm{\delta_1}$ & $\bm{\delta_2}$ & $\bm{\delta_3}$\\
\midrule
 & 21.94 & 40.09 & 39.20 & 43.21 & 41.50 & 38.29 & 40.65 & 28.28 & 30.41 & 30.30\\

\multirow{-2}{*}{\centering\arraybackslash 5} & (0.0030) & (0.0060) & (0.0047) & (0.0059) & (0.0058) & (0.0050) & (0.0041) & (0.0041) & (0.0058) & (0.0074)\\
\cmidrule{1-11}
 & 14.18 & 39.50 & 41.54 & 41.64 & 41.52 & 39.58 & 41.82 & 21.40 & 22.92 & 22.00\\

\multirow{-2}{*}{\centering\arraybackslash 10} & (0.0025) & (0.0053) & (0.0050) & (0.0058) & (0.0054) & (0.0044) & (0.0043) & (0.0037) & (0.0043) & (0.0057)\\
\cmidrule{1-11}
 & 4.64 & 37.41 & 44.6 & 40.71 & 41.69 & 41.92 & 44.31 & 10.18 & 10.57 & 10.18\\

\multirow{-2}{*}{\centering\arraybackslash 25} & (0.0013) & (0.0045) & (0.0048) & (0.0050) & (0.0049) & (0.0042) & (0.0040) & (0.0025) & (0.0028) & (0.0025)\\
\cmidrule{1-11}
 & 0.64 & 37.48 & 46.80 & 41.62 & 43.43 & 44.16 & 46.61 & 3.86 & 4.04 & 3.86\\

\multirow{-2}{*}{\centering\arraybackslash 50} & (0.0000) & (0.0042) & (0.0037) & (0.0044) & (0.0041) & (0.0042) & (0.0040) & (0.0014) & (0.0015) & (0.0014)\\
\cmidrule{1-11}
 & 0.04 & 36.07 & 47.88 & 39.93 & 43.17 & 44.97 & 47.54 & 0.69 & 0.76 & 0.69\\

\multirow{-2}{*}{\centering\arraybackslash 100} & (0.0000) & (0.0038) & (0.0036) & (0.0042) & (0.0032) & (0.0035) & (0.0037) & (0.0000) & (0.0000) & (0.0000)\\
\cmidrule{1-11}
 & 0.00 & 35.82 & 49.42 & 40.45 & 46.26 & 46.44 & 48.62 & 0.01 & 0.01 & 0.01\\

\multirow{-2}{*}{\centering\arraybackslash 250} & (0.0000) & (0.0036) & (0.0023) & (0.0035) & (0.0023) & (0.0039) & (0.0032) & (0.0000) & (0.0000) & (0.0000)\\
\cmidrule{1-11}
 & 0.00 & 35.78 & 49.83 & 39.76 & 48.05 & 46.74 & 49.66 & 0.00 & 0.00 & 0.00\\

\multirow{-2}{*}{\centering\arraybackslash 500} & (0.0000) & (0.0032) & (0.0022) & (0.0033) & (0.0016) & (0.0040) & (0.0025) & (0.0000) & (0.0000) & (0.0000)\\
\cmidrule{1-11}
 & 0.00 & 35.27 & 50.24 & 39.69 & 49.64 & 47.78 & 49.53 & 0.00 & 0.00 & 0.00\\

\multirow{-2}{*}{\centering\arraybackslash 1000} & (0.0000) & (0.0035) & (0.0022) & (0.0031) & (0.0000) & (0.0041) & (0.0029) & (0.0000) & (0.0000) & (0.0000)\\
\bottomrule
\end{tabular}}
\end{table}

\vspace{2cm}

\begin{table}[!htb]
\centering
\caption{Estimated misclassification probabilities (in percentages) with standard errors (in parentheses) for $\delta_1$, $\delta_2$, $\delta_3$, and popular classifiers \textbf{Simulated Example 4}.}
\label{table-Ex4-updated}
\resizebox{\linewidth}{!}{
\begin{tabular}{ccccccccccc}
\toprule
\multirow[b]{2}{*}{\thead{\textbf{d}}}
        & \multicolumn{7}{c}{\thead{\textbf{Popular Classifiers}}}
            & \multicolumn{3}{c}{\thead{\textbf{Proposed Classifiers}}}\\
\cmidrule(lr){2-8}\cmidrule(lr){9-11}

& \textbf{Bayes} & \textbf{GLMNET} & \textbf{NN-RP} & \textbf{SVM-LIN} & \textbf{SVM-RBF} & \textbf{N-NET} & \textbf{1-NN} & $\bm{\delta_1}$ & $\bm{\delta_2}$ & $\bm{\delta_3}$\\
\midrule
 & 28.90 & 45.85 & 44.24 & 46.37 & 40.02 & 41.82 & 40.68 & 43.39 & 31.5 & 31.52\\

\multirow{-2}{*}{\centering\arraybackslash 5} & (0.0025) & (0.0023) & (0.0023) & (0.0017) & (0.0033) & (0.0023) & (0.0028) & (0.0046) & (0.0033) & (0.0034)\\
\cmidrule{1-11}
 & 22.44 & 45.54 & 45.88 & 45.32 & 37.28 & 41.90 & 40.46 & 39.29 & 25.16 & 25.20\\

\multirow{-2}{*}{\centering\arraybackslash 10} & (0.0026) & (0.0020) & (0.0021) & (0.0017) & (0.0024) & (0.0020) & (0.0024) & (0.0045) & (0.0026) & (0.0028)\\
\cmidrule{1-11}
 & 11.17 & 45.43 & 48.18 & 45.24 & 34.90 & 42.38 & 41.53 & 34.06 & 16.56 & 16.94\\

\multirow{-2}{*}{\centering\arraybackslash 25} & (0.0019) & (0.0016) & (0.0014) & (0.0015) & (0.0021) & (0.0020) & (0.0026) & (0.0041) & (0.0030) & (0.0029)\\
\cmidrule{1-11}
 & 4.13 & 44.99 & 49.01 & 44.54 & 35.38 & 43.53 & 43.92 & 27.92 & 10.53 & 11.00\\

\multirow{-2}{*}{\centering\arraybackslash 50} & (0.0014) & (0.0014) & (0.0000) & (0.0015) & (0.0027) & (0.0020) & (0.0021) & (0.0038) & (0.0028) & (0.0029)\\
\cmidrule{1-11}
 & 0.76 & 44.97 & 49.68 & 43.85 & 38.31 & 44.98 & 45.96 & 21.30 & 6.22 & 6.76\\

\multirow{-2}{*}{\centering\arraybackslash 100} & (0.0000) & (0.0017) & (0.0000) & (0.0018) & (0.0028) & (0.0025) & (0.0016) & (0.0035) & (0.0021) & (0.0022)\\
\cmidrule{1-11}
 & 0.00 & 45.06 & 49.85 & 44.80 & 44.62 & 45.44 & 47.43 & 10.96 & 1.76 & 2.06\\

\multirow{-2}{*}{\centering\arraybackslash 250} & (0.0000) & (0.0015) & (0.0000) & (0.0014) & (0.0021) & (0.0026) & (0.0013) & (0.0025) & (0.0012) & (0.0012)\\
\cmidrule{1-11}
 & 0.00 & 44.97 & 49.94 & 44.38 & 48.02 & 45.48 & 48.44 & 4.19 & 0.20 & 0.27\\

\multirow{-2}{*}{\centering\arraybackslash 500} & (0.0000) & (0.0012) & (0.0000) & (0.0017) & (0.0012) & (0.0030) & (0.0010) & (0.0018) & (0.0000) & (0.0000)\\
\cmidrule{1-11}
 & 0.00 & 44.78 & 49.92 & 45.10 & 49.78 & 46.08 & 49.02 & 1.18 & 0.00 & 0.00\\

\multirow{-2}{*}{\centering\arraybackslash 1000} & (0.0000) & (0.0017) & (0.0000) & (0.0014) & (0.0000) & (0.0019) & (0.0000) & (0.0000) & (0.0000) & (0.0000)\\
\bottomrule
\end{tabular}}
\end{table}

\newpage

\begin{table}[!htb]
\centering
\caption{Estimated misclassification probabilities (in percentages) with standard errors (in parentheses) for $\delta_1$, $\delta_2$, $\delta_3$, and popular classifiers \textbf{Simulated Example 5}.}
\label{table-Ex5-updated}
\resizebox{\linewidth}{!}{
\begin{tabular}{ccccccccccc}
\toprule
\multirow[b]{2}{*}{\thead{\textbf{d}}}
        & \multicolumn{7}{c}{\thead{\textbf{Popular Classifiers}}}
            & \multicolumn{3}{c}{\thead{\textbf{Proposed Classifiers}}}\\
\cmidrule(lr){2-8}\cmidrule(lr){9-11}

& \textbf{Bayes} & \textbf{GLMNET} & \textbf{NN-RP} & \textbf{SVM-LIN} & \textbf{SVM-RBF} & \textbf{N-NET} & \textbf{1-NN} & $\bm{\delta_1}$ & $\bm{\delta_2}$ & $\bm{\delta_3}$\\
\midrule
 & 20.78 & 48.63 & 42.47 & 46.85 & 32.21 & 40.90 & 39.40 & 38.14 & 24.97 & 25.32\\

\multirow{-2}{*}{\centering\arraybackslash 5} & (0.0027) & (0.0041) & (0.0032) & (0.0039) & (0.0047) & (0.0053) & (0.0039) & (0.0048) & (0.0033) & (0.0034)\\
\cmidrule{1-11}
 & 11.77 & 48.20 & 45.82 & 47.60 & 28.02 & 40.71 & 40.31 & 33.56 & 17.26 & 17.71\\

\multirow{-2}{*}{\centering\arraybackslash 10} & (0.0022) & (0.0041) & (0.0026) & (0.0037) & (0.0048) & (0.0045) & (0.0029) & (0.0047) & (0.0033) & (0.0033)\\
\cmidrule{1-11}
 & 2.86 & 47.45 & 49.40 & 48.30 & 24.94 & 44.00 & 46.39 & 25.09 & 7.98 & 8.33\\

\multirow{-2}{*}{\centering\arraybackslash 25} & (0.0012) & (0.0038) & (0.0017) & (0.0035) & (0.0044) & (0.0044) & (0.0022) & (0.0043) & (0.0023) & (0.0023)\\
\cmidrule{1-11}
 & 0.36 & 46.66 & 49.91 & 47.84 & 23.55 & 45.47 & 48.87 & 17.22 & 3.08 & 3.43\\

\multirow{-2}{*}{\centering\arraybackslash 50} & (0.0000) & (0.0036) & (0.0015) & (0.0035) & (0.0035) & (0.0038) & (0.0021) & (0.0033) & (0.0015) & (0.0016)\\
\cmidrule{1-11}
 & 0.00 & 47.26 & 49.83 & 47.59 & 24.30 & 46.46 & 49.77 & 9.60 & 0.67 & 0.84\\

\multirow{-2}{*}{\centering\arraybackslash 100} & (0.0000) & (0.0038) & (0.0014) & (0.0032) & (0.0028) & (0.0036) & (0.0019) & (0.0025) & (0.0000) & (0.0000)\\
\cmidrule{1-11}
 & 0.00 & 46.08 & 50.00 & 48.01 & 29.72 & 48.06 & 50.33 & 2.56 & 0.02 & 0.04\\

\multirow{-2}{*}{\centering\arraybackslash 250} & (0.0000) & (0.0040) & (0.0017) & (0.0026) & (0.0026) & (0.0035) & (0.0020) & (0.0013) & (0.0000)& (0.0000)\\
\cmidrule{1-11}
 & 0.00 & 45.47 & 50.08 & 48.43 & 37.14 & 48.22 & 49.88 & 0.44 & 0.00 & 0.00\\

\multirow{-2}{*}{\centering\arraybackslash 500} & (0.0000) & (0.0030) & (0.0016) & (0.0030) & (0.0025) & (0.0037) & (0.0019) & (0.0000) & (0.0000) & (0.0000)\\
\cmidrule{1-11}
 & 0.00 & 45.25 & 49.84 & 48.59 & 44.47 & 49.34 & 49.90 & 0.01 & 0.00 & 0.00\\

\multirow{-2}{*}{\centering\arraybackslash 1000} & (0.0000) & (0.0028) & (0.0023) & (0.0032) & (0.0015) & (0.0042) & (0.0021) & (0.0000) & (0.0000) & (0.0000)\\
\bottomrule
\end{tabular}}
\end{table}


\end{document}